\documentclass[runningheads]{llncs}

\usepackage[mobile]{eccv}

\usepackage{eccvabbrv}

\usepackage{graphicx}
\usepackage{booktabs}
\usepackage{pifont}
\usepackage{makecell}
\usepackage{multirow}
\usepackage{titletoc}
\usepackage{tocloft}
\usepackage{placeins}
\usepackage[table]{xcolor}
\usepackage{listings}
\usepackage{xcolor}
\usepackage[shortcuts]{extdash}
\usepackage[strings]{underscore}
\usepackage{seqsplit}
\definecolor{myGreen}{RGB}{0, 140, 0}
\newcommand{\cmark}{\textcolor{myGreen}{\ding{51}}} 
\newcommand{\xmark}{\textcolor{black}{\ding{55}}} 
\providecommand{\authcount}[1]{}
\usepackage[accsupp]{axessibility}  

\usepackage{hyperref}

\usepackage{orcidlink}

\begin{document}

\title{MMEarth-Bench: Global Model Adaptation via Multimodal Test-Time Training} 

\titlerunning{MMEarth-Bench}

\author{Lucia Gordon\inst{1,2}\orcidlink{0000-0003-3219-6960} \and
Serge Belongie\inst{2}\orcidlink{0000-0002-0388-5217} \and
Christian Igel\inst{2}\orcidlink{0000-0003-2868-0856}\and Nico Lang\inst{2}\orcidlink{0000-0001-8434-027X}}

\authorrunning{L.~Gordon et al.}

\institute{Harvard University, USA \and
University of Copenhagen, Denmark
\\
\email{luciagordon@g.harvard.edu, nila@di.ku.dk}}

\maketitle

\begin{abstract}
  Recent research in geospatial machine learning demonstrates that models pretrained with self-supervised learning on Earth observation data can perform well on downstream tasks with limited labeled data. However, most benchmark datasets have few data modalities and poor global representation, limiting the ability to evaluate multimodal pretrained models at global scales. In order to fill this gap, we introduce \textsc{MMEarth-Bench}, a collection of five new environmental tasks with 12 modalities, globally distributed data, and both random and geographic test splits. We benchmark a diverse set of pretrained models and find that while (multimodal) pretraining tends to improve model robustness in limited data settings, geographic generalization abilities remain poor. Moreover, a simple randomly initialized multimodal model is competitive given enough labeled data. Although data is abundant, models can currently only make use of the modalities on which they were pretrained. To solve this problem, we propose using all the modalities available at test time as auxiliary tasks for test-time adaptation. Our model-agnostic method for test-time training with multimodal reconstruction (\textsc{TTT-MMR}) can improve performance across all models and tasks on both test splits. Furthermore, geographic batching leads to a good trade-off between regularization and specialization during TTT, which can be beneficial for long-tail distributions. Our dataset, code, and visualization tool are linked on the project page: \href{https://lgordon99.github.io/mmearth-bench/}{lgordon99.github.io/mmearth-bench}.
\end{abstract}

\section{Introduction}
\label{sec:intro}

\begin{figure}[t]
    \centering
    \includegraphics[width=\linewidth]{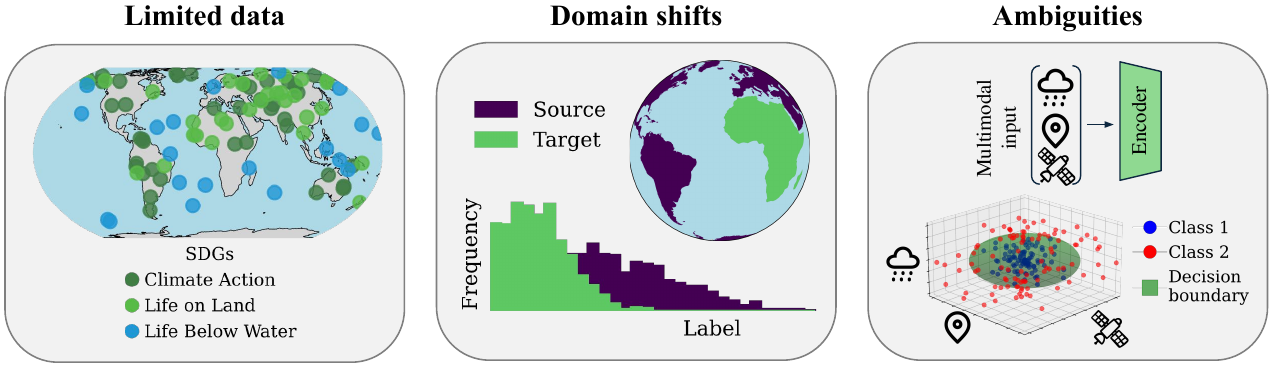}
    \caption{\textbf{Self-supervised multimodal pretraining promises to overcome three grand challenges in Earth observation.} Crucial applications have to rely on \emph{limited and sparse} and \emph{geographically biased} training data. Furthermore, the \emph{ambiguities} inherent to modeling biophysical quantities with remotely sensed data may be resolved by models conditioned on \emph{multiple modalities}.}
    \label{fig:teaser}
\end{figure}

Earth observation (EO) data helps us monitor our planet's health, respond to natural disasters, and tackle societal challenges such as the Sustainable Development Goals (SDGs) \cite{persello2022deep}. A multitude of sensors provide unprecedented data about the planet but also far exceed what most data processing pipelines can handle.
To realize the full potential of EO data, raw observations, such as optical satellite images, must be interpreted at both local \emph{and} global scales. Deep learning seems like a natural approach to analyzing copious unlabeled data, and supervised learning has shown great success in the case of abundant reference data \cite{BookDLRS2017}, but we identify \emph{three grand challenges} yet to be overcome (see \cref{fig:teaser}). 
The first long-standing challenge has been the sparsity and limited amount of labeled data, as is often the case in crucial applications where reference data is collected through field measurements~\cite{BookDLRS2017}.
The second challenge is geographic generalization, \ie the domain shift that occurs when using a model trained on one geographic region to make predictions on another~\cite{pangaea}. 
Lastly, the third main challenge is to minimize the ambiguities inherent to estimating biophysical quantities from observational, especially unimodal, EO data~\cite{using_multiple_modalities}.
The research community working at the intersection of machine learning and EO has identified self-supervised learning (SSL) as a promising way forward~\cite{tseng2025galileo, anysat, jakubik2025terramind, mmearth}. Pretraining models with global unlabeled data promises to facilitate model adaptation to tasks with limited labels and ameliorate performance drops in geographic regions without reference data. Because EO data is georeferenced, the multitude of sensors and satellites means that there are myriad data modalities that can be automatically aligned~\cite{mmearth}. This is an opportunity for multimodal data fusion to resolve ambiguities in model predictions~\cite{Sialelli2025, NEURIPS2024_e4e7de47, teng2023satbird, 10943541}. Hence, EO is relevant to the computer vision community for advancing multimodal representation learning~\cite{4m}.
While existing EO benchmarks \cite{copernicusfm,pangaea,phileo,fomo,geobench} have helped to develop and evaluate pretrained geospatial models, they limit the evaluation of progress towards overcoming these three grand challenges, as most of their downstream tasks are i) regional rather than global, ii) not evaluated on geographic splits with an entire held-out region, and iii) only provide few modalities per task.

In this work we introduce \textsc{MMEarth-Bench}, a multimodal EO benchmark dataset containing five global tasks designed to evaluate pretrained geospatial models in these three challenging yet crucial settings. As in prior work, we use the term ``modality'' to describe sensor measurements, metadata, and derived products~\cite{mmearth}. We evaluate 8 pretrained models on our benchmark and find that multimodal pretraining does improve performance in low-shot settings and under geographic shifts. However, we show that given enough labeled data, training a multimodal model from scratch is a strong baseline. As is the case with our evaluation and more generally, there is a mismatch between the \emph{input modalities} on which a model was pretrained and the \emph{task modalities} available at inference time~\cite{pangaea}. The standard practice is simply to use only the task modalities that the pretrained model can take as input.
To address this inefficiency, we introduce a method for \emph{using all available modalities at test time, whether or not they are compatible with the pretrained model}. Complementary to using multimodal data for self-supervised pretraining, we propose using the modalities as reconstruction tasks in the regime of test-time adaptation, which uses unlabeled test data to adapt the model at test time. We formulate our proposed method for Test-Time Training with MultiModal Reconstruction (\textsc{TTT-MMR}), along with a second variant using geographic batching (\textsc{TTT-MMR-Geo}), which leads to a good trade-off between specialization and regularization, a crucial aspect in TTT \cite{goyal2022conjugate,tan2025searchtta}. We summarize our major contributions:

\begin{enumerate}
    \item \textbf{MMEarth-Bench dataset:} We introduce five multimodal environmental monitoring tasks for benchmarking pretrained models. Each task consists of 12 aligned modalities and provides globally distributed data with both random and geographic test splits.
    \item \textbf{Benchmarking:} We benchmark 8 recent pretrained models covering a range of architectures and including both uni- and multimodal models. We find that all models still struggle to generalize geographically and that training a multimodal model from scratch is competitive with global pretraining.
    \item \textbf{Method:} To advance model adaptation when labels are limited and geographically biased, we propose repurposing multimodal pretext tasks for test-time adaptation, using modality reconstruction losses as a test-time adaptation signal. Our model-agnostic approach yields performance improvements across all models and tasks on both test splits.
\end{enumerate}

\section{Related Work}
\subsection{Low-shot learning}
\label{subsec:low-shot-learning}
\noindent\textbf{Pretrained models.}
One strategy to facilitate low-shot learning is self\-/supervised pretraining. In contrast to vision foundation models trained on web data \cite{simeoni2025dinov3,imagenet}, EO models are pretrained on large amounts of unlabeled remotely sensed data. Hong \etal~\cite{hong2026foundation} provide an overview of the myriad foundation models in remote sensing. While some geospatial models are still pretrained on RGB imagery~\cite{GeoKR,gassl,seco,scalemae,simeoni2025dinov3}, more recent unimodal models include additional spectral bands
~\cite{satmae2022,satlas,mmearth}. We benchmark various pretrained models for global adaptation with limited and geographically biased labels, as found in real applications.

\textbf{Multimodal fusion.}
Subsequently, additional work aimed to learn richer geospatial representations by pretraining on multimodal data ~\cite{dofa,anysat,jakubik2025terramind,copernicusfm} or training multimodal models from scratch~\cite{using_multiple_modalities}. We evaluate the benefits of modality fusion in \emph{pretrained} models under geographic shifts. Moreover, whereas multimodal models use multiple modalities as input, our method uses them as \emph{targets}.

\textbf{Joint training (JT).}
Joint training trains a model simultaneously on both the main task and auxiliary tasks in order to improve robustness on the main task, especially when training data is limited~\cite{ssl_few_shot_learning,hendrycks,sun19ttt}. The model has a shared encoder and separate heads for the main task and self-supervised tasks~\cite{ssl_few_shot_learning}. This method only uses multimodal data to generate reconstruction losses at train time. In contrast, our method also leverages the reconstruction losses as an \emph{adaptation signal at test time}. We use JT as a baseline in our TTT experiments. 
 
\subsection{Domain adaptation}
\textbf{Unsupervised domain adaptation (UDA).}
Unsupervised domain adaptation uses labeled data from the source domain and unlabeled data from the target domain to adapt to the target domain \cite{uda_review}. UDA-SS \cite{udass} extends JT by including unlabeled target data during the training process to facilitate the encoder's adaptation to the target domain. Like many common UDA methods, this is primarily designed for covariate shifts across domains~\cite{adversarial_uda,uda_review}. The geographic generalization problem differs, as we also encounter \emph{label} distribution shifts. Scheibenreif \etal \cite{geospatial_domain_adaptation} insert adapters into a pretrained encoder and train them with SSL using target domain modalities before finetuning the model with target domain labels. In contrast, our problem setup assumes that we neither have access to labeled nor unlabeled data from the target domain during training. Our model-agnostic approach also avoids modifying the encoder architecture.

\textbf{Test-time adaptation (TTA).}
Unlike traditional UDA which uses target distribution data during training, test-time adaptation, or equivalently test-time training (TTT), updates the model during testing to address distribution shifts~\cite{sun19ttt,gandelsman2022testtime}.
Following JT, rather than discarding the self-supervised task head~\cite{udass}, TTT uses this head at test time to calculate a loss for adapting the encoder before making a final prediction~\cite{sun19ttt,gandelsman2022testtime}. Past works have used rotation prediction \cite{sun19ttt} or masked reconstruction \cite{gandelsman2022testtime} as auxiliary tasks.
To avoid having to define a self-supervised auxiliary task, TENT \cite{wang2021tent} uses the prediction entropy as an adaptation signal. This makes it a natural approach for classification tasks but not for regression, which is common for biophysical variables (4 of our 5 tasks). Unlike TENT, which updates the encoder's batch norm statistics, our proposed TTT method works with any architecture and loss. This is especially relevant as many pretrained EO models use a transformer architecture without batch norm layers. 
Sparse or coarse labels have also been used to produce a TTA signal~\cite{rna}. In the spirit of using weak labels for TTT, we use the data modalities available at test time as reconstruction targets.
While TTA has been used with remote sensing imagery to adapt to distribution shifts in street-level RGB imagery~\cite{tta_remote_sensing} and guide robots’ search for targets in an environment~\cite{tan2025searchtta}, we adapt pretrained models to \emph{diverse} environmental tasks and domains at \emph{global} scales.

\textbf{Regularization in TTA.}
Regularization is crucial for TTA \cite{goyal2022conjugate,tan2025searchtta}, and simply using batched test data is a regularization method, as it results in less noisy gradients~\cite{nano_adapt,stable_tta}. Test batches can be comprised of multiple augmentations of a single image~\cite{sun19ttt} or just be random samples of the test data \cite{gandelsman2022testtime}. We propose geographic batching as a compromise between regularization and specialization. Updating only the batch norm parameters~\cite{wang2021tent} or FiLM layers~\cite{perez2018film,rna} are commonly used regularization approaches. In model-agnostic fashion, our method's adaptation signal uses 12 modality reconstruction losses, as opposed to just one, for regularization. Each individual modality's gradient may be noisy, but averaging over them reduces noise. In this way, we avoid needing to insert batch norm or FiLM layers. The S4T method~\cite{synchronizing_task_behavior} uses a vision transformer as a task behavior synchronizer to lessen degradations in model performance due to the ideal number of TTT iterations differing among multiple auxiliary tasks. We achieve regularization with a lightweight linear task modality decoder.

\textbf{Multimodal TTA.}
Shin \etal~\cite{mmtta} study multimodal TTA for 3D semantic segmentation by producing pseudo-labels for use as a self-training signal at test time given RGB images and LiDAR point clouds. Like TENT, their method updates batch norm parameters in the architecture. READ~\cite{tta_multimodal_reliability} addresses the scenario where modalities vary in their reliability when subject to domain shifts. They encode each modality with a transformer and modulate attention-based fusion layers at test time. Unlike READ, our approach does not require encoding the auxiliary modalities, instead solely using them as reconstruction targets. 

\subsection{Existing benchmark datasets}
\begin{table}[b]
\centering
\caption{\textbf{Comparison of benchmark datasets for EO (vision) models.} ``M'': number of modalities per task. ``\cmark/\xmark'': a property holds for a subset of tasks.}
\label{tab:benchmark-datasets}
\resizebox{0.6\columnwidth}{!}{%
\setlength{\tabcolsep}{5pt}
\begin{tabular}{@{} 
    l   
    l   
    r   
    c   
    c   
@{}}
\toprule
\textbf{Benchmark} & 
\textbf{Domain} &
\textbf{M} & 
\textbf{Global} &
\textbf{Climate} \\
\midrule
Copernicus-Bench~\cite{copernicusfm} & Mixed & 1 & \cmark/\xmark & \cmark/\xmark \\
PANGAEA~\cite{pangaea} & Mixed & 1-3 & \cmark/\xmark & \xmark  \\
PhilEO Bench~\cite{phileo} & Built-up & 1 & \cmark & \xmark  \\
FoMo-Bench~\cite{fomo} & Forests & 1-4 & \cmark/\xmark & \cmark/\xmark \\
GEO-Bench~\cite{geobench} & Mixed & 1-3 & \xmark & \xmark \\
SustainBench~\cite{yeh2021sustainbench} & SDGs & 1-10 & \cmark/\xmark & \xmark \\ \midrule
MMEarth-Bench (ours) & Nature & 12 & \cmark & \cmark \\
\bottomrule
\end{tabular}
}
\end{table}

\textbf{Overview.}
Past work in benchmark datasets for pretrained EO models has guided progress in model development and pretraining strategies. We summarize the most related benchmarks in \cref{tab:benchmark-datasets}. 
\emph{SustainBench}~\cite{yeh2021sustainbench} is a collection of 15 tasks designed to track progress in 7 sustainable development goals, including no poverty, quality education, and climate action. The modalities vary by task and can include imagery from Landsat, Sentinel-2 (S2), MODIS, NAIP, as well as geolocation and time. 
\emph{GEO-Bench}~\cite{geobench} contains 12 tasks ranging from building and solar panel classification to tree segmentation. The modalities vary by task and can include S2, Sentinel-1 (S1), Landsat-8, RGB, hyperspectral imagery, or elevation data. None of the tasks are globally distributed. 
\emph{FoMo-Bench}~\cite{fomo} contains 15 datasets for forest monitoring, where each task has up to 4 modalities such as S2, S1, LiDAR, elevation, or meteorological data. 
\emph{PhilEO Bench}~\cite{phileo} contains task data for building density estimation, road segmentation, and landcover classification. All tasks have S2 as the single input modality. 
\emph{Copernicus-Bench}~\cite{copernicusfm} is a benchmark focusing on cross-modal model evaluation that includes 15 unimodal tasks including land cover, biomass, and air quality. The input modality varies across tasks and can include sensors such as Sentinel-1, -2, -3, or -5P. PANGAEA~\cite{pangaea} is a collection of existing benchmark datasets, each of which has up to 3 modalities such as S2, S1, Planet, or Maxar. 

\textbf{Limitations.}
None of the existing benchmark datasets share the same multiple modalities across all downstream tasks, and the vast majority of tasks contain no more than 3 modalities. This limits the evaluation of multimodal pretrained models, since likely not all of their modalities are available at test time. Furthermore, the vast majority of downstream tasks are limited to a single geographic region. Even if a benchmark contains tasks covering different regions, this does not make it possible to determine whether a pretrained model is able to generalize from one region to another. While PANGAEA explicitly mentions a geographic test split for one task and SustainBench includes geographic splits for several tasks, benchmarks rarely explicitly state whether a test split represents an entire held-out region or simply avoids overlap with the training data. This is likely due to a combination of geographic generalization not being prioritized and many benchmarks collecting existing datasets and using their data splits. The degree to which a geographic test split can be a distribution shift is also limited by the geographic spread of the data in the first place. Additionally, few downstream tasks include climate data as a modality, limiting development in fusion methods for optical EO and climate data. While there are several downstream tasks focusing on the natural world, many tasks focus on the human-nature interface and the built-up world. Moreover, the same tasks appear in multiple benchmarks, most of which include preexisting datasets. We contribute five novel, global datasets for evaluating pretrained models on biomass, soil property, and species occurrence prediction. Our tasks focus on the natural world with 12 shared modalities across imagery, map products, and climate variables. These design choices should facilitate future development of multimodal models.

\section{Dataset}
\textbf{Overview.}
\textsc{MMEarth-Bench} is comprised of five new datasets, summarized in \cref{tab:downstream-tasks}, for environmental tasks: aboveground biomass, soil nitrogen (N), soil organic carbon (OC), soil pH, and species occurrence. Each single-timestamp dataset includes 12 modalities along with task data for each 128$\times$128-pixel tile. Every pixel represents 10m on the ground, so each tile spans an area of $\approx$ 1.6~km$^2$. All five tasks share the same 12 modalities (see \cref{tab:modalities}) present in the MMEarth pretraining dataset~\cite{mmearth}. Within each task, we ensure no overlap among tiles. Note that the \textsc{MMEarth-Bench} datasets are not designed for producing the next best task-specific model or for time-series modeling. Rather, we design our datasets for systematic evaluation of pretrained models that aim to generalize to various downstream tasks with limited data under geographic shifts.

\begin{figure*}[t]
    \centering
\includegraphics[width=\linewidth]{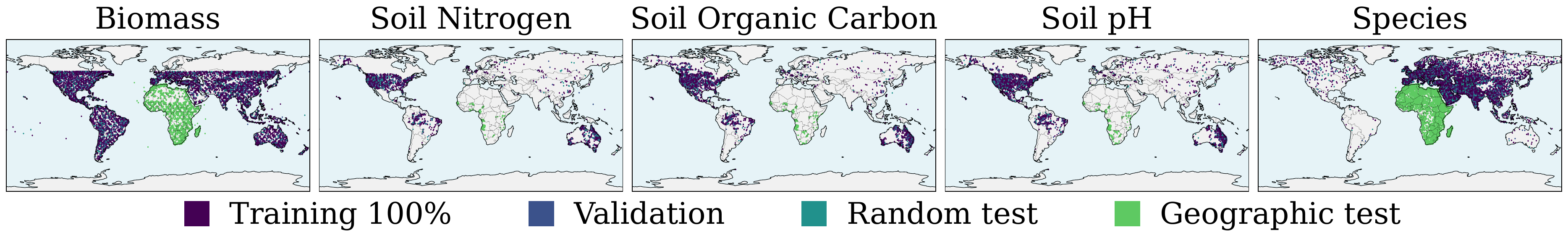}
    \caption{\textbf{Data splits in \textsc{MMEarth-Bench}.} Each of the 5 tasks consists of a geographic test split (``Africa'') and splits the rest of the world randomly into training (70\%), validation (15\%), and random test (15\%) sets. Subsets with 50\% and 5\% of the training data for even lower-shot experiments can be seen with our \href{https://mmearth-bench.com/\#explorer}{Explorer}.}
    \label{fig:split_maps}
\end{figure*}

\textbf{Splits.}
Each task's tiles are split into train, validation, random test, and geographic test sets (see \cref{fig:split_maps}). The tiles in Africa define the geographic test set, as it is often underrepresented in labeled data derived from field measurements, as is the case for our soil tasks~\cite{inat,WEF_2021_EO_Africa}. In practice, models finetuned with non-Africa data might be used to make predictions in Africa, making it especially relevant to understand how well this generalization works. The remaining tiles are randomly split into train/validation/random test with ratios 70\%/15\%/15\%. We include a random and a geographic test set to evaluate model performance both in- and out-of-distribution. To allow for evaluating models in very low data regimes, we also provide train sets containing 50\% and 5\% of the training tiles, selected randomly but where the 5\% train set is a subset of the 50\% train set.

\begin{table}[t]
\centering
\caption{\textbf{\textsc{MMEarth-Bench} tasks.}}
\label{tab:downstream-tasks}
\resizebox{0.8\columnwidth}{!}{%
\setlength{\tabcolsep}{5pt}
\begin{tabular}{@{} 
    l   
    r   
    l   
    l   
    l   
    l   
@{}}
\toprule
\textbf{Task} & 
\textbf{\# Tiles} & 
\textbf{Unit} &
\textbf{Scale} & 
\textbf{Type} &
\textbf{License} \\
\midrule
\textbf{Biomass}              & 18{,}393 & Mg/ha & Pixel-level & Regression & CC BY \\
\textbf{Soil Nitrogen}        &  5{,}643 & g/kg & Tile-level  & Regression & CC BY-NC\\
\makecell[l]{\textbf{Soil Organic Carbon}}  &  7{,}982 & g/kg & Tile-level  & Regression & CC BY-NC \\
\textbf{Soil pH}              &  8{,}508 & Unitless & Tile-level  & Regression & CC BY-NC \\
\textbf{Species}   &  36{,}410 & \makecell[l]{Occurrence} & Tile-level & \makecell[c]{Multi-label classification} & \href{https://www.iucnredlist.org/terms/terms-of-use}{Terms of Use} \\
\bottomrule
\end{tabular}
}
\end{table}

\begin{table*}[t]
\centering
\caption{\textbf{\textsc{MMEarth-Bench} modalities.}}
\label{tab:modalities}
\resizebox{!}{1.12cm}{
\setlength{\tabcolsep}{5pt}
\begin{tabular}{ll}
\multicolumn{2}{c}{\textbf{Pixel-level}}\\
\toprule
\textbf{Modality} & \textbf{Bands / Variables} \\
\midrule
\textbf{Sentinel-2 (S2)} &
B1--B8, B8A, B9, B11, B12 \\
\textbf{Sentinel-1 (S1)} &
(Asc., Desc.) $\times$ (VV, VH, HH, HV) \\
\textbf{ASTER GDEM} &
Elevation, slope \\
\textbf{ETH Canopy Height} &
Height, uncertainty \\
\textbf{Dynamic World} &
Landcover (9 categories) \\
\textbf{ESA WorldCover} &
Landcover (11 categories) \\
\bottomrule
\end{tabular}
}
\hfill
\resizebox{!}{1.12cm}{
\setlength{\tabcolsep}{5pt}
\begin{tabular}{ll}
\multicolumn{2}{c}{\textbf{Tile-level}}\\
\toprule
\textbf{Modality} & \textbf{Bands / Variables} \\
\midrule
\textbf{Precipitation} &
Last month, month, year \\
\textbf{Temperature} &
(Last month, month, year) $\times$ (max, mean, min) \\
\textbf{Geolocation} &
Longitude, latitude \\
\textbf{S2 Date} &
Date \\
\textbf{Biome} &
Biome (14 categories) \\
\textbf{Ecoregion} &
Ecoregion (846 categories) \\
\bottomrule
\end{tabular}
}
\end{table*}

\textbf{Modalities.}
All tiles include 12 modalities, six of which are pixel-level (S2, S1, ASTER GDEM \cite{aster_dem}, ETH Global Canopy Height \cite{lang2023high}, Dynamic World \cite{brown2022dynamic} and ESA WorldCover \cite{zanaga_2021_5571936}) and six of which are tile-level (ERA5 precipitation and temperature, geolocation, Sentinel-2 date, biome and ecoregion \cite{ecoregions}). Eight of the modalities are continuous-valued and four are categorical-valued. All of the modality data is accessed through Google Earth Engine (GEE) \cite{GORELICK201718}.

\textbf{Tasks.}
Our biomass dataset is sourced from aboveground biomass estimates from NASA's 2020 GEDI mission~\cite{dubayah2020global}. We sample measurements with GEE such that our tiles are approximately balanced across biomes. We obtain our soil data from the WoSIS December 2023 snapshot~\cite{wosis}. We select soil N, soil OC, and soil pH due to their environmental and ecological significance. Nitrogen is crucial for plant growth, organic carbon is an indicator of soil quality, and pH has a strong influence on biogeochemical processes in soil~\cite{butterfly}. For each soil property we aggregate data across time within the depth range 0-5 cm. Biomass and the soil properties are regression tasks. We extract our species range data from the IUCN Red List's terrestrial mammal range polygons~\cite{iucn}. We filter by species whose historic range covers at least 6,000 km$^2$ both in and outside Africa and then take the first 100 species. Sampling tiles from these overlapping ranges yields multiple species per tile, making this a multi-label classification task. The minimum number of tiles in which a species appears by split is 187 for train 100\%, 101 for train 50\%, 7 for train 5\%, 32 for validation, 46 for random test, and 173 for geographic test. See the Appendix for more details.

\section{Method}
\begin{figure}[t]
    \centering
    \includegraphics[width=0.75\linewidth]{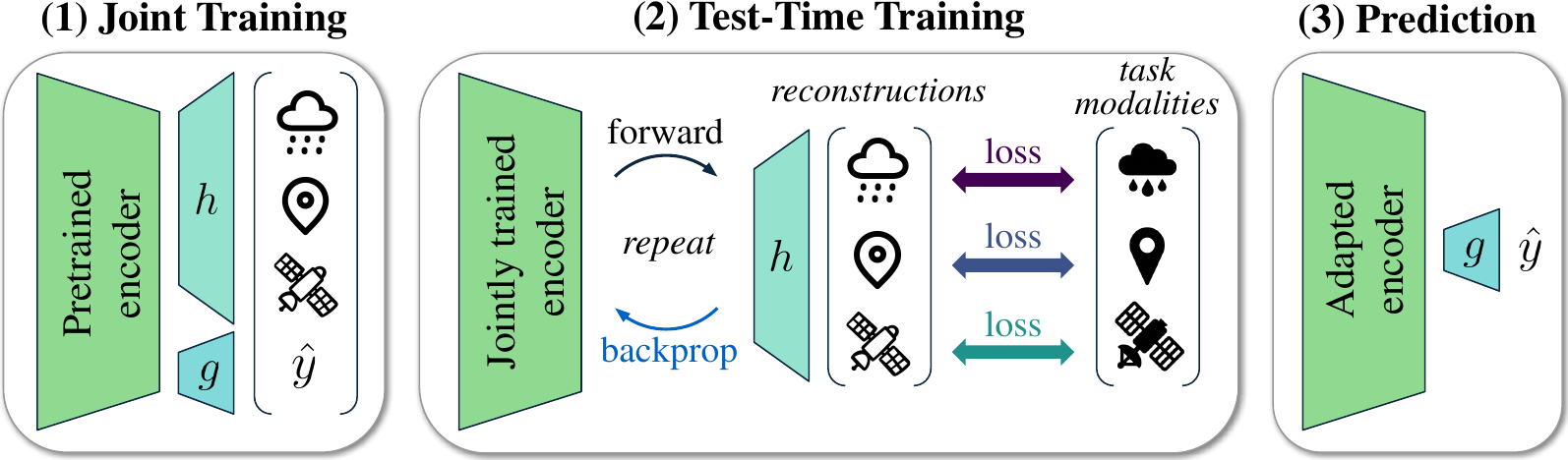}
    \caption{\textbf{TTT with multimodal reconstruction (\textsc{TTT-MMR}).} 
    (1) A (pretrained) \emph{encoder} is jointly trained with the \emph{main task decoder} $g$ and \emph{task modality decoder} $h$ to get a prediction $\hat{y}$ for the main task and a reconstruction $\hat{d}_m$ for every task modality $m\in M$. Then for each batch at test time, (2) the modality reconstruction losses $\{R_m\}_{m\in M}$ are used as an adaptation signal to update the encoder iteratively, and (3) the adapted encoder is used to yield improved predictions for the main task.}
    \label{fig:ttt_diagram}
\end{figure}

\textbf{Multimodal adaptation signal.}
We propose a method for multimodal test-time training, \textsc{TTT-MMR}, in which we formulate reconstruction tasks using the \emph{task modalities}, \ie the set of modalities available at test time. As we do \emph{not} use test labels, we treat these modalities as proxies for our task of interest. While we know that canopy height is correlated with biomass~\cite{lang2023high,dubayah2020global}, for example, it is generally nontrivial to preselect the best proxies for a given task. Thus, we propose a method for using all available modalities in a test-time adaptation signal.
A task modality decoder is used to reconstruct all the task modalities given the encoder's embedding of the \emph{input modalities}, \ie the subset of modalities accepted by the encoder. For a given task, we first perform joint training, during which we train the task modality decoder with the mean of the modality reconstruction losses, the main task decoder with the task loss, and the encoder with the sum of these two losses.
Then at test time the reconstruction losses serve as an adaptation signal for the encoder (see \cref{fig:ttt_diagram}). \textsc{TTT-MMR} normalizes the gradients of the task modality reconstruction losses by modality and then averages them across modalities (excluding missing modalities) so that each modality contributes equally to the adaptation signal, regardless of its original scale. We freeze the task modality decoder to force the encoder to adapt.

\textbf{Formalization.}
Each batch $B$ contains a set of $|B|$ tiles. Each tile $t\in B$ has data for the input modalities, collectively $i_t$, and data for each task modality $m\in M$ given by $d_{m,t}$. The encoder is a function $e$ with parameters $\theta$ of the input modalities, which produces input embeddings $\{e_\theta(i_t)\}_{t\in B}$. The task modality decoder is a function $h$ with parameters $\alpha$ of the input embeddings, which produces modality reconstructions $\{\hat{d}_{m,t}=h_\alpha(e_\theta(i_t))_{m,t}\}_{m\in M,t\in B}$. Let $R_{m,t}(d_{m,t},\hat{d}_{m,t})$ be the reconstruction loss for tile $t$'s modality $m$. For each modality, the reconstruction loss is averaged across all tiles in the batch: $\textstyle R_m = \frac{1}{|B|}\sum_{t\in B}R_{m,t}(d_{m,t},\hat{d}_{m,t})$.
We take the gradient of each per-modality reconstruction loss with respect to the encoder parameters: 
$\textstyle\frac{\partial R_m}{\partial\theta}=\frac{\partial R_m}{\partial h_\alpha}\frac{\partial h_\alpha}{\partial e_\theta}\frac{\partial e_\theta}{\partial\theta}$.
To put equal weight on each modality, we separately normalize their gradients, 
$\frac{\partial R_m}{\partial\theta}\to\frac{1}{\|\frac{\partial R_m}{\partial\theta}\|}\frac{\partial R_m}{\partial\theta}$,
before averaging the gradients across modalities. We adapt the parameter $\theta_j$ by backpropagating this mean gradient using stochastic gradient descent with learning rate 
$\lambda$: $\textstyle\theta_j\to\theta_j-\frac{\lambda}{|M|}\sum_{m\in M}\frac{1}{\|\frac{\partial R_m}{\partial\theta}\|}\frac{\partial R_m}{\partial\theta_j}$.

In an iterative optimization process, we can use the updated encoder to generate new embeddings and reconstructions to compute new reconstruction losses and gradients, which can then update the encoder again. After TTT, the adapted encoder $e_{\theta'}$ produces embeddings that the task decoder $g$ uses to generate a final prediction $\hat{y}=g(e_{\theta'}(i))$, after which the encoder is reset to its post-JT state for the next batch. We perform $I_{\text{max}}$ iterations on every batch in the validation set and then use the mean of the batches' best iteration number during testing.
In our experiments, we use $|B|=8$, $\lambda=10^{-2}$, and $I_{\text{max}}=5$.

\textbf{Batching.}
We compare \textsc{TTT-MMR}, in which batches are random, non-overlapping samples of the test data, with \textsc{TTT-MMR-Geo}, which batches the test data based on geographic proximity as a proxy for sample similarity. 
The latter allows more specialization to the geographic domain defined by the batch, at the cost of some regularization that comes from averaging over more geographically diverse tiles. We generate the geographic batches through recursive spatial partitioning, mimicking a k-d tree~\cite{kdtree}. We recursively subdivide the region containing the split's tiles until only one subregion has more than $|B|$ tiles. This results in non-overlapping batches that are contiguous geographic regions.

\section{Experiments}
We address the following research questions: (1) \textit{How well can pretrained models transfer to downstream tasks with limited reference data?} (2) \textit{Can pretrained models generalize geographically?} (3) \textit{When do pretrained multimodal models benefit from using multiple input modalities at test time?} Lastly, we show that multimodal TTT improves models with limited and geographically biased data.
    
\label{sec:experiments}
\subsection{Setup}
\textbf{Benchmarking.} 
\begin{table}[t]
\centering
\caption{\textbf{Pretrained models benchmarked.}}
\label{tab:pretrained-encoders}
\rowcolors{2}{gray!15}{white}
\resizebox{\columnwidth}{!}{%
\setlength{\tabcolsep}{5pt}
\begin{tabular}{@{} 
    l   
    l   
    l   
    l   
@{}}
\toprule
\textbf{Method} & 
\textbf{Architecture} & 
\textbf{Pretraining dataset} & 
\textbf{Input modalities} \\
\midrule
Scale-MAE~\cite{scalemae} & ViT-L & FMoW-RGB~\cite{fmow} & RGB \\
DINOv3 Web~\cite{simeoni2025dinov3} & ViT-L/16 distilled & LVD-1689M~\cite{simeoni2025dinov3} & RGB \\
DINOv3 Sat~\cite{simeoni2025dinov3} & ViT-L/16 distilled & SAT-493M~\cite{simeoni2025dinov3} & RGB \\
SatlasNet~\cite{satlas} & Swin-v2-B & SatlasPretrain~\cite{satlas} & S2 \\
MPMAE~\cite{mmearth} & ConvNeXt V2-A & MMEarth~\cite{mmearth} & S2 \\
TerraMind~\cite{jakubik2025terramind} & TerraMindv1-B & TerraMesh~\cite{terramesh} & S2, S1, DEM, RGB \\
Copernicus-FM~\cite{copernicusfm} & ViT-B & Copernicus-Pretrain~\cite{copernicusfm} & S2, S1, DEM, Geolocation, Date \\
Galileo~\cite{tseng2025galileo} & ViT-B & Galileo dataset~\cite{tseng2025galileo} & 
\makecell[l]{S2, S1, NDVI, Temperature, Precipitation, \\DEM, Dynamic World, Geolocation, Month} \\
\bottomrule
\end{tabular}
}
\end{table}
We benchmark 8 pretrained models: 3 RGB-only, 2 S2-only, and 3 multimodal models (see \cref{tab:pretrained-encoders}). For comparison, we also evaluate a randomly initialized ConvNeXtV2A model with RGB input and another variant (ConvNeXtV2A-MM) with all modalities as input. We implement all models with a linear task decoder. For the tile-level tasks, the task decoder consists of global average pooling followed by layer norm and a fully connected layer, and for the pixel-level task, it bilinearly upsamples the embeddings to the tile size and then applies a convolutional layer with a $1\times1$ kernel. We finetune by training all model parameters. Linear probing results are provided in the Appendix.

\textbf{Multimodal models.}
The TerraMind encoder contains a series of blocks in which modality-specific tokens are conditioned on one another through the attention mechanism. For modality fusion, we use the default implementation, which averages the patch embeddings from the final encoder block across modalities. In contrast, cross-modal Copernicus-FM does not have per-modality encoding. We implement it as a Siamese network, sharing parameters across modalities. We separately compute embeddings for S2, S1, and DEM and then average them. Galileo linearly projects the modalities separately into tokens, adds modality-specific embeddings to them, concatenates all the tokens into a single sequence, and then performs self-attention in shared transformer blocks. For ConvNeXtV2A-MM we one-hot-encode the categorical modalities, add spatial dimensions to the tile-level modalities, stack all the pixel- and tile-level bands in a single tensor, and set the number of input channels to 926.

\textbf{JT and TTT.}
JT and TTT use a linear task modality decoder that bilinearly upsamples the input embeddings to the tile size and then passes them through a 2D convolution with a $1\times1$ kernel that reconstructs every band in every task modality at the tile resolution. We group the channels by modality and perform global average pooling on the bands of the tile-level modalities, yielding the modality reconstructions. For the reconstruction losses we use cross-entropy loss and MSE loss for the categorical- and continuous-valued modalities, respectively.

\textbf{Implementation.}
We run our experiments on an NVIDIA H200 140GB GPU. The regression tasks use MSE loss and species uses BCE with logits loss. Our hyperparameter settings are specified in the Appendix. In all results, ``performance'' refers to R$^2$ for the regression tasks and mean average precision (mAP) for species. Following training, we select the checkpoint with the highest performance on the validation set and report its results on both test sets. The input data is normalized according to the pretrained model's normalization method and statistics or center-normalized using the mean and STD of the bands in the training set for the randomly initialized models and the task modalities. Shaded regions in plots reflect 1 standard error across three distinct random seeds.

\subsection{Results}

\begin{figure}[t]
    \centering
    \includegraphics[width=\columnwidth]{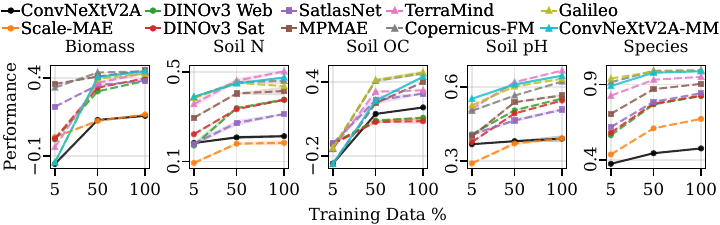}
    \caption{\textbf{Low-shot in-distribution performance.} Finetuning on training subsets. Symbology: $\bullet$=RGB, $\blacksquare$=S2, $\blacktriangle$=multimodal, solid=random init., dashed=pretrained.}
    \label{fig:rq1}
\end{figure}
\begin{figure}[t]
    \centering
    \includegraphics[width=\columnwidth]{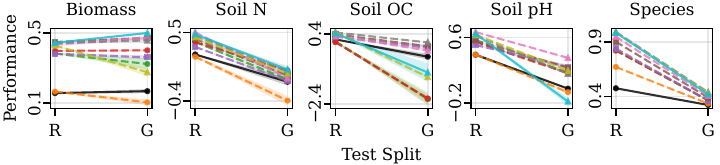}
    \caption{\textbf{Geographic generalization.} Performance comparison on random (R) \vs geographic (G) test splits using all training data.}
    \label{fig:rq2}
\end{figure}

\textbf{Finetuning with limited reference data.}
We finetune all 8 pretrained models and the randomly initialized ConvNeXtV2A models on three subsets of the training data: 5\%, 50\%, and 100\%. We evaluate the models on the random, in-distribution test split to determine how effectively they transfer to the five tasks in \textsc{MMEarth-Bench}, with results shown in \cref{fig:rq1}. Overall, the models that take multimodal input data outperform the unimodal models and especially the RGB-only models.
MPMAE, an S2-only model, shows competitive or superior performance to the multimodal models on the biomass and soil OC tasks especially. Surprisingly, the randomly initialized ConvNeXtV2A-MM model generally performs on par with the pretrained multimodal models, ranking 1st when averaging rankings across tasks for 100\% training data (see Appendix). As expected, its performance relative to the pretrained multimodal models worsens with less training data, but it still ranks 4th overall with 5\% training data, only beaten by the other multimodal models. The RGB-only ConvNeXtV2A model ranks poorly overall, but it performs similarly to or better than at least one of the RGB-pretrained models on 4 out of 5 tasks. Biomass and soil OC appear to be most sensitive to limited training data, with none of the models achieving a positive R$^2$ with only 5\% of the training data on the soil OC task, which existing work in soil modeling has found to be challenging~\cite{soc_england}. Comparing DINOv3 Web and Sat shows that the domain-specific pretraining of DINOv3 Sat had little benefit on our tasks.
Copernicus-FM originally used multimodal inputs during linear probing and unimodal inputs during finetuning in a cross-modal evaluation~\cite{copernicusfm}. We provide a \emph{multimodal} finetuning evaluation of Copernicus-FM.

\textbf{Geographic generalization challenge.}
Next, we evaluate all the models on the geographic test split after finetuning on all the training data to determine how effectively they generalize to Africa, for which they have not seen labels. As shown in \cref{fig:rq2}, for all tasks except biomass the models perform significantly worse on the geographic test split than on the random one, despite being pretrained globally. 
Overall, the multispectral models perform best, though for soil N and soil OC, even the multimodal models struggle on the geographic split, suggesting that these properties are especially challenging to predict in new regions. The randomly initialized ConvNeXtV2A-MM ranks 4th overall, while the multimodal pretrained Galileo model ranks 5th. 
ConvNeXtV2A-MM's lower ranking on the geographic than the random test split can be explained by i) it has not seen any pretraining data from Africa and ii) being trained on all the modalities may increase the risk of overfitting to the training distribution. This means that the representations learned for Africa during pretraining can be beneficial when labeled data is unavailable, especially for soil OC and pH. The DINOv3 models have likely also seen unlabeled data from Africa during pretraining, but they are still outperformed by ConvNeXtV2A-MM on 4 of the 5 tasks on the geographic split. ConvNeXtV2A-MM's strong performance on biomass and soil N suggests that the multimodal inputs are helping to resolve ambiguities that pretraining is not able to achieve. It is also possible that finetuning on the world without Africa erases some of the information the pretrained models had encoded about the region, hindering geographic generalization.
All the soil tasks and species occurrence offer an opportunity for globally pretrained models to improve their methodology to better facilitate geographic generalization on downstream tasks.

\begin{figure}[tb]
    \centering
    \includegraphics[width=\linewidth]{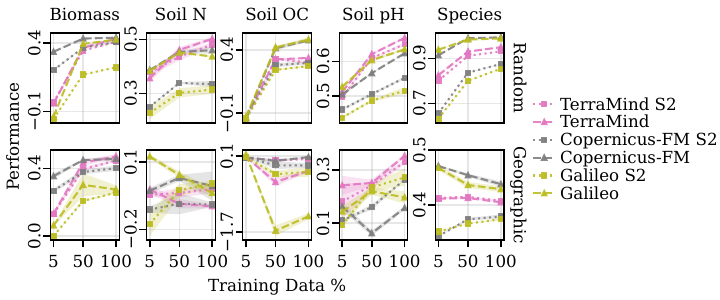}
    \caption{\textbf{Unimodal \vs multimodal input data.} Finetuning performance of S2-only (dotted-square) \vs multimodal (dashed-triangle) variants.}
    \label{fig:rq3_plot}
\end{figure}

\textbf{Unimodal \vs multimodal input data.}
We compare multimodal and S2-only variants of TerraMind, Copernicus-FM, and Galileo to evaluate the effect of multimodal input data on their performance. \cref{fig:rq3_plot} shows that the models can use the additional modalities to produce better task-specific representations on both test splits. While the multimodal version usually outperforms the unimodal one (less pronounced for TerraMind), there are exceptions in the geographic split of the soil tasks. While \cref{fig:rq2} showed that multimodal pretraining tends to lead to better geographic generalization than unimodal pretraining, \cref{fig:rq3_plot} suggests that \emph{finetuning} the same multimodally pretrained model with multi- rather than unimodal inputs can harm generalization. The additional modalities could lead to overfitting to the non-Africa training domain, which is harmful if the modalities undergo domain shifts between non-Africa and Africa. This effect appears more strongly for TerraMind and Galileo than Copernicus-FM, which could be because our Siamese Copernicus-FM simply averages embeddings across modalities, but the fusion is learned in the other two. This behavior has been observed in prior work~\cite{using_multiple_modalities}.
Furthermore, TerraMind and Galileo have modality-specific encoders and thus more capacity to fit the training data.

\begin{figure}[tb]
    \centering
    \includegraphics[width=\linewidth]{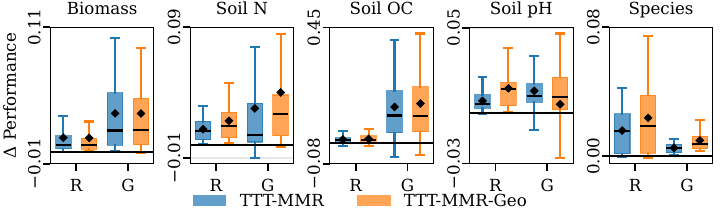}
    \caption{\textbf{Multimodal test-time training improvement per task.} Improvement of \textsc{\textsc{TTT-MMR}} (random batching) and \textsc{\textsc{TTT-MMR-Geo}} (geographic batching) over joint training. Boxplots show the distribution over all models and seeds. $\blacklozenge$=mean, $\mathbf{-}$=median, whiskers are based on the 1.5 IQR value. $\Delta$ reflects absolute change.}
    \label{fig:tta_results_per_task}
\end{figure}

\textbf{Multimodal test-time training.}
After performing JT, we apply \textsc{TTT-MMR} or \textsc{TTT-MMR-Geo} to the models to improve performance. \cref{fig:tta_results_per_task} shows that both of our proposed approaches improve performance over JT on all tasks. For biomass, soil OC, and soil pH, both variants perform similarly, with \textsc{TTT-MMR-Geo} displaying better performance on soil N and species. One-sided (greater) Wilcoxon tests with Holm-Bonferroni correction demonstrate statistical significance of the performance improvement after TTT for each boxplot with $p<0.05$. In addition, both methods improve performance on both test splits, with larger gains on the geographic split for biomass and soil OC.

The method rankings per model averaged over tasks shown in \cref{tab:ttt_ranks_by_model} demonstrate that both our TTT methods outperform JT for all models. \textsc{TTT-MMR-Geo} is the best method for all models except SatlasNet, Galileo, and ConvNeXtV2A\-/MM, indicating that these may benefit from the additional regularization that comes from \textsc{TTT-MMR}'s random batching. The RGB-only models tend to undergo larger performance improvements from TTT than the multispectral and multimodal models (shown in the Appendix). These are the models that otherwise have the least `information' about a given tile, but both \textsc{TTT-MMR} variants allow them to make use of multimodal data. The pretrained S2 and multimodal models still get additional modalities through \textsc{TTT-MMR} but also get more spectral information as input. Although MPMAE was pretrained by reconstructing these same modalities, we do not observe any special behavior in its performance gains from TTT. Overall, all benchmarked models benefit from our methods, despite having diverse architectures, pretraining strategies, and input modalities. This demonstrates that our TTT methods are model- and task-agnostic. Combining all 12 task modality reconstruction losses helps prevent overfitting to any single reconstruction task, producing a more robust adaptation signal that is more likely to align with the actual task of interest. Moreover, our per-modality gradient normalization improves stability during TTT.

\begin{table*}[t]
\centering
\caption{\textbf{Multimodal TTT improvement per model.} Average ranks of JT (joint training baseline), \textsc{TTT-MMR} (random batching), and \textsc{TTT-MMR-Geo} (geographic batching). Ranks are mean $\pm$ standard error averaged over tasks and seeds.}
\resizebox{\linewidth}{!}{%
\setlength{\tabcolsep}{5pt}
\centering
\begin{tabular}{@{}llcccccccccc@{}}
\toprule
\textbf{Test split} & \textbf{Method} & \makecell{\textbf{ConvNeXt} \\ \textbf{V2A}} & \makecell{\textbf{Scale-}\\\textbf{MAE}} & \makecell{\textbf{DINOv3} \\ \textbf{Web}} & \makecell{\textbf{DINOv3} \\ \textbf{Sat}} & \makecell{\textbf{Satlas}\\\textbf{Net}} & \textbf{MPMAE} & \makecell{\textbf{Terra}\\\textbf{Mind}} & \makecell{\textbf{Copernicus}\\\textbf{-FM}} & \textbf{Galileo} & \makecell{\textbf{ConvNeXt} \\ \textbf{V2A-MM}} \\
\midrule
\multirow{3}{*}{\textbf{Random}} & JT & $2.9 \pm 0.1$ & $3.0 \pm 0.0$ & $3.0 \pm 0.0$ & $3.0 \pm 0.0$ & $2.9 \pm 0.1$ & $3.0 \pm 0.0$ & $3.0 \pm 0.0$ & $2.8 \pm 0.1$ & $2.2 \pm 0.2$ & $2.5 \pm 0.2$ \\
 & TTT-MMR & $2.1 \pm 0.1$ & $1.8 \pm 0.1$ & $1.8 \pm 0.1$ & $1.7 \pm 0.1$ & $1.8 \pm 0.1$ & $1.6 \pm 0.1$ & $\mathbf{1.5 \pm 0.1}$ & $1.7 \pm 0.2$ & $\mathbf{1.7 \pm 0.1}$ & $\mathbf{1.7 \pm 0.2}$ \\
 & TTT-MMR-Geo & $\mathbf{1.1 \pm 0.1}$ & $\mathbf{1.2 \pm 0.1}$ & $\mathbf{1.2 \pm 0.1}$ & $\mathbf{1.3 \pm 0.1}$ & $\mathbf{1.3 \pm 0.1}$ & $\mathbf{1.4 \pm 0.1}$ & $\mathbf{1.5 \pm 0.1}$ & $\mathbf{1.5 \pm 0.1}$ & $2.1 \pm 0.2$ & $1.8 \pm 0.2$ \\
\midrule
\multirow{3}{*}{\textbf{Geographic}} & JT & $3.0 \pm 0.0$ & $2.8 \pm 0.1$ & $3.0 \pm 0.0$ & $3.0 \pm 0.0$ & $2.7 \pm 0.2$ & $2.7 \pm 0.2$ & $2.9 \pm 0.1$ & $2.8 \pm 0.1$ & $2.9 \pm 0.1$ & $2.5 \pm 0.2$ \\
 & TTT-MMR & $\mathbf{1.5 \pm 0.1}$ & $1.9 \pm 0.2$ & $1.7 \pm 0.1$ & $\mathbf{1.5 \pm 0.1}$ & $\mathbf{1.6 \pm 0.2}$ & $2.0 \pm 0.1$ & $1.7 \pm 0.1$ & $1.8 \pm 0.1$ & $1.6 \pm 0.1$ & $1.9 \pm 0.1$ \\
 & TTT-MMR-Geo & $\mathbf{1.5 \pm 0.1}$ & $\mathbf{1.3 \pm 0.1}$ & $\mathbf{1.3 \pm 0.1}$ & $\mathbf{1.5 \pm 0.1}$ & $1.7 \pm 0.2$ & $\mathbf{1.3 \pm 0.2}$ & $\mathbf{1.3 \pm 0.2}$ & $\mathbf{1.4 \pm 0.2}$ & $\mathbf{1.5 \pm 0.2}$ & $\mathbf{1.7 \pm 0.2}$ \\
\bottomrule
\end{tabular}
}
\label{tab:ttt_ranks_by_model}
\end{table*}

\definecolor{freq_gray}{HTML}{696969}

\begin{figure}[t]
    \centering
    \includegraphics[width=\linewidth]{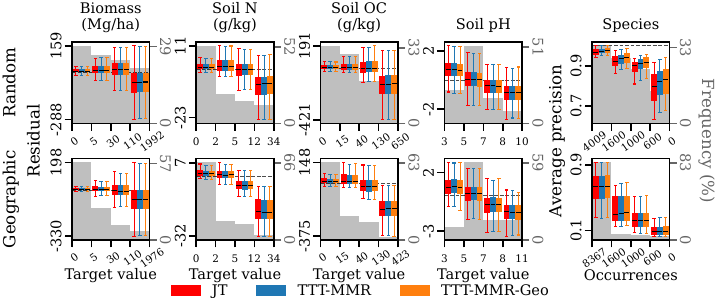}
    \caption{\textbf{Stratified multimodal test-time training performance.} Residuals are computed after pooling predictions across architectures. Average precision is computed per species and per architecture and then averaged across architectures, so the boxes show the distribution of those species-level means for the species within the bin. Frequency and species occurrence count are computed separately for each split. For the regression tasks, \textcolor{freq_gray}{frequency} is the percentage of test samples in each target bin. For species, \textcolor{freq_gray}{frequency} is the percentage of the summed occurrences of the bin's species relative to the total over all species. Boxplots show the median and whiskers based on the 1.5 IQR value. Dashed lines mark a residual of zero and an average precision of 1. Results reflect one seed.}
    \label{fig:ttt_long_tail}
\end{figure}
\begin{figure}
    \centering
    \includegraphics[width=\linewidth]{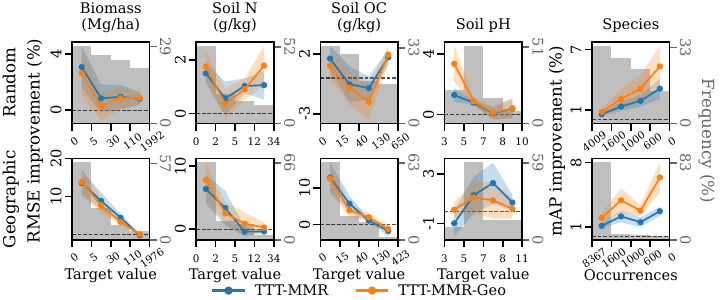}
    \caption{\textbf{Stratified multimodal test-time training relative improvement over joint training.} A positive value is an improvement. For the regression tasks, the improvement for each architecture and bin is $100\times(\text{RMSE}_{\text{JT}}-\text{RMSE}_{\text{TTT}})/\text{RMSE}_{\text{JT}}$. For species occurrence, the average precision is computed per architecture per species, and then the values are averaged across the species in the bin, yielding a mAP per architecture. The improvement is $100\times(\text{mAP}_{\text{TTT}}-\text{mAP}_{\text{JT}})/\text{mAP}_{\text{JT}}$. Lines are the mean across architectures and the bands are $\pm1$ standard error. Dashed lines mark 0. Results reflect one seed.}
    \label{fig:ttt-comparison-lines}
\end{figure}

\textbf{Test-time training performance stratified by frequency.}
We demonstrated in \cref{fig:tta_results_per_task} that TTT improves performance on all tasks and both splits. To investigate where the errors and performance gains come from, we stratify those results by target value for the regression tasks and by occurrence count for species. In \cref{fig:ttt_long_tail} we show an absolute stratified performance comparison among JT, \textsc{TTT-MMR}, and \textsc{TT-MMR-Geo}, and in \cref{fig:ttt-comparison-lines} we compare the relative performance gains from each method over JT. \cref{fig:ttt_long_tail} shows that for joint training and both TTT variants, errors concentrate where data are scarce. Biomass, soil N, and soil OC are strongly right-skewed with long tails, and models systematically underpredict high biomass, soil N, and soil OC values. Soil pH values cluster around 6-7, with fewer samples at the acidic and alkaline extremes, and models overpredict low pH values and underpredict high ones. Species average precision falls with rarity. \textsc{TTT-MMR} and \textsc{TTT-MMR-Geo} closely track JT across the regression bins, indicating that TTT does not substantially correct this error pattern. The clearest visible gain from TTT is for species on the random split, with \textsc{TTT-MMR-Geo} improving performance more than \textsc{TTT-MMR}. 
In \cref{fig:ttt-comparison-lines} we see that for biomass, soil N, and soil OC, on the random split RMSE relative improvement is greatest for common and rare target values and least in between. On the geographic split, the TTT benefit is greatest for common values and drops with increasing rarity, reaching approximately zero in the last bin. Overall, the relative gains are larger on the geographic split than on the random split, and both TTT methods provide similar relative performance improvements. For soil pH, on the random split improvement is greatest for low values, especially with \textsc{TTT-MMR-Geo}, and on the geographic split greatest for more common values. For species, the relative performance gains from TTT increase with rarity on both splits, and the benefit of geographic batching over random batching increases with rarity as well. As the relative improvement from TTT varies across bins, tasks, and splits, future work could explore when and how strongly to adapt to unlabeled samples at test time.

\section{Conclusion}
\label{sec:conclusion}

Models pretrained on large, unlabeled Earth observation datasets aim to generalize to new tasks and geographic domains with limited data.
To evaluate their progress, we present \textsc{MMEarth-Bench}, a new multimodal benchmark dataset that contains 12 aligned EO modalities and task data for biomass, soil nitrogen, soil organic carbon, soil pH, and species occurrence. 
We benchmark 8 pretrained (RGB, multispectral, and multimodal) models on these tasks and find that all models exhibit a geographic generalization gap on 4 out of 5 tasks.
Self-supervised multimodal pretraining improves downstream performance in low-shot regimes and under geographic shifts, but with sufficient labels a simple multimodal model trained from scratch is competitive. To enable \emph{any} (pretrained) model to employ all available modalities when making predictions, we propose self-supervised multimodal \emph{test-time} training. 
Our proposed \textsc{TTT-MMR} method uses all modalities as reconstruction tasks at test time to provide a self-supervised adaptation signal. We show that \textsc{TTT-MMR} can improve performance on all models, tasks, and splits over a joint training baseline, with geographic batching outperforming random batching for most models on both splits. Thus, we have formulated an efficient, model-agnostic method for improving adaptation performance at test time.

\subsection*{Acknowledgments}
We appreciate the open data policies of the Copernicus program and its partners ESA and ECMWF. We thank Harvard's FAS Research Computing cluster and also Google Earth Engine for free access to the data. This work was supported by the Pioneer Centre for AI (DNRF grant no. P1), research grant Global Wetland Center (grant no. NNF23OC0081089) from the Novo Nordisk Foundation, and the European Union project ELIAS (grant agreement no. 101120237). LG was supported by the National Science Foundation Graduate Research Fellowship (grant no. DGE 2140743) and VILLUM FONDEN (grant no. VIL70006).
\newpage
%
%

\bibliographystyle{splncs04}
\bibliography{main}
\clearpage

 




\providecommand{\authcount}[1]{}
\lstset{
  language=Python,
  basicstyle=\ttfamily\small,
  keywordstyle=\color{blue},
  commentstyle=\color{gray},
  stringstyle=\color{teal},
  frame=single,
  breaklines=true,
  showstringspaces=false,
  tabsize=2
}
\raggedbottom


%




\title{MMEarth-Bench: Global Model Adaptation via Multimodal Test-Time Training\\\textsc{Appendix}} 

\titlerunning{MMEarth-Bench Appendix}

\author{Lucia Gordon\inst{1,2}\orcidlink{0000-0003-3219-6960} \and
Serge Belongie\inst{2}\orcidlink{0000-0002-0388-5217} \and
Christian Igel\inst{2}\orcidlink{0000-0003-2868-0856}\and Nico Lang\inst{2}\orcidlink{0000-0001-8434-027X}}

\authorrunning{L.~Gordon et al.}

\institute{Harvard University, USA \and
University of Copenhagen, Denmark
\\
\email{luciagordon@g.harvard.edu, nila@di.ku.dk}}

\maketitle
\setcounter{section}{0}
\setcounter{figure}{0}
\setcounter{table}{0}

\makeatletter
\setcounter{tocdepth}{2}
\startcontents
\printcontents{}{1}{}
\setcounter{tocdepth}{0}
\makeatother

\renewcommand{\thesection}{A.\arabic{section}}
\renewcommand{\theHsection}{A.\thesection}
\renewcommand{\thefigure}{A.\arabic{figure}}
\renewcommand{\thetable}{A.\arabic{table}}

\setlength{\cftsecnumwidth}{2.3em}
\setlength{\cftsubsecnumwidth}{3em}

\section{MMEarth-Bench dataset}
\subsection{Overview and format}
\begin{figure*}[h]
    \centering
    \begin{subfigure}{0.19\linewidth}
        \centering
        \includegraphics[width=\linewidth]{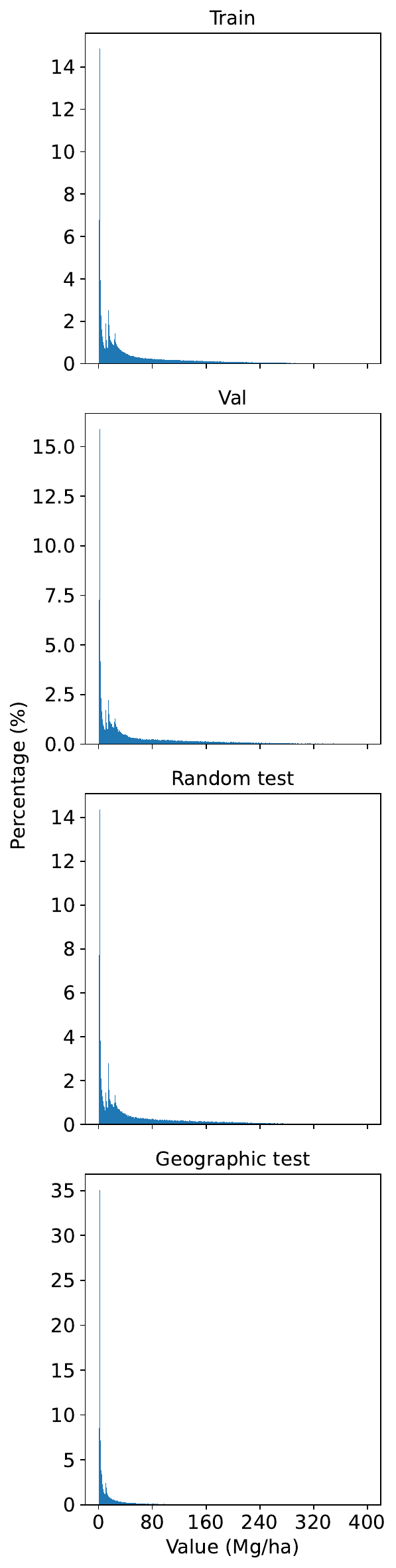}
        \caption{Biomass}
        \label{fig:biomass_distributions}
    \end{subfigure}
    \begin{subfigure}{0.19\linewidth}
        \centering
        \includegraphics[width=\linewidth]{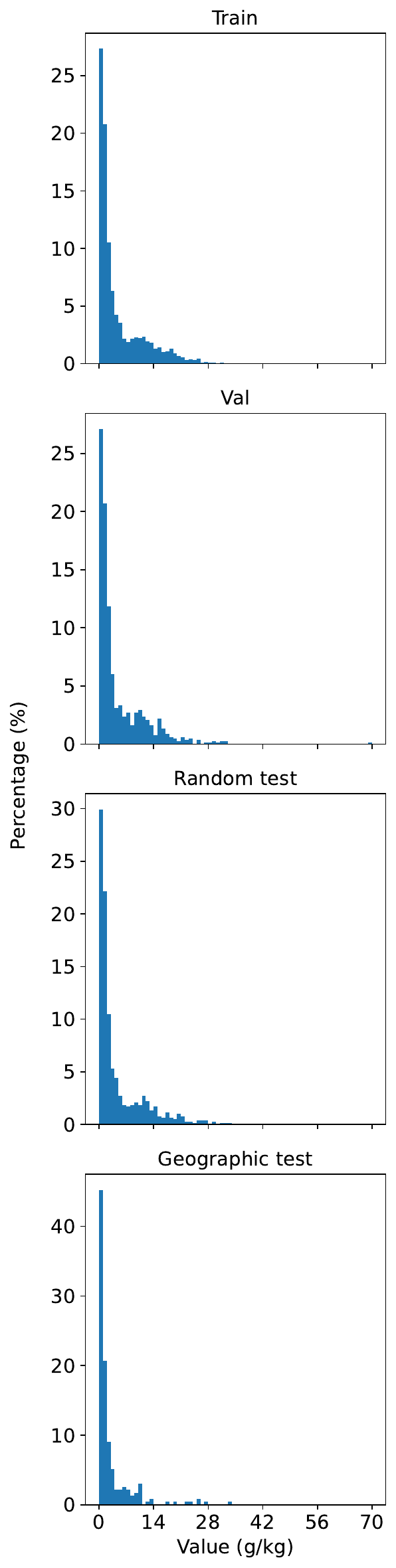}
        \caption{Soil N}
        \label{fig:soil_nitrogen_distributions}
    \end{subfigure}
    \begin{subfigure}{0.19\linewidth}
        \centering
        \includegraphics[width=\linewidth]{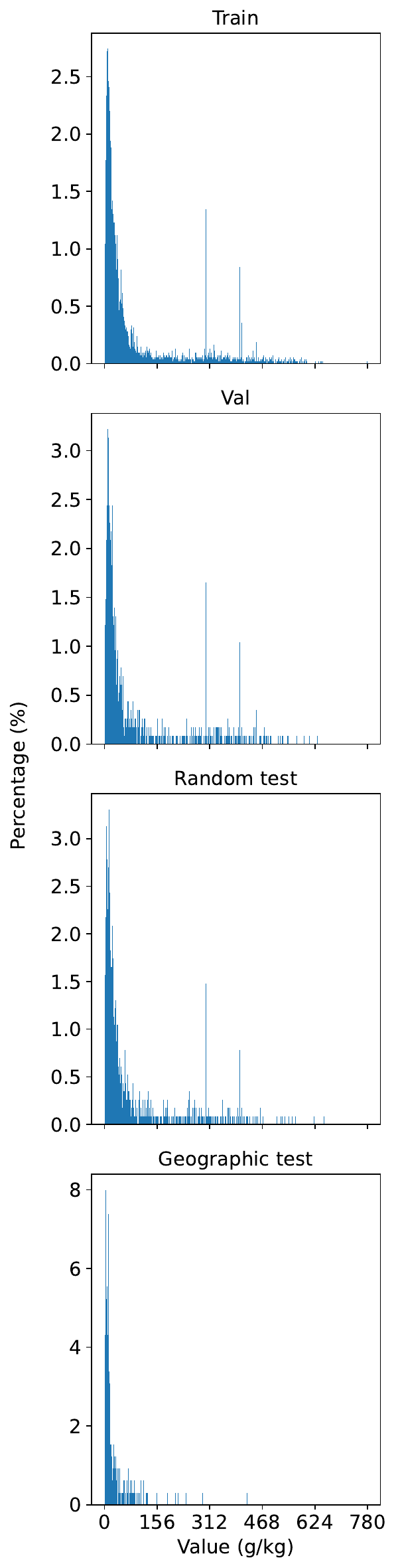}
        \caption{Soil OC}
        \label{fig:soil_organic_carbon_distributions}
    \end{subfigure}
    \begin{subfigure}{0.19\linewidth}
        \centering
        \includegraphics[width=\linewidth]{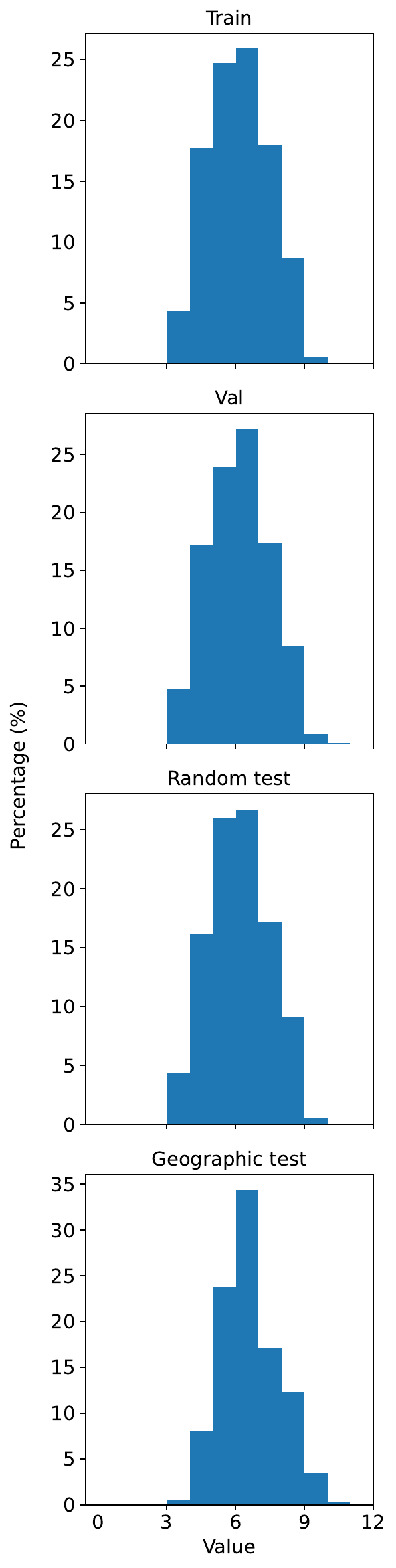}
        \caption{Soil pH}
        \label{fig:soil_pH_distributions}
    \end{subfigure}
    \begin{subfigure}{0.19\linewidth}
        \centering
        \includegraphics[width=\linewidth]{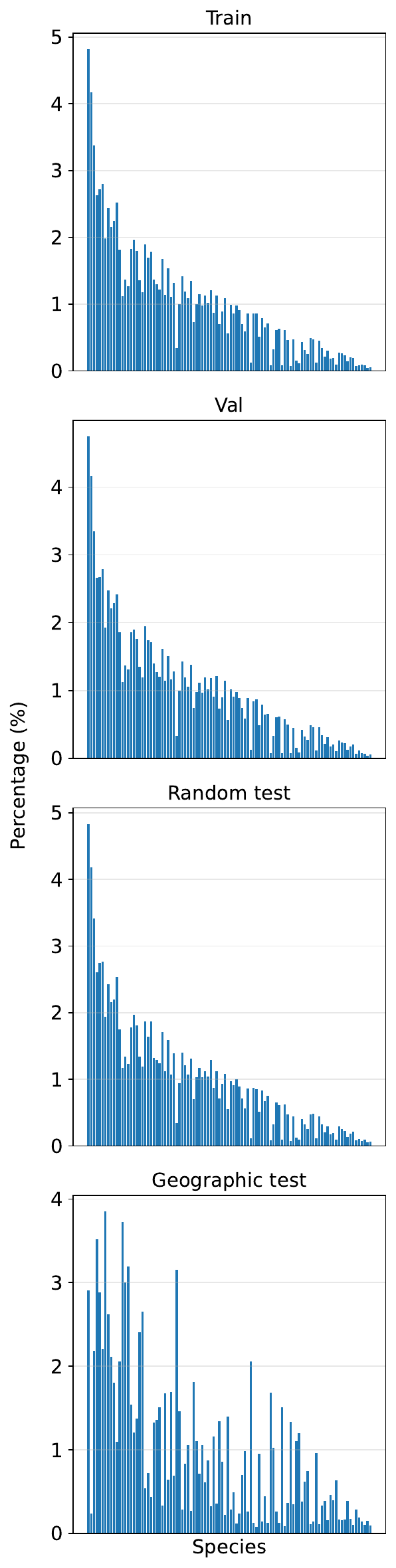}
        \caption{Species}
        \label{fig:species_distributions}
    \end{subfigure}
    \caption{\textbf{Distributions across splits for MMEarth-Bench tasks.} The species are ordered as in \cref{fig:tiles_per_species}.}
    \label{fig:distributions}
\end{figure*}

We show the number of tiles in each split for all the MMEarth-Bench tasks in \cref{tab:tiles_per_split}. We provide information on the modalities in MMEarth-Bench, now including the no-data values, in \cref{tab:modalities_full}. We also display summary statistics for the modalities in \cref{tab:modality_stats}. Summary statistics for the four regression tasks can be found in \cref{tab:regression_split_stats} and \cref{tab:regression_split_ranges}. \cref{fig:distributions} visualizes the data distributions by task for all splits. The data for each task dataset is stored in an H5 file, with sizes shown in \cref{tab:dataset_h5_sizes}. Users can read in data from one of our task H5 files using the code in \cref{fig:reading_data_code}. The split data for each task containing the tile indices for the various splits along with the normalization statistics is stored in JSON files. The data is hosted on our \href{https://sid.erda.dk/cgi-sid/ls.py?share_id=cbMhbwV1yP&current_dir=.&flags=f}{university server}.

\begin{table}[h]
\centering
\caption{\textbf{Number of tiles per split by task.}}
\label{tab:tiles_per_split}
\begin{tabular}{lccccc}
\toprule
\textbf{Split} & \textbf{Biomass} & \textbf{Soil N} & \textbf{Soil OC} & \textbf{Soil pH} & \textbf{Species} \\
\midrule
Train 100\% & 10665 & 3787 & 5359 & 5711 & 18882 \\
Train 50\% & 5332 & 1894 & 2680 & 2856 & 9441 \\
Train 5\% & 533 & 189 & 268 & 286 & 944 \\
Validation & 2286 & 812 & 1149 & 1224 & 4046 \\
Random test & 2286 & 812 & 1149 & 1224 & 4047 \\
Geographic test & 3156 & 232 & 325 & 349 & 9435 \\
\bottomrule
\end{tabular}
\end{table}

\begin{table*}[h]
\centering
\caption{\textbf{MMEarth-Bench modalities.} Including the number of bands and no-data value for each modality.}
\label{tab:modalities_full}
\setlength{\tabcolsep}{5pt}
\resizebox{\linewidth}{!}{%
\rowcolors{2}{gray!15}{white}
\begin{tabular}{@{} 
    l   
    r   
    >{\raggedright\arraybackslash}p{0.37\linewidth}   
    c   
    c   
    r   
@{}}
\toprule
\textbf{Modality} & 
\textbf{\# Bands} & 
\textbf{Band names} & 
\textbf{Scale} & 
\textbf{Type} &
\textbf{No-data value} \\
\midrule
Sentinel-2               & 12 & \makecell[l]{B1, B2, B3, B4, B5, B6, B7, \\B8, B8A, B9, B11, B12}  
                         & Pixel-level & Continuous & 65535 \\

Sentinel-1               &  8 & \makecell[l]{Ascending VV, VH, HH, HV;\\Descending VV, VH, HH, HV}  
                         & Pixel-level & Continuous & -9999 \\

ASTER GDEM                 &  2 & Elevation, slope  
                         & Pixel-level & Continuous & -9999 \\

ETH Global Canopy Height &  2 & Height, uncertainty  
                         & Pixel-level & Continuous & 255 \\

Dynamic World            &  1 & Landcover  
                         & Pixel-level & Categorical & 9 \\

ESA WorldCover           &  1 & Landcover  
                         & Pixel-level & Categorical & 11 \\

Precipitation            &  3 & Last month, month, year  
                         & Tile-level  & Continuous & -9999 \\

Temperature              &  9 & \makecell[l]{Last month max, mean, min; \\Month max, mean, min; \\Year max, mean, min}  
                         & Tile-level  & Continuous & -9999 \\

Geolocation              &  2 & Longitude, latitude  
                         & Tile-level  & Continuous & N/A \\

Sentinel-2 date                    &  1 & Date  
                         & Tile-level  & Continuous & N/A \\

Biome                    &  1 & Biome number  
                         & Tile-level  & Categorical & 14 \\

Ecoregion                &  1 & Ecoregion number  
                         & Tile-level  & Categorical & 846 \\
\bottomrule
\end{tabular}}
\end{table*}

\begin{table*}[h]
\centering
\caption{\textbf{Modality statistics.} NaN percentage and [min, max] range across all tasks.}
\label{tab:modality_stats}
\resizebox{\linewidth}{!}{%
\begin{tabular}{lcccccccccc}
\toprule
\multirow{2}{*}{\textbf{Modality}} & \multicolumn{2}{c}{\textbf{Biomass}} & \multicolumn{2}{c}{\textbf{Soil Nitrogen}} & \multicolumn{2}{c}{\textbf{Soil Organic Carbon}} & \multicolumn{2}{c}{\textbf{Soil pH}} & \multicolumn{2}{c}{\textbf{Species}} \\
\cmidrule(lr){2-3}\cmidrule(lr){4-5}\cmidrule(lr){6-7}\cmidrule(lr){8-9}\cmidrule(lr){10-11}
 & \textbf{NaN \%} & \textbf{[Min, Max]} & \textbf{NaN \%} & \textbf{[Min, Max]} & \textbf{NaN \%} & \textbf{[Min, Max]} & \textbf{NaN \%} & \textbf{[Min, Max]} & \textbf{NaN \%} & \textbf{[Min, Max]} \\
\midrule
Sentinel-2 & 1.79e-03\% & [0.00, 2.07e+04] & 9.84e-04\% & [0.00, 1.81e+04] & 1.63e-03\% & [0.00, 2.03e+04] & 9.28e-04\% & [0.00, 2.03e+04] & 7.88e-04\% & [0.00, 2.27e+04] \\
Sentinel-1 & 6.08e+01\% & [-70.29, 31.73] & 6.12e+01\% & [-64.31, 32.83] & 6.19e+01\% & [-64.31, 37.43] & 6.11e+01\% & [-64.31, 37.43] & 5.68e+01\% & [-75.43, 29.50] \\
ASTER GDEM elevation & 0.0\% & [-186.35, 6117.91] & 0.0\% & [-32.49, 5557.40] & 0.0\% & [-32.49, 5557.40] & 0.0\% & [-31.00, 5557.40] & 0.0\% & [-427.00, 6649.21] \\
ASTER GDEM slope & 0.0\% & [0.00, 77.56] & 0.0\% & [0.00, 78.61] & 0.0\% & [0.00, 77.77] & 0.0\% & [0.00, 78.61] & 0.0\% & [0.00, 79.94] \\
ETH GCH height & 1.48\% & [0.00, 59.96] & 1.70\% & [0.00, 64.00] & 2.33\% & [0.00, 60.00] & 1.85\% & [0.00, 64.00] & 2.74\% & [0.00, 68.34] \\
ETH GCH uncertainty & 1.48\% & [0.00, 246.00] & 1.70\% & [0.00, 252.00] & 2.33\% & [0.00, 252.00] & 1.85\% & [0.00, 252.00] & 2.74\% & [0.00, 253.00] \\
DynamicWorld & 1.03\% & [0.00, 8.00] & 4.07e-02\% & [0.00, 8.00] & 4.17e-02\% & [0.00, 8.00] & 3.19e-02\% & [0.00, 8.00] & 3.02e-01\% & [0.00, 8.00] \\
ESA WorldCover & 0.0\% & [0.00, 10.00] & 0.0\% & [0.00, 10.00] & 2.51e-02\% & [0.00, 10.00] & 2.35e-02\% & [0.00, 10.00] & 0.0\% & [0.00, 10.00] \\
Precipitation & 1.54\% & [1.80e-05, 12.07] & 1.22\% & [1.80e-05, 7.33] & 1.01\% & [1.80e-05, 5.60] & 7.17e-01\% & [1.80e-05, 7.33] & 9.23e-01\% & [1.80e-05, 10.91] \\
Temperature & 1.54\% & [227.58, 324.67] & 1.22\% & [220.34, 320.92] & 1.01\% & [220.34, 320.92] & 7.17e-01\% & [220.34, 320.92] & 9.23e-01\% & [218.25, 326.06] \\
Geolocation & 0.0\% & [-179.97, 179.60] & 0.0\% & [-160.03, 175.45] & 0.0\% & [-161.96, 175.45] & 0.0\% & [-160.03, 172.73] & 0.0\% & [-179.89, 179.81] \\
Geolocation encoding & 0.0\% & [-1.00, 1.00] & 0.0\% & [-1.00, 1.00] & 0.0\% & [-1.00, 1.00] & 0.0\% & [-1.00, 1.00] & 0.0\% & [-1.00, 1.00] \\
Month encoding & 0.0\% & [-1.00, 1.00] & 0.0\% & [-1.00, 1.00] & 0.0\% & [-1.00, 1.00] & 0.0\% & [-1.00, 1.00] & 0.0\% & [-1.00, 1.00] \\
Biome & 0.0\% & [0.00, 13.00] & 4.08e-01\% & [0.00, 12.00] & 3.01e-01\% & [0.00, 13.00] & 1.18e-01\% & [0.00, 13.00] & 8.18e-01\% & [0.00, 13.00] \\
Ecoregion & 0.0\% & [3.00, 845.00] & 4.08e-01\% & [2.00, 828.00] & 3.01e-01\% & [2.00, 844.00] & 1.18e-01\% & [2.00, 844.00] & 8.18e-01\% & [2.00, 843.00] \\
MSK CLDPRB & 0.0\% & [0.00, 100.00] & 0.0\% & [0.00, 100.00] & 0.0\% & [0.00, 100.00] & 0.0\% & [0.00, 100.00] & 0.0\% & [0.00, 100.00] \\
S2CLOUDLESS & 0.0\% & [0.00, 100.00] & 0.0\% & [0.00, 100.00] & 0.0\% & [0.00, 100.00] & 0.0\% & [0.00, 100.00] & 0.0\% & [0.00, 100.00] \\
SCL & 1.47e-04\% & [2.00, 11.00] & 4.11e-04\% & [2.00, 11.00] & 3.92e-04\% & [2.00, 11.00] & 1.46e-04\% & [2.00, 11.00] & 3.14e-04\% & [2.00, 11.00] \\
MSK CLDPRB CLOUDY PIXEL FRACTION & 0.0\% & [0.00, 0.18] & 0.0\% & [0.00, 0.16] & 0.0\% & [0.00, 0.17] & 0.0\% & [0.00, 0.17] & 0.0\% & [0.00, 0.23] \\
S2CLOUDLESS CLOUDY PIXEL FRACTION & 0.0\% & [0.00, 0.19] & 0.0\% & [0.00, 0.17] & 0.0\% & [0.00, 0.17] & 0.0\% & [0.00, 0.17] & 0.0\% & [0.00, 0.17] \\
SCL NO DATA PIXEL FRACTION & 0.0\% & [0.00, 0.02] & 0.0\% & [0.00, 0.02] & 0.0\% & [0.00, 0.02] & 0.0\% & [0.00, 0.01] & 0.0\% & [0.00, 0.04] \\
\bottomrule
\end{tabular}
}
\end{table*}

\begin{table}[h]
\centering
\caption{\textbf{Split statistics for regression tasks.} Values are mean $\pm$ standard deviation. Units are Mg/ha for biomass and g/kg for soil N and soil OC.}
\label{tab:regression_split_stats}
\setlength{\tabcolsep}{5pt}
\begin{tabular}{lcccc}
\toprule
\textbf{Split} & \textbf{Biomass} & \textbf{Soil N} & \textbf{Soil OC} & \textbf{Soil pH} \\
\midrule
Train 100\% & $62.51 \pm 99.24$ & $4.98 \pm 6.08$ & $81.21 \pm 125.10$ & $6.10 \pm 1.32$ \\
Train 50\% & $63.86 \pm 102.75$ & $5.01 \pm 6.11$ & $81.77 \pm 126.70$ & $6.10 \pm 1.33$ \\
Train 5\% & $61.09 \pm 101.79$ & $4.47 \pm 5.99$ & $75.40 \pm 128.10$ & $6.18 \pm 1.35$ \\
Validation & $63.80 \pm 101.03$ & $4.97 \pm 6.42$ & $81.67 \pm 123.34$ & $6.08 \pm 1.32$ \\
Random test & $65.14 \pm 99.81$ & $4.66 \pm 6.08$ & $74.72 \pm 111.76$ & $6.12 \pm 1.30$ \\
Geographic test & $26.09 \pm 66.16$ & $3.03 \pm 4.97$ & $25.38 \pm 43.25$ & $6.59 \pm 1.24$ \\
\bottomrule
\end{tabular}
\end{table}

\begin{table}[h]
\centering
\caption{\textbf{Split ranges for regression tasks.} Values are [min, max]. Units are Mg/ha for biomass and g/kg for soil N and soil OC.}
\label{tab:regression_split_ranges}
\setlength{\tabcolsep}{5pt}
\begin{tabular}{lcccc}
\toprule
\textbf{Split} & \textbf{Biomass} & \textbf{Soil N} & \textbf{Soil OC} & \textbf{Soil pH} \\
\midrule
Train 100\% & [0.00, 1991.21] & [0.00, 38.81] & [0.00, 779.00] & [3.00, 10.70] \\
Train 50\% & [0.00, 1991.21] & [0.00, 38.80] & [0.00, 779.00] & [3.00, 10.70] \\
Train 5\% & [0.00, 1794.24] & [0.04, 38.80] & [0.00, 779.00] & [3.40, 9.50] \\
Validation & [0.00, 1981.19] & [0.00, 69.07] & [0.00, 630.30] & [3.20, 10.00] \\
Random test & [0.00, 1991.65] & [0.00, 33.86] & [0.00, 650.00] & [3.00, 9.90] \\
Geographic test & [0.67, 1975.70] & [0.07, 33.70] & [0.30, 422.80] & [3.60, 10.30] \\
\bottomrule
\end{tabular}
\end{table}

\begin{table}[h]
\centering
\caption{\textbf{Data volume.} One H5 file per task. Size recorded in gigabytes (G).}
\begin{tabular}{ccccc}
\toprule
\textbf{Biomass} &
\textbf{Soil N} &
\textbf{Soil OC} &
\textbf{Soil pH} &
\textbf{Species} \\
\midrule
16.0G & 4.3G & 6.0G & 6.4G & 28.0G \\
\bottomrule
\end{tabular}
\label{tab:dataset_h5_sizes}
\end{table}

\begin{figure*}[h]
\begin{lstlisting}
import h5py
import json

with h5py.File(PATH_TO_H5, 'r') as h5_file:
    if KEY in ['sentinel2_date', 'crs', 'sentinel2_system_index']:
        key_data = h5_file[KEY].asstr()[...]
    elif KEY in ['missing_modalities', 'species']:
        key_data = [json.loads(lst) for lst in h5_file[KEY].asstr()[...]]
    elif KEY in ['Sentinel2', 'Sentinel1', 'ASTER_GDEM', 'ETH_GCH', 'DynamicWorld', \
                'ESA_WorldCover', 'precipitation', 'temperature', 'geolocation_encoding', \
                'month_encoding', 'biome', 'ecoregion', 'biomass', 'soil_nitrogen', \
                'soil_organic_carbon', 'soil_pH', 'MSK_CLDPRB', 'S2CLOUDLESS', 'SCL', \
                'geolocation', 'MSK_CLDPRB_CLOUDY_PIXEL_FRACTION', \ 
                'S2CLOUDLESS_CLOUDY_PIXEL_FRACTION', 'SCL_NO_DATA_PIXEL_FRACTION', 'id', \
                'transform']:
        key_data = h5_file[KEY][:]
\end{lstlisting}
\caption{\textbf{Python code for reading data from H5 files in MMEarth-Bench.} Both modalities and tasks can be provided as keys.}
\label{fig:reading_data_code}
\end{figure*}

\FloatBarrier
\subsection{Biomass}
Our biomass dataset is sourced from above-ground biomass measurements collected by NASA's GEDI mission. To guide our sampling, we divide the planet into its 14 terrestrial biomes, which can be further divided into 846 total ecoregions. The geographic extents of these ecoregions are captured in Google Earth Engine's \href{https://developers.google.com/earth-engine/datasets/catalog/RESOLVE_ECOREGIONS_2017#description}{RESOLVE Ecoregions 2017} dataset~\cite{ecoregions}. The GEDI mission only collects data within the latitude band spanning from 51.6$^\circ$ S to 51.6 N, so we do not consider ecoregions outside of this range, leaving 774 remaining. For ecoregions overlapping the boundary of the GEDI range, we crop out the area that does not intersect with the range. We aim to generate 20,000 total biomass tiles. We want the tiles to be distributed approximately evenly across biomes, so we set a target number of tiles per biome given by 1,429, which is the rounded-up quotient of 20,000 tiles divided by 14 biomes. Next, we compute the area of each ecoregion (cropped to the GEDI range) and then sum the areas of the ecoregions within each biome to get the total area of each biome. We then calculate the number of tiles to be generated in each ecoregion by scaling the number of tiles per biome by the fraction of the area the ecoregion covers within its biome and round up to ensure at least 1 tile is generated per ecoregion. This results in 20,421 total tiles. We save this data to a JSON file \texttt{ecoregion\_tile\_counts.json}.

Now we can begin fetching GEDI data from Google Earth Engine. We loop through the biomes and consider the ecoregions one-by-one. We check the number of running SLURM jobs every second and wait to proceed with running the ecoregion until there are maximum 30 jobs running to avoid hitting memory limits on Google Earth Engine. We then submit a 3-day SLURM job with 500MB of RAM and pass in the number of tiles we pre-calculated for the ecoregion.

The process to generate tiles for a given ecoregion is as follows. GEDI L4A is the dataset containing above-ground biomass measurements. We first use Google Earth Engine's \href{https://developers.google.com/earth-engine/datasets/catalog/LARSE_GEDI_GEDI04_A_002_INDEX#table-schema}{GEDI L4A table index} dataset~\cite{agb}, which contains a list of the GEDI L4A collection IDs along with their start and end times. We only consider data from the year 2020, so we filter the list to include only those collections that have data in 2020. We then extract the geographic range of the ecoregion cropped to the GEDI range as discussed earlier. We use this ecoregion range to filter the table index dataset to yield only those collections that have data for the desired ecoregion in the year 2020. We then create a list of their collection IDs and query the collections that are valid assets. We merge all these collections into a single collection. We then filter the collection by the bounds of the ecoregion to exclude GEDI points outside of the ecoregion. We then apply quality filters to only keep points that have ``degrade\_flag = 0,'' ``l2\_quality\_flag = 1,'' ``l4\_quality\_flag = 1,'' ``leaf\_off\_flag = 0'' (growing season) and ``region\_class $>$ 0'' (land). We then apply a filter to only keep points whose leaf-off day is after their leaf-on day. This is to make it more straight-forward to generate a leaf-on date filter later on when downloading the data modalities. Finally, we filter by GEDI points with biomass value $\leq$2,000 since the plots in Duncanson et al.~\cite{biomass_models} only show AGBD values up to $\approx$ 2,000.

Next, we randomly shuffle the GEDI points and select however many points are needed from the beginning of the list, possibly fewer depending on how many there are. We then export the collection of points to a bucket in Google Cloud Storage in GeoJSON format. Once it has been uploaded, we download the file. If there is at least one point in the collection, we generate an outer tile around each point of size 1,300x1,300-m. The outer tile is larger than our desired tile by one pixel on each side. We do this because later on we construct the tile on the Sentinel-2 grid, and because in that grid a single pixel spans 10m across, our resultant tile could be shifted from our original intended tile by a maximum of 1 pixel or 10m in each direction. Hence we make an ``outer tile`` that is guaranteed to contain our final tile for the purpose of checking overlap between tiles. We then remove points with overlapping tiles and shuffle those remaining. We then select the number of points we need from the start of the list, possibly fewer depending on if we lost some points due to overlap removal. If the number of points is equal to the number of desired ecoregion tiles or the number of points before removing overlaps was less than the number of ecoregion tiles, we save the points' data as a GeoJSON. Otherwise, we select twice however many points are needed from the shuffled list of GEDI points and repeat the above process. If there are still not enough points, then we select three times however many points are needed and continue this loop until there are enough points left over after removing overlaps. We then merge all of the ecoregion points into a single list, remove any overlaps among points from all the ecoregions, and then save the remaining points as a single GeoJSON file containing 19,834 points. Note that this is 587 points fewer than our intended 20,421. This is because 14 ecoregions had 0 GEDI points when 1 was desired and one ecoregion had only 38 GEDI points when 578 were desired. Finally, we concatenate all the points from all the ecoregions. The biomass point generation process took $\approx$ 50 hours. The points list is then shuffled and each point is assigned an ID. \Cref{fig:biomass_missing_modality_counts} shows how many points are missing each of the modalities after generating tiles. \Cref{fig:biomass_distributions} visualizes the distribution in biomass values across the train, validation, random test, and geographic test sets.

\subsection{Soil Nitrogen, Organic Carbon, \& pH}
Our soil datasets are sourced from the \href{https://data.isric.org/geonetwork/srv/api/records/e50f84e1-aa5b-49cb-bd6b-cd581232a2ec}{WoSIS December 2023 snapshot}~\cite{wosis}. We consider the datasets for nitrogen, organic carbon, and pH. \href{https://soil.copernicus.org/articles/7/217/2021/}{SoilGrids}, a model that predicts soil properties based on environmental covariates, exhibited a decrease in performance with increasing soil depth due to a weaker relationship between environmental covariates and soil properties in the deeper layers~\cite{soilgrids}. Thus, we filter the three datasets by measurements within the depth range 0-5 cm, which is the shallowest soil layer of the six standard layers defined in the \href{https://www.sciencedirect.com/science/article/abs/pii/B9780128001370000030?via%3Dihub}{GlobalSoilMap} specifications~\cite{globalsoilmap}. We then filter measurements by those whose positional uncertainty is ``circa 100 m,'' which is the strictest option. We then save the remaining points along with their outer tiles and ``value\_avg''. This results in 8,991 soil nitrogen points, 11,593 soil organic carbon points, and 12,283 soil pH points. For each task, we shuffle the points, remove overlaps among the outer tiles as we did for biomass, assign an ID to each point, and then save the data to a GeoJSON. This yields 5,884 points for soil nitrogen, 8,277 points for soil organic carbon, and 8,836 points for soil pH.

\Cref{fig:soil_nitrogen_missing_modality_counts,fig:soil_organic_carbon_missing_modality_counts,fig:soil_pH_missing_modality_counts} show how many points are missing each of the modalities for the three soil tasks after generating tiles. \Cref{fig:soil_nitrogen_distributions,fig:soil_organic_carbon_distributions,fig:soil_pH_distributions} visualize the distribution in soil nitrogen, organic carbon, and pH values, respectively, across the train, validation, random test, and geographic test sets.

\subsection{Species}
Our species data is sourced from the terrestrial mammals polygon shapefile, which is one of the data products on the \href{https://www.iucnredlist.org/resources/spatial-data-download}{IUCN Red List's Spatial Data Download} page~\cite{iucn}. The data we use comes from Version 2025-1, last updated on March 27, 2025. We downloaded the data on June 13, 2025. We later use Africa as our geographic held-out test region, so we need to ensure that our selected species occur both in and out of Africa. We use the QGIS software to draw a Polygon representing Africa for the purpose of our geographic split. We base our boundary on the \href{https://www.usgs.gov/programs/earthquake-hazards/google-earthtmkml-files}{Tectonic Plate Boundaries} KMZ file provided by the USGS~\cite{tectonic_plate_boundaries} as well as Esri's \href{https://hub.arcgis.com/datasets/esri::world-continents/about}{World Continents} GeoJSON available on ArcGIS Hub~\cite{world_continents}. We export our Africa polygon as a GeoJSON.

\begin{figure}
    \centering
    \includegraphics[width=0.6\linewidth]{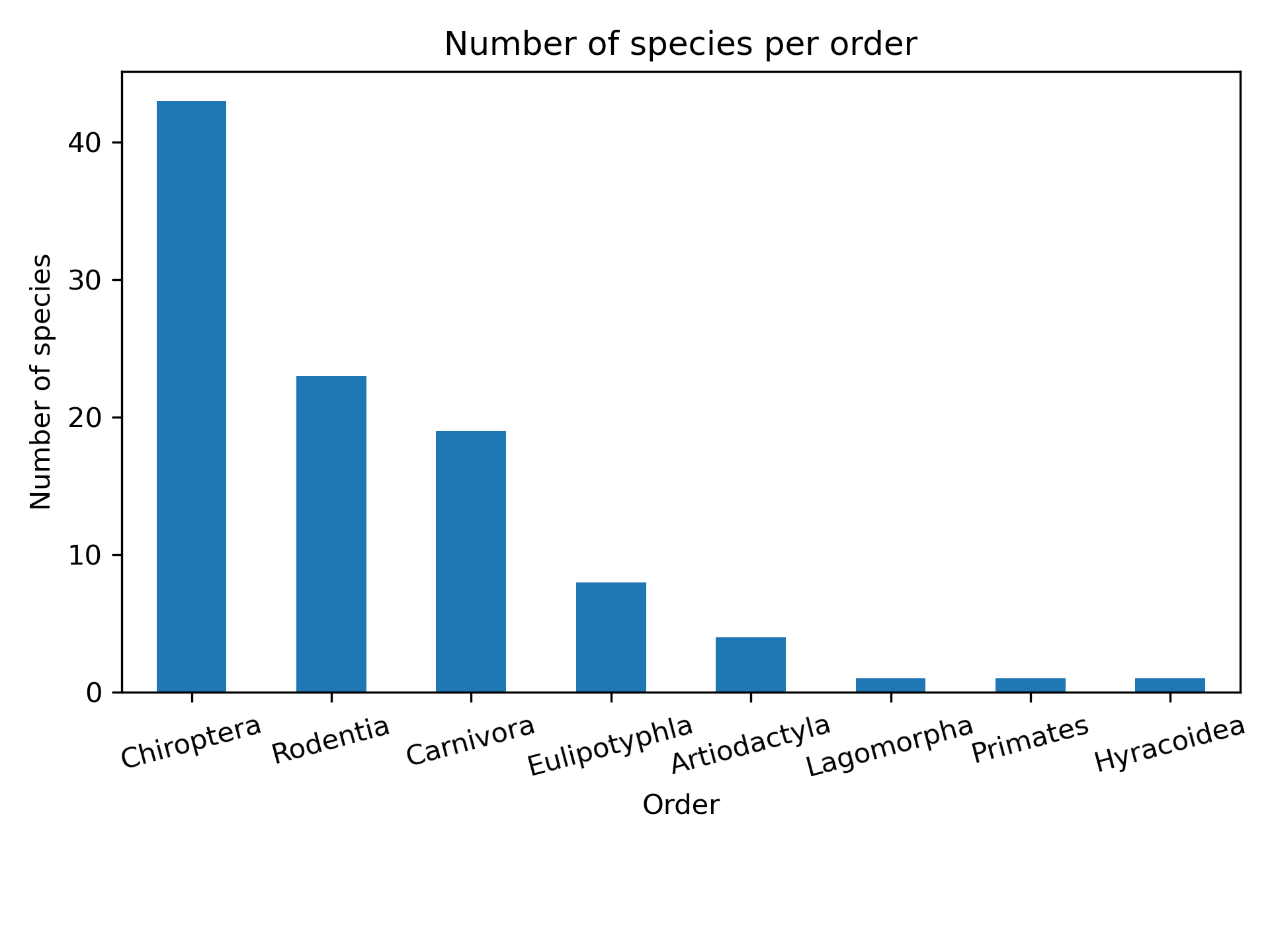}
    \caption{\textbf{Species distribution per taxonomic order.}}
    \label{fig:species_per_order}
\end{figure}

The terrestrial mammals shapefile has a row for each disjoint part of a species' range. We dissolve the GeoDataFrame by species name so that each row corresponds to a unique species and the multiple Polygons per species are combined into a single MultiPolygon. This results in 5,675 rows. We then filter the GeoDataFrame by species whose range intersects with Africa, leaving 1,451 species. We then reproject the species GeoDataFrame and the Africa geometry to EPSG:6933, the global equal area projection. In this projection, we compute for each species how much of its range is in and outside Africa in square kilometers. There are 122 species with nonzero area both in and outside Africa. We then filter by species whose range covers at least 6,000 km$^2$ both in and outside Africa, leaving 101 species. We take the first 100 in the list. Of these final 100 selected species, the minimum area in Africa is 12,018.38 km$^2$ and the minimum area outside Africa is 6,218.04 km$^2$. We save the ranges of these 100 species to a GeoJSON. \Cref{fig:species_per_order} shows the taxonomic orders of the selected species.

\begin{figure*}
    \centering
    \includegraphics[width=\linewidth]{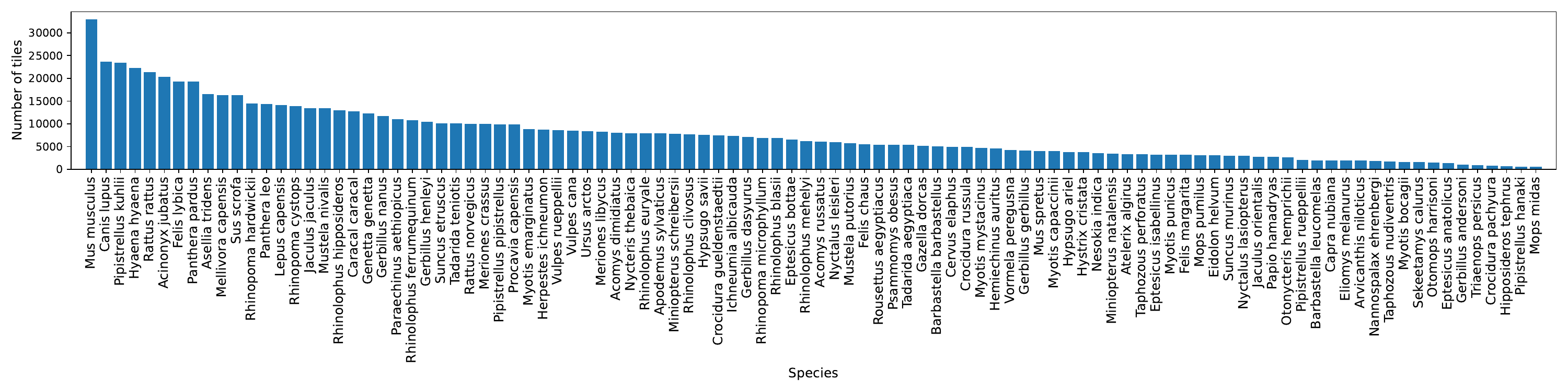}
    \caption{\textbf{Species occurrence across all tiles.} Multiple species can be present per tile.}
    \label{fig:tiles_per_species}
\end{figure*}

\begin{figure}
    \centering
    \includegraphics[width=0.6\linewidth]{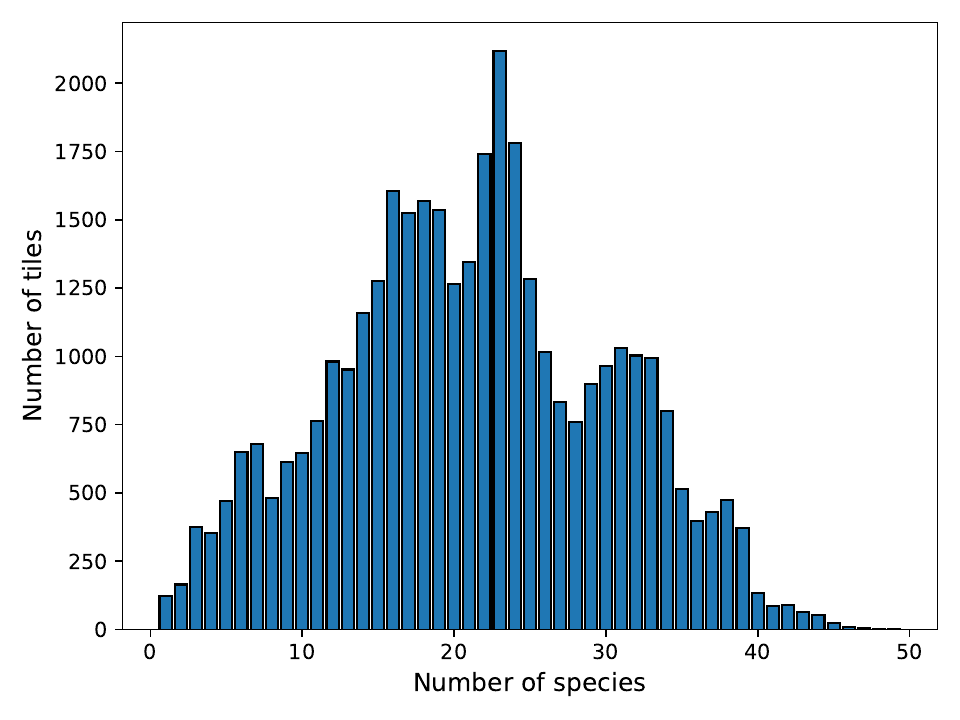}
    \caption{{\textbf{Species richness per tile.} Visualization of how many tiles contain each number of species. For example, we see that 23 is the most common number of species to occur in a tile, making this a strongly multi-label task.}}
    \label{fig:tile_species_counts}
\end{figure}

Now, we need to generate points from these species ranges. For each row in the GeoDataFrame we just created, we extract the species name as well as its range in and out of Africa using the GeoJSON for the boundary of Africa we made before. We aim to generate 100 points in Africa and 300 out of Africa. We first extract the bounding box of the range in Africa. We then uniformly randomly pick $x$ and $y$ coordinates within the bounding box. We create a point from these coordinates with an outer tile. If the outer tile intersects the species' range in Africa, we save the point to a list along with the name of the species and its outer tile. We continue generating coordinates until we have points for the species in Africa. This same process is then repeated for outside Africa. Afterwards, we combine the in-Africa and out-of-Africa points into a single list and remove overlaps among them. We append this list to an overall list of points. When this process has been completed for all species, our overall list has 39,676 points. We remove overlaps among all of those, resulting in 39,011 points. We then loop through each point in this list and check whether its outer tile overlaps with any of the other species' ranges. If so, we add the species to the point's species list. Finally, we shuffle the points and assign them IDs before saving the data to a GeoJSON. The maximum and minimum numbers of points per species are 35,468 and 530, respectively. 

\Cref{fig:species_missing_modality_counts} shows how many points are missing each of the modalities after generating tiles. The maximum and minimum numbers of tiles per species are 32,982 and 525, respectively. We assign each species a distinct integer label from 0-99 to be used during modeling. The minimum number of tiles in which a species appears by split is 187 for train, 32 for validation, 46 for random test, and 173 for geographic test. \cref{fig:tiles_per_species} shows how many tiles each species occurs in, and \cref{fig:tile_species_counts} plots how many tiles contain a certain number of species. 

\subsection{Generating tiles}
Once we have saved the GeoJSONs containing the coordinates of all the points, we are ready to generate tiles with paired modalities and task data. For a given task, for each point, we first need to determine a date range to be used for filtering modality data. For biomass, we first extract the GEDI points within the point's outer tile following the same procedure as described earlier. Then, we extract each GEDI point's leaf-on and leaf-off day and create a list with unique pairs of (leaf-on, leaf-off) days. We then convert each integer day into a valid date format. The date filter to be used is then any date that falls within any of those date ranges, corresponding to a leaf-on time period. For the soil and species tasks, we estimate the growing season by using the months May-September for northern-hemisphere points and November-March for southern-hemisphere points and use that as the date filter. We generate a 1280x1280-m tile centered on the point by applying a buffer.

\subsubsection{Sentinel-2}
We obtain Sentinel-2 images from the \href{https://developers.google.com/earth-engine/datasets/catalog/COPERNICUS_S2_SR_HARMONIZED}{Harmonized Sentinel-2 MSI: MultiSpectral Instrument, Level-2A} dataset on Google Earth Engine. We first filter by the preselected date range to extract all images with a valid date. Then we filter by images that have some overlap with the tile. We then apply a stricter filter to ensure that we only keep images that contain the tile plus a 200-m buffer to avoid edge effects. If no images pass all the filters, then we do not include the point in the resulting dataset. If at least one image did pass the filters, we use the MSK\_CLDPRB band to compute a cloudy pixel fraction for each image within the tile, where we define a cloudy pixel as one with $>=$ 10\% cloud probability. We then filter to only keep images with $<10$\% cloudy pixels. We again check if any images have passed the filter and skip over the point if none have. If there is at least one image left, we apply a second cloud filter using the \href{https://developers.google.com/earth-engine/datasets/catalog/COPERNICUS_S2_CLOUD_PROBABILITY}{Sentinel-2: Cloud Probability} dataset from Google Earth Engine. We apply the same three filters as above. We then save the data as an additional band for each of the images, using the ``system:index'' value for pairing. We then use this S2CLOUDLESS band to compute another cloudy pixel fraction, again using 10\% as the threshold for classifying a pixel as cloudy and filtering by images with $<10$\% cloudy pixels. We again skip over the point if no images have passed the filter. If there are some left, we then apply a filter against invalid pixels using the SCL band in Sentinel-2 images. We only keep images that have less than 10\% no-data pixels in the SCL band. We then sort the images in ascending order by their fraction of no-data SCL pixels and take the one with the lowest value. We once again skip over the point if no images have passed the filter. We access the date property of the image, the projection of the B4 band, and the CRS of the B4 band. We then project the coordinates of the tile using that projection and round the coordinates to land on the nearest pixel intersection. We then set the tile to be the 128x128-pixel square centered at that pixel intersection. We then apply a bilinear projection to the bands in the image with continuous pixel values and unmask to the Sentinel-2 no-data value of 65,535. We apply a nearest-neighbor projection to SCL which has categorical pixel values and unmask it to 0, which was its original no-data value. We then save the data for all of the bands. We save B1, B2, B3, B4, B5, B6, B7, B8, B8A, B9, B11, and B12 as our Sentinel-2 modality bands, and we also save MSK\_CLDPRB, S2CLOUDLESS, and SCL as additional metadata. This results in a 12-band continuous-valued modality.

\subsubsection{Sentinel-1}
We obtain Sentinel-1 images from the \href{https://developers.google.com/earth-engine/datasets/catalog/COPERNICUS_S1_GRD}{Sentinel-1 SAR GRD: C-band Synthetic Aperture Radar Ground Range Detected, log scaling} dataset on Google Earth Engine. We apply the same three filters as for Sentinel-2 and also filter by images with the interferometric wide swath mode. We then calculate the difference between the date of each image and the date of the Sentinel-2 image we selected. We sort the images in ascending order by how many days off they are from the Sentinel-2 image date. We extract all the ascending images and then take the first one. We also extract all the descending images and then take the first one. For both the ascending and descending image, we save the VV, VH, HH, and HV bands bilinearly projected to match Sentinel-2 if they exist, otherwise we save them as a NaN band using the Sentinel-1 no-data value of -9,999. We also unmask the image to the Sentinel-1 no-data value. This results in an 8-band continuous-valued modality. If there is neither an ascending nor descending image, we mark the modality as missing.

\subsubsection{ASTER GDEM}
We access elevation data from the GDEM dataset within the sat-io Aster data available on Google Earth Engine. We mask no-data values and then calculate the slope from the elevation data. We then bilinearly project both the elevation and slope bands and unmask them to the no-data value. This results in a 2-band continuous-valued modality. If all the pixels in the elevation band are no-data, we mark the modality as missing.

\subsubsection{ETH Global Canopy Height}
We access canopy height data from the ETH Global Canopy Height 2020 10m map~\cite{lang2023high} available on Google Earth Engine. We use both the height and uncertainty bands, bilinearly reproject them, and unmask to the original no-data value of 255 for the product, resulting in a 2-band continuous-valued modality. If all the pixels in the height band are no-data, we mark the modality as missing.

\subsubsection{Dynamic World}
We obtain Dynamic World data from the \href{https://developers.google.com/earth-engine/datasets/catalog/GOOGLE_DYNAMICWORLD_V1}{Dynamic World V1} collection on Google Earth Engine, which has 9 landcover classes~\cite{dynamic_world}. We filter by images within 2020 and those that have some overlap with the selected tile. We then get the mode for each pixel across images and use that as our image. If no image was found, we create a constant-valued image filled with the no-data value of 9 and reproject it using nearest-neighbor resampling. Otherwise, we reproject the image we obtained and unmask it to the no-data value. This yields a 1-band categorical-valued modality. If all the pixels in the image are no-data, we mark the modality as missing.

\subsubsection{ESA WorldCover}
We obtain ESA WorldCover data from the \href{https://developers.google.com/earth-engine/datasets/catalog/ESA_WorldCover_v100}{ESA WorldCover 10m v100} collection on Google Earth Engine, which has 11 landcover classes~\cite{esa_worldcover}. We remap the class labels such that $10\to0$, $20\to1$, $30\to2$, $40\to3$, $50\to4$, $60\to5$, $70\to6$, $80\to7$, $90\to8$, $95\to9$, and $100\to10$. We project the image using nearest-neighbor resampling and unmask it to the no-data value of 11, yielding a 1-band categorical-valued modality. If all the pixels in the image are no-data, we mark the modality as missing.

\subsubsection{Precipitation \& Temperature}
We obtain precipitation and temperature data from the \href{https://developers.google.com/earth-engine/datasets/catalog/ECMWF_ERA5_LAND_MONTHLY_AGGR}{ERA5-Land Monthly Aggregated - ECMWF Climate Reanalysis} dataset on Google Earth Engine. We collect temperature mean, min, and max as well as total precipitation data for the month corresponding to the Sentinel-2 image, the month prior, and the year ending with the selected month. This data already exists at the month-level and we compute it at the year-level. For each band (3 for precipitation and 9 for temperature), we compute the mean over the tile. If the value is None, then we instead use the no-data value of -9,999. This results in a 3-band continuous-valued modality for precipitation and a 9-band continuous-valued modality for temperature.

\subsubsection{Geolocation}
We extract the longitude and latitude of the centroid of the tile that has been snapped to the Sentinel-2 grid, yielding a 2-band continuos-valued modality. We also compute a geolocation encoding:\\$[\cos(\text{lon}), \sin(\text{lon}), \cos(\text{lat}), \sin(\text{lat})]$.

\subsubsection{Sentinel-2 date}
We save the date of the selected Sentinel-2 image as a string in YYYY-MM-dd format, yielding a 1-band continuous-valued modality. We also extract the month from this date and compute a month encoding by $[\cos(\pi\times\text{month}/6), \sin(\pi\times\text{month}/6)]$.

\subsubsection{Biome \& Ecoregion}
We store the biome and ecoregion names along with their corresponding numbers in \texttt{biome\_labels.json} and \texttt{ecoregion\_labels.json}, respectively. We access the same ecoregions dataset as before and filter it by the centroid of the final tile. We then extract the biome and ecoregion information from that if it is available. If it is, we map each biome name to an integer in 0-12 and each ecoregion name to an integer in 0-845 using the JSONs mentioned above. If the biome information is not available, we set biome to the no-data value of 13. If the ecoregion information is not available. we set ecoregion to the no-data value of 846. This results in a 1-band categorical-valued modality for biomass and a 1-band categorical-valued modality for ecoregion. 

\subsubsection{Task Data}
We must also include the data for each task in the tiles. For biomass, we initialize a NaN image with -9,999 constant values and then write over it with the GEDI points we extracted for the tile. We reproject this to match the Sentinel-2 grid with nearest-neighbor resampling. For the soil properties, we save the value of the property together with the point in the GeoJSON, so we simply read in that value. For species, we save the list of species present in the tile together with the point in the GeoJSON, so we can simply read in that list.

\subsubsection{Saving Data}
\begin{figure}
    \centering
    \begin{subfigure}{\linewidth}
        \centering
        \includegraphics[width=0.5\linewidth]{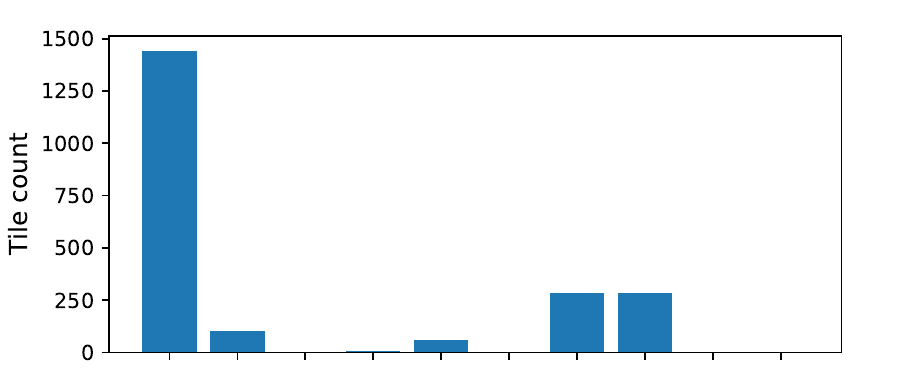}
        \caption{Biomass}
        \label{fig:biomass_missing_modality_counts}
    \end{subfigure}
    \\
    \begin{subfigure}{\linewidth}
        \centering
        \includegraphics[width=0.5\linewidth]{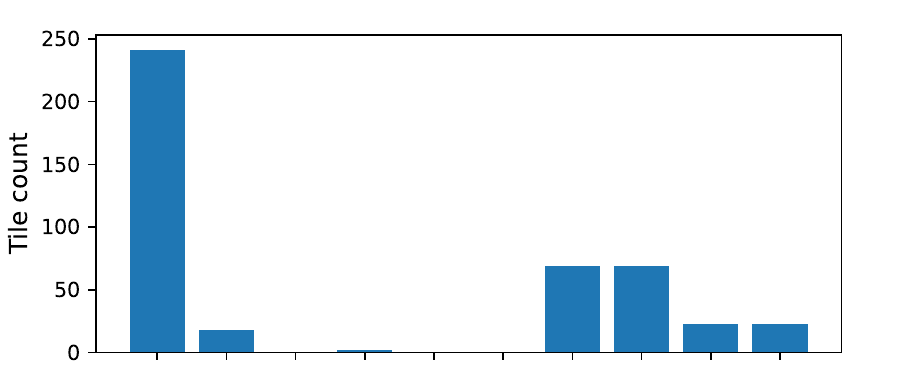}
        \caption{Soil nitrogen}
        \label{fig:soil_nitrogen_missing_modality_counts}
    \end{subfigure}
    \\
    \begin{subfigure}{\linewidth}
        \centering
        \includegraphics[width=0.5\linewidth]{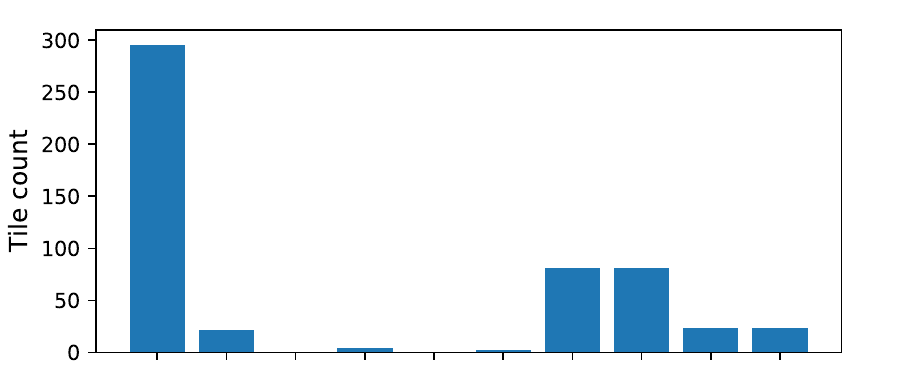}
        \caption{Soil organic carbon}
        \label{fig:soil_organic_carbon_missing_modality_counts}
    \end{subfigure}
    \\
    \begin{subfigure}{\linewidth}
        \centering
        \includegraphics[width=0.5\linewidth]{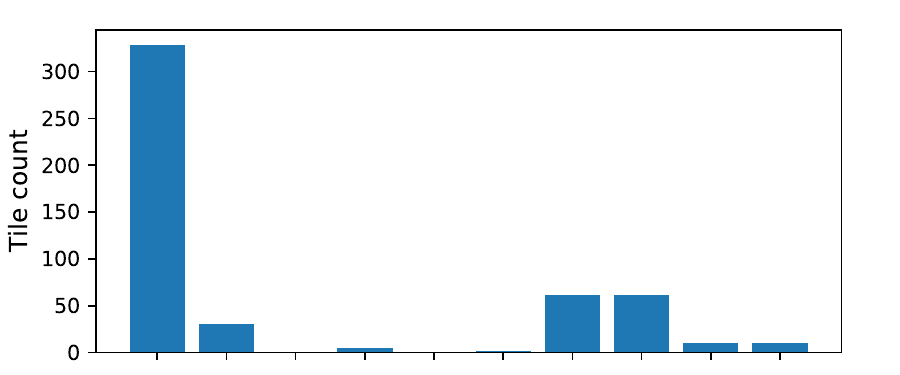}
        \caption{Soil pH}
        \label{fig:soil_pH_missing_modality_counts}
    \end{subfigure}
    \\
    \begin{subfigure}{\linewidth}
    \centering
    \includegraphics[width=0.5\linewidth]{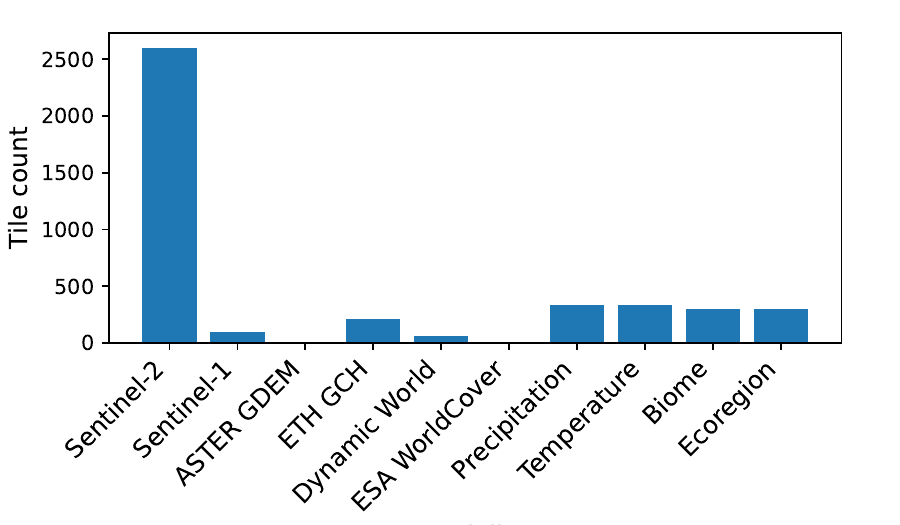}
            \caption{Species}
\label{fig:species_missing_modality_counts}
\end{subfigure}
    \caption{\textbf{Missing modality counts per task.} When downloading modality data from GEE for the tiles, we record which modalities are missing. If Sentinel-2 was missing, we did not save the tile or attempt to download the rest of the modalities.}
    \label{fig:missing_modality_counts}
\end{figure}

All modalities except for geolocation and date could in theory be unavailable for a tile. In \cref{fig:missing_modality_counts}, we visualize how many tiles for each task have missing modalities. If Sentinel\-/2 was missing, we did not save the tile. The data for all the other tiles is downloaded and saved as a TIFF. Each TIFF contains an array of shape (29,128,128) except for biomass, which has an additional band for the biomass data, yielding shape (30,128,128). The band names are shown in \cref{tab:tiff_bands}. Each TIFF also has a dictionary of tags containing the tile\-/level data (tile\-/level modalities along with the geolocation encoding, month encoding, missing modalities, MSK\_CLDPRB cloudy pixel fraction, S2CLOUDLESS cloudy pixel fraction, SCL no data fraction, and task value if the task is not biomass). The TIFF file also contains the tile ID, CRS, and transform. We save TIFFs for all tiles we can generate and then merge the TIFFs into a single H5 file for each task. The keys in the H5 file are Sentinel2, Sentinel1, ASTER\_GDEM, ETH\_GCH, DynamicWorld, ESA\_WorldCover, precipitation, temperature, geolocation, sentinel2\_date, biome, ecoregion, geolocation\_encoding, month\_encoding, MSK\_CLDPRB, MSK\_CLDPRB\_CLOUDY\_PIXEL\_FRACTION, S2CLOUDLESS, \seqsplit{S2CLOUDLESS\_CLOUDY\_PIXEL\_FRACTION}, SCL, SCL\_NO\_DATA\_PIXEL\_FRACTION, crs, id, missing\_modalities, transform, sentinel2\_system\_index, and task, where ``task'' is the name of the task. Sentinel-2 system index uniquely identifies the Sentinel-2 image used for each tile.

\begin{table}
\centering
\caption{\textbf{TIFF band indices and names.} Note that Band 29 ``biomass'' is only present for the biomass task.}
\label{tab:tiff_bands}
\begin{tabular}{@{} 
    r   
    l   
@{}}
\toprule
\textbf{Band index} & 
\textbf{Band name} \\
\midrule
0 & Sentinel2\_B1  \\
1 & Sentinel2\_B2 \\
2 & Sentinel2\_B3 \\
3 & Sentinel2\_B4 \\
4 & Sentinel2\_B5 \\
5 & Sentinel2\_B6 \\
6 & Sentinel2\_B7 \\
7 & Sentinel2\_B8 \\
8 & Sentinel2\_B8A \\
9 & Sentinel2\_B9 \\
10 & Sentinel2\_B11 \\
11 & Sentinel2\_B12 \\
12 & MSK\_CLDPRB \\
13 & S2CLOUDLESS \\
14 & SCL \\
15 & Sentinel1\_ascending\_VV \\
16 & Sentinel1\_ascending\_VH \\
17 & Sentinel1\_ascending\_HH \\
18 & Sentinel1\_ascending\_HV \\
19 & Sentinel1\_descending\_VV \\
20 & Sentinel1\_descending\_VH \\
21 & Sentinel1\_descending\_HH \\
22 & Sentinel1\_descending\_HV \\
23 & ASTER\_GDEM\_elevation \\
24 & ASTER\_GDEM\_slope \\
25 & ETH\_GCH\_height \\
26 & ETH\_GCH\_uncertainty \\
27 & DynamicWorld \\
28 & ESA\_WorldCover \\
29 & biomass \\
\bottomrule
\end{tabular}
\end{table}
\FloatBarrier

\subsection{Dataset splits}
We split each task dataset into a train set, validation set, random test set, and geographic test set. We use the tiles in Africa as the geographic test set, and we randomly split the tiles in the rest of the world into train, validation, and random test sets with ratios 70\%/15\%/15\%.

We do this by first extracting the bounding box for each tile and reprojecting it into EPSG 4326. Each tile is also assigned an index. We split these boxes into an Africa group and a non-Africa group. The Africa group contains all tiles that intersect our Africa polygon, and the non-Africa group contains the rest. We randomly shuffle the non-Africa tiles and then assign the first 70\% for training, the next 15\% for validation, and the remaining 15\% for random testing. We calculate normalization statistics on the training tiles for the Sentinel-2, Sentinel-1, ASTER GDEM, ETH GCH, precipitation, and temperature modalities. In particular, we mask out the no-data value for each modality and then compute the mean and standard deviation for each band. We save the task's train, validation, and random test indices along with the normalization statistics to a JSON file.

\subsection{MMEarth-Bench Explorer}
\begin{figure*}[ht]
    \centering
    \begin{subfigure}[b]{0.45\textwidth}
        \centering
        \includegraphics[width=\textwidth]{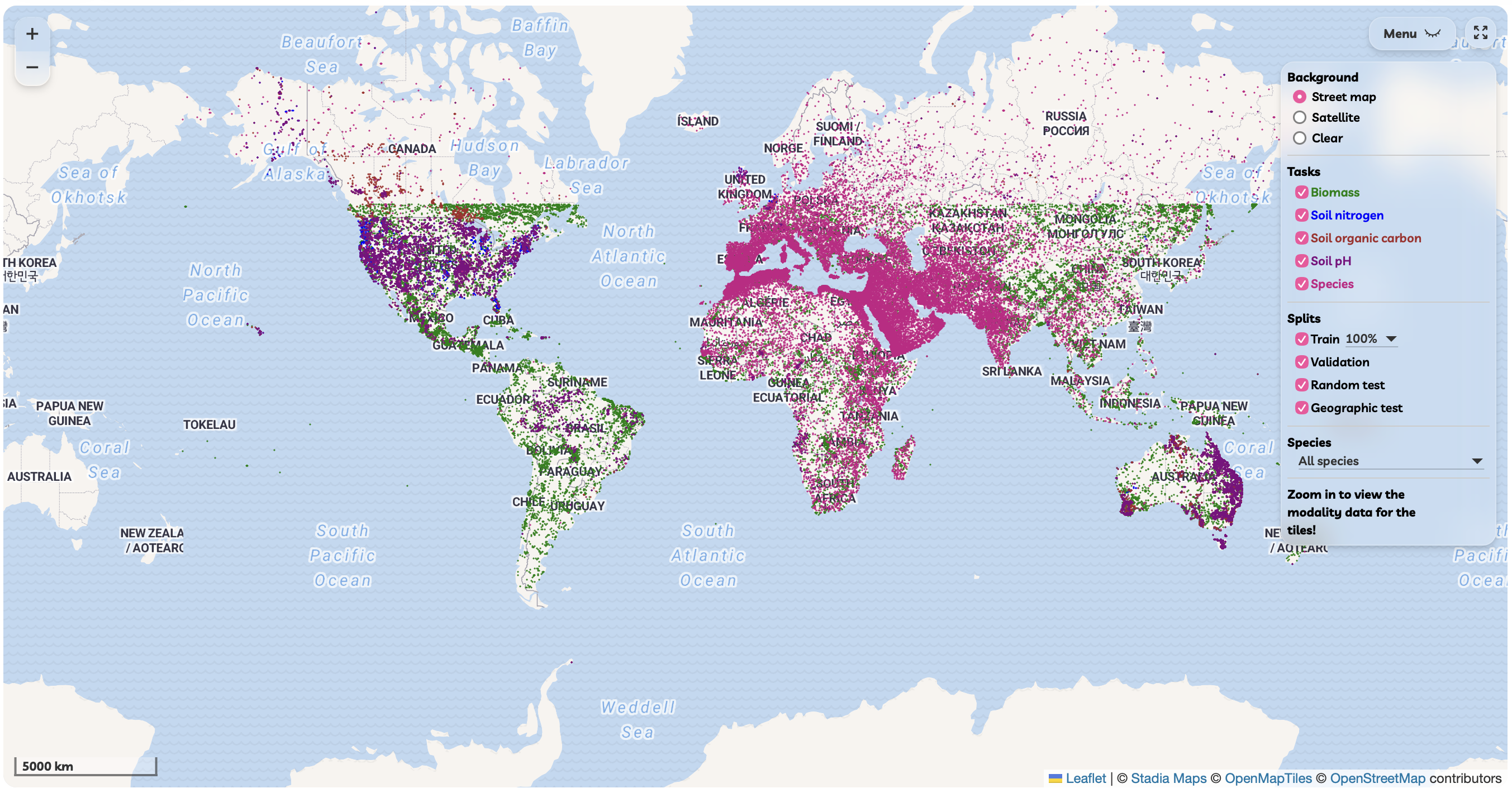}
        \caption{Full map view.}
        \label{fig:a}
    \end{subfigure}
    \hfill
    \begin{subfigure}[b]{0.45\textwidth}
        \centering
        \includegraphics[width=\textwidth]{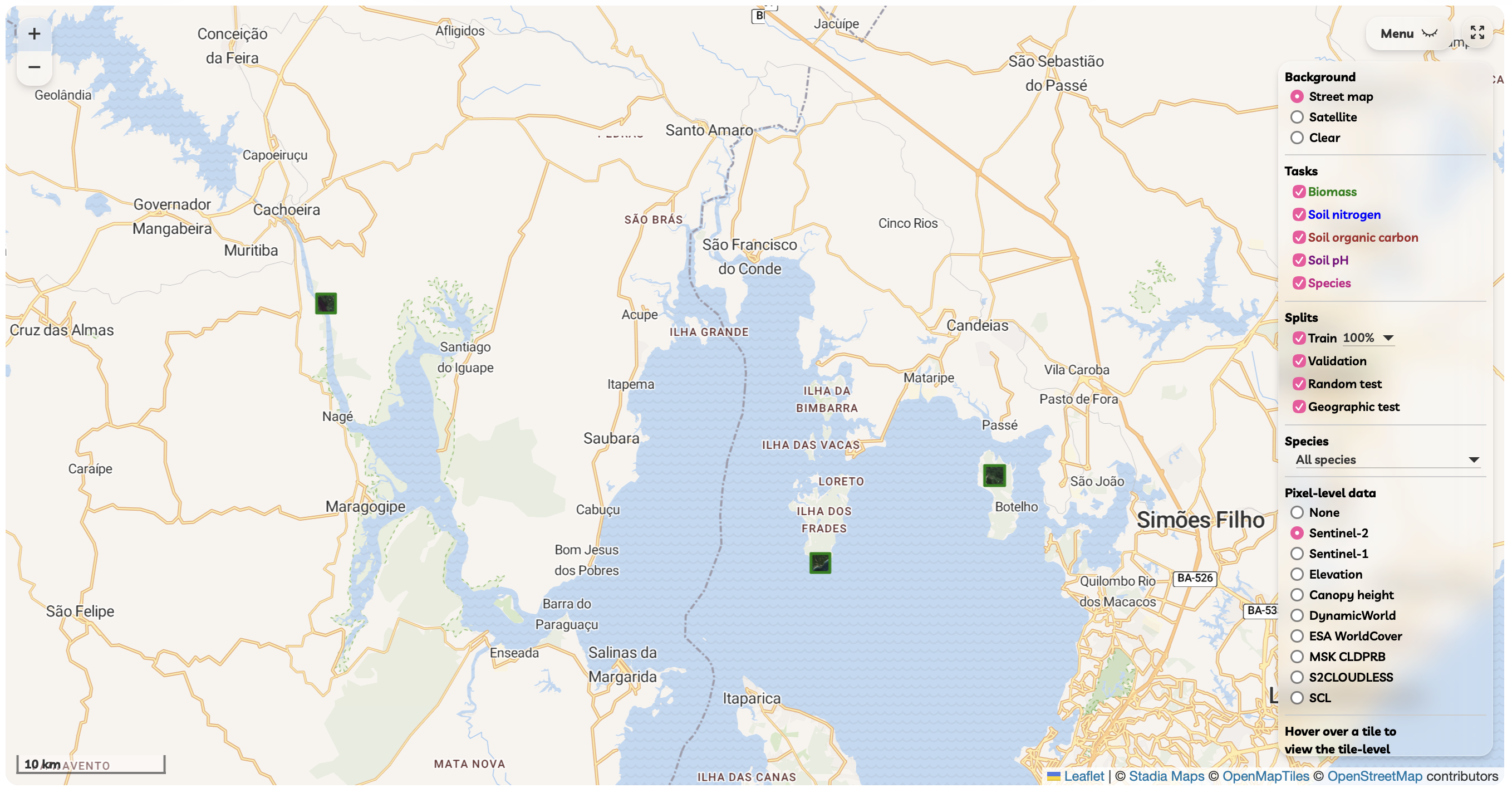}
        \caption{Zooming in to three biomass tiles and displaying S2.}
        \label{fig:b}
    \end{subfigure}

    \vspace{1em} 

    \begin{subfigure}[b]{0.45\textwidth}
        \centering
        \includegraphics[width=\textwidth]{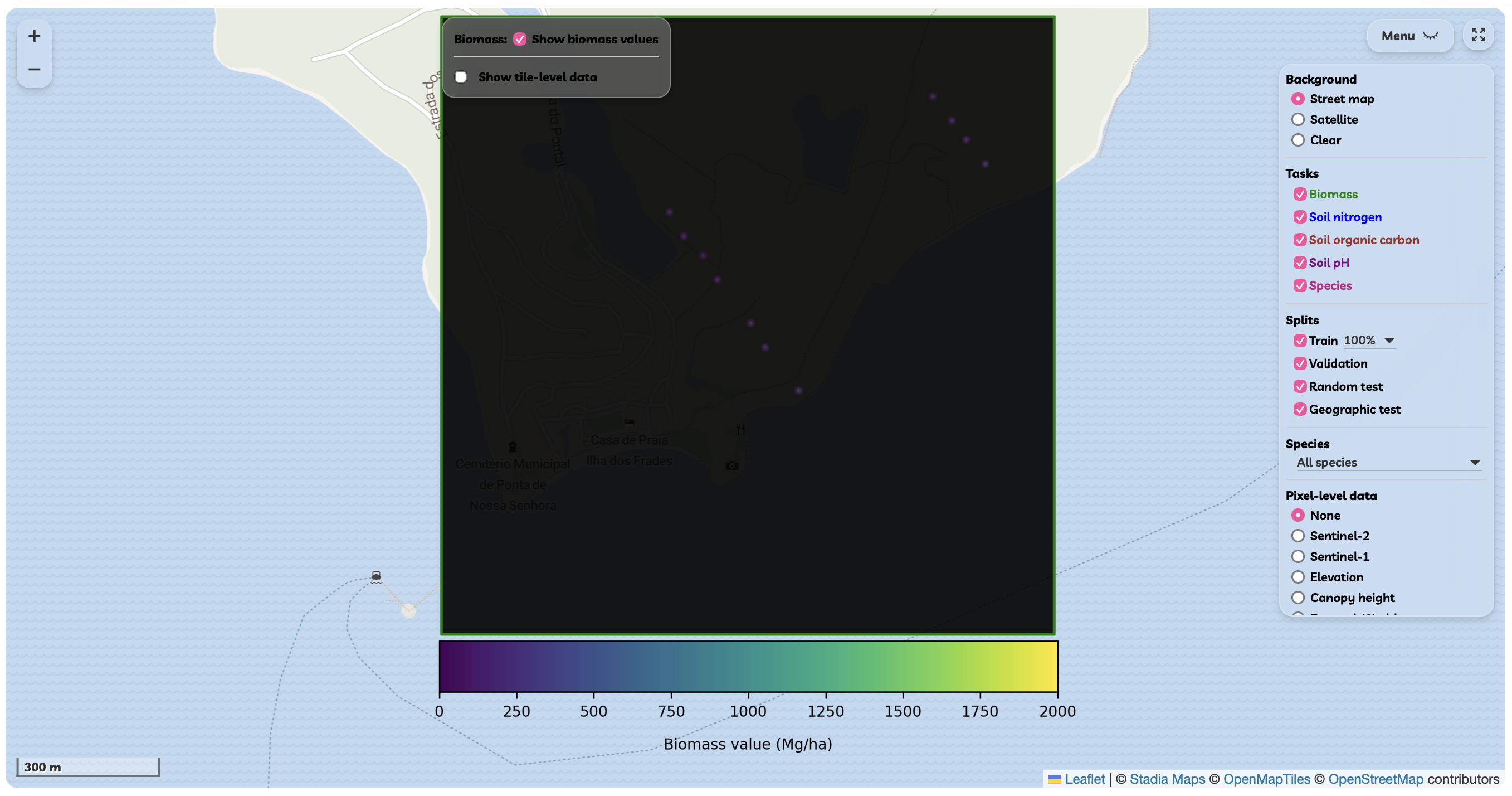}
        \caption{Displaying biomass pixel-level values.}
        \label{fig:c}
    \end{subfigure}
    \hfill
    \begin{subfigure}[b]{0.45\textwidth}
        \centering
        \includegraphics[width=\textwidth]{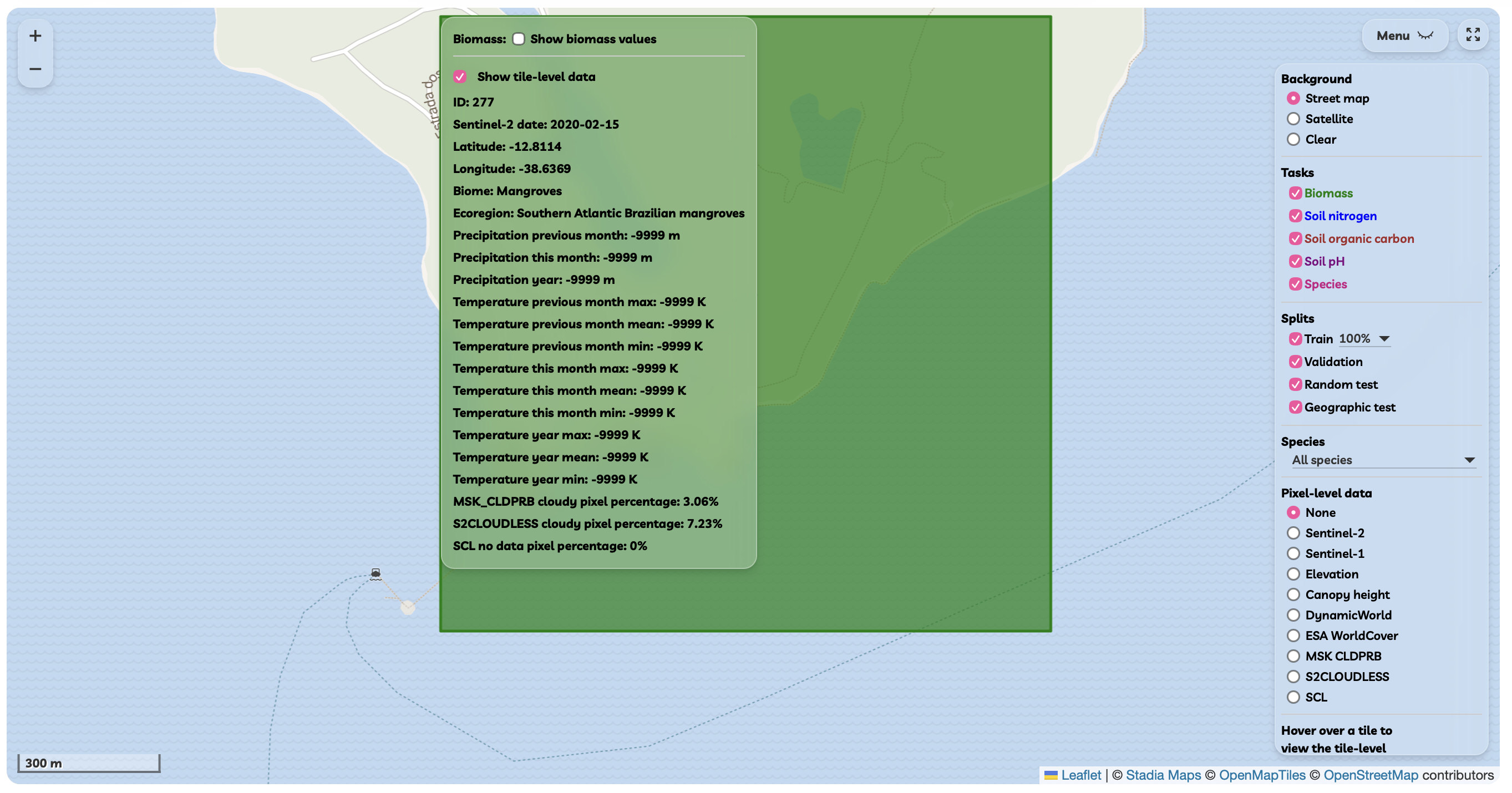}
        \caption{Tile-level data for the tile.}
        \label{fig:d}
    \end{subfigure}
    \caption{\href{https://lgordon99.github.io/mmearth-bench-app/}{\textbf{MMEarth-Bench explorer.}}
    \cref{fig:a} shows the full map view, where the tiles for all the tasks are overlaid in distinct colors. The user can toggle tasks on and off. Zooming into the map reveals the option to view modality data for the tiles, as shown in \cref{fig:b}. The user can click on any of the pixel-level modalities to see the corresponding images overlaid on the tiles in view. Hovering the mouse over a tile reveals the option to see more data, as shown in \cref{fig:c}. For biomass, the pixel-level biomass data can be shown. For all the tasks, the user can click to see the values for the tile's tile-level modalities, as shown in \cref{fig:d}. Additionally, we incorporate split data into the map so that users can toggle the tiles by split (train 100\%, train 50\%, train 5\%, validation, random test, geographic test).}
    \label{fig:visualization_tool}
\end{figure*}

We have developed an interactive map visualization tool available at 
\\\href{https://lgordon99.github.io/mmearth-bench-app/}{https://lgordon99.github.io/mmearth-bench-app/} to visualize the\\MMEarth-Bench dataset.
The app was made with HTML, CSS, and JavaScript and runs in the browser. We show selected screenshots in \cref{fig:visualization_tool}.

\FloatBarrier
\section{Practitioner's guide to using \textsc{TTT-MMR}}
Our test-time training method is not limited to the MMEarth-Bench tasks. Others wishing to adapt a pretrained model to their downstream task can extract the MMEarth modalities using our code and then run JT followed by \textsc{TTT-MMR} in order to get more accurate predictions. Downloading the MMEarth modalities requires a GeoJSON of (lon, lat) coordinates for the task data. See \cref{fig:downloading_mmearth_modalities} for an example. This GeoJSON should be accessible at a path such as \texttt{DATA\_DIR/TASK/TASK\_points.geojson}. The modalities can then be downloaded using \texttt{python get\_tile\_data.py TASK}.

Our TTT approach takes $<$1 sec/sample at inference time, making it highly usable for practitioners (see \cref{tab:compute-time}). With $<$10k samples per test set, it takes maximum 2.5 hours on a single GPU.

\begin{figure}
\begin{lstlisting}
{
    "type": "FeatureCollection",
    "features": [
        {
            "type": "Feature",
            "geometry": {
                "type": "Point",
                "coordinates": [
                    lon0,
                    lat0
                ]
            },
            "id": 0
        },
        {
            "type": "Feature",
            "geometry": {
                "type": "Point",
                "coordinates": [
                    lon1,
                    lat1
                ]
            },
            "id": 1
        }
    ]
}
\end{lstlisting}
\caption{\textbf{Example GeoJSON for downloading the MMEarth modalities from (lon, lat) coordinates.} The user should populate the file with the locations where they have downstream task data.}
\label{fig:downloading_mmearth_modalities}
\end{figure}

\begin{table}
\centering
\caption{Compute time for inference (seconds/sample)}
\setlength{\tabcolsep}{5pt}
\begin{tabular}{lccc}
\toprule
& DINOv3 Sat & MPMAE & Copernicus-FM \\
 \midrule
Standard forward pass & 0.007 & 0.002 & 0.009 \\
\textsc{TTT-MMR-Geo} (5 iters) & 0.910 & 0.367 & 0.706 \\
\bottomrule
\end{tabular}
\label{tab:compute-time}
\end{table}

\section{Experimental setup}
\subsection{Hyperparameter settings}

The hyperparameters for all our experiments are listed in \cref{tab:hyperparam-settings}. We train for 100 epochs with batch size 64. We use a maximum learning rate of $10^{-4}$ and a weight decay of 0.05~\cite{pangaea}. We use the AdamW optimizer~\cite{adamw} and a learning rate scheduler with a linear warm-up of 10 epochs followed by cosine annealing~\cite{copernicusfm} with a minimum learning rate of $10^{-6}$~\cite{multimae}.

\begin{table}
\centering
\caption{
\textbf{Hyperparameter settings.}
}
\begin{tabular}{ll}
\toprule
\textbf{Hyperparameter} & \textbf{Value}  \\

\midrule
Optimizer & AdamW~\cite{adamw} \\ 
Learning rate scheduler & Linear warmup with cosine annealing\\
Max learning rate & $10^{-4}$ \\
Min learning rate & $10^{-6}$ \\
Weight decay & 0.05 \\
Batch size & 64 \\
Training epochs & 100 \\
Warm-up epochs & 10 \\
\bottomrule
\end{tabular}

\label{tab:hyperparam-settings}
\end{table}

\subsection{Pretrained geospatial models benchmarked}
\label{sec:pretrained_models}

\subsubsection{RGB input}
Scale-MAE~\cite{scalemae} pretrains a ViT-L model according to the masked autoencoding framework with the addition of a ground sampling distance (GSD) positional encoding. They pretrain on the fMoW-RGB~\cite{fmow} dataset, which contains remotely sensed images of varying resolution and GSD. DINOv3~\cite{simeoni2025dinov3} trains a 7B-parameter DINO model, a variant of the ViT architecture, with the new Gram anchoring strategy to remove noise from the feature maps. Subsequently, this model is distilled into smaller models, such as ViT-L. We benchmark DINOv3 Web, the ViT-L distilled from the 7B-parameter model pretrained on the web imagery dataset LVD-1689M~\cite{simeoni2025dinov3}, and DINOv3 Sat, the ViT-L distilled from the 7B-parameter model pretrained on their SAT-493M dataset~\cite{simeoni2025dinov3} derived from Maxar RGB images.

\subsubsection{Sentinel-2 input}
SatlasNet~\cite{satlas} contains a Swin Transformer backbone and separate heads for different label types (e.g., segmentation or point). They train both multi-image (time series) and single-image models. All models are pretrained on SatlasPretrain~\cite{satlas}, a dataset of 856K tiles with an associated time series of remote sensing imagery of varying resolution (Sentinel-2 and NAIP) and labels drawn from 137 categories. We benchmark the single-image (9-band) Sentinel-2 pretrained Swin-v2-B backbone. MPMAE~\cite{mmearth} is a multi-pretext masked autoencoder approach to training a ConvNeXtV2-based architecture ~\cite{woo2023convnext}. They pretrain in a self-supervised manner on multimodal pretext tasks using the data in the MMEarth dataset~\cite{mmearth} in order to learn semantic representations for Sentinel-2, the model's input. The MMEarth pretraining dataset contains the same modalities as our MMEarth-Bench. We benchmark their atto-sized model that was pretrained on MMEarth64 with an uncertainty-weighted loss on 56$\times$56-pixel images.

\subsubsection{Multimodal input}
TerraMind is the first any\-/to\-/any generative multimodal geospatial model~\cite{jakubik2025terramind}. It is pretrained using a dual token- and pixel-level approach on TerraMesh, a multimodal pretraining dataset with Sentinel\-/2, Sentinel\-/1, land use/land cover, NDVI, DEM, geolocation, and captions~\cite{terramesh}. We benchmark their TerraMindv1-B model, an architecture based on ViT-B. An extension of DOFA~\cite{dofa}, Copernicus-FM is a model that can process data from any spectral or non-spectral sensor with varying resolution through dynamic hypernetworks and also integrates metadata (e.g., geolocation and date) ~\cite{copernicusfm}. They pretrain a ViT-B model on the Copernicus-Pretrain dataset~\cite{copernicusfm}, which includes data from Sentinel-1, -2, -3, and 5P, as well as Copernicus DEM. Galileo also uses an architecture based on ViT, modified to accept channel groups and variable input shapes and timesteps~\cite{tseng2025galileo}. They also use a novel pretraining objective with contributions from both the pixel and the latent space, encouraging learning global and local features. Their pretraining dataset of $\approx$ 100k instances includes Sentinel-1 and -2 data, elevation and slope, Dynamic World and World Cereal landcover maps, ERA5, TerraClimate, VIIRS nighttime lights, LandScan, and geolocation. We benchmark their ViT-B-based pretrained model.

As these models mix ascending and descending Sentinel-1 images during training, for each tile we compute the number of valid pixels in the ascending VV and VH bands as well as the number of valid pixels in the descending VV and VH bands. We then take the pair of VV, VH bands that has more valid pixels, either ascending or descending. In the case of a tie we select the ascending VV and VH bands.

\subsection{Modality reconstruction}
When reconstructing modalities during JT and \textsc{TTT-MMR}, we reconstruct the cyclic-encoded geolocation and month as opposed to the raw values, following prior work \cite{mmearth}.

\subsection{\textsc{TTT-MMR-Geo} batching}
\begin{figure}
    \centering
    \includegraphics[width=0.6\linewidth]{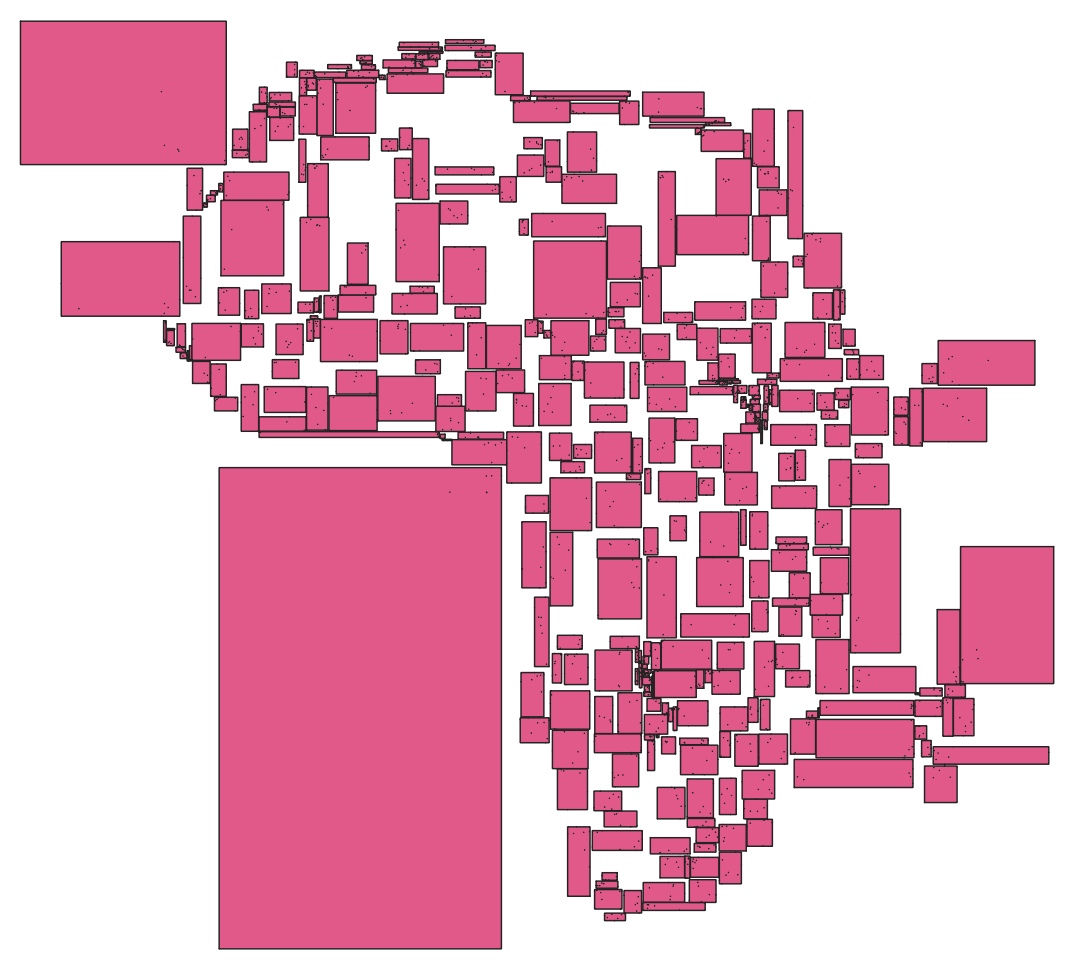}
    \caption{\textbf{Biomass geographic test split batches for \textsc{TTT-MMR-Geo}.} The pink rectangles represent the bounding boxes for the batches. Each black point is a tile in the geographic test set. Each rectangle contains 8 tiles except for one, which contains the batch size (8) plus any remainder (4 in this case).}
    \label{fig:ttt-mmr-geo-batches}
\end{figure}

In \cref{fig:ttt-mmr-geo-batches} we show an example of the geographically contiguous, non\-/overlapping batches in \textsc{TTT-MMR-Geo}.

\subsection{Evaluation metrics}
\subsubsection{Coefficient of determination ($R^2$)}
We use $R^2$ as the evaluation metric for our regression tasks, where $R^2\in(-\infty,1]$.
$$R^2=1-\frac{\sum_i(y_i-f(x_i))^2}{\sum_i(y_i-\overline{y})^2}$$
A negative $R^2$ means that the predictions of the model $f$ are worse than predicting the mean of the test data $\overline{y}$.

\subsubsection{Mean Average Precision (mAP)}
We use mean average precision as the evaluation metric for our multi-label classification task, where $\text{mAP}\in[0,1]$. This is equivalent to the area under the precision-recall curve.
$$\text{AP}=\sum_n(R_n-R_{n-1})P_n$$
$P_n$ and $R_n$ are the precision and recall, respectively, at threshold $n$. To calculate mAP, we calculate AP for each class and then compute the mean over all classes.

\subsubsection{Relative Improvement (RI)}
In Tab. 5 we show the average rankings of JT, \textsc{TTT-MMR}, and \textsc{TTT-MMR-Geo} averaged over tasks for each model. Since not all tasks share the same metric, performance deltas cannot simply be averaged across tasks. Moreover, even averaging raw deltas is not ideal for $R^2$, as it is nonlinear in the error. One option is to convert changes in $R^2$ and mAP to a relative improvement metric that can then be averaged across tasks. 
$$\text{RI}=\frac{\text{metric}_\text{new}-\text{metric}_\text{old}}{1-\text{metric}_\text{old}}$$
For $R^2$, this RI metric can be interpreted as the decrease in the fraction of unexplained variance between two results. For mAP, it represents the decrease in the gap to perfect performance. The fractional relative improvement is multiplied by 100 to yield a relative percent improvement.

\section{Additional results}
\subsection{Finetuning \vs linear probing}
\label{sec:FTvsLP}
\begin{figure}
    \centering
    \includegraphics[width=\linewidth]{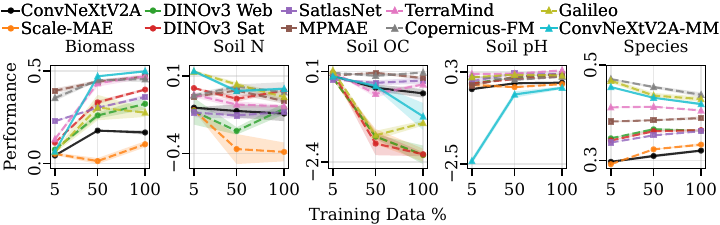}
    \caption{\textbf{Finetuning performance on geographic split.} We observe a weaker relationship between more training data and an improvement in performance than for the random split (Fig. 4).}
    \label{fig:rq1_FT_geographic}
\end{figure}

\begin{figure}
    \centering
    \includegraphics[width=\linewidth]{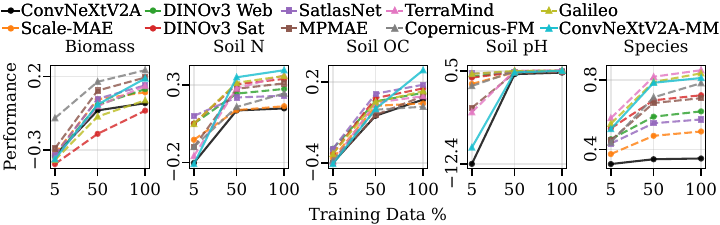}
    \caption{\textbf{Linear probing performance on random split.} As for finetuning in Fig. 4., performance improves with more training data on the random split.}
    \label{fig:rq1_LP_random}
\end{figure}

\begin{figure}
    \centering
    \includegraphics[width=\linewidth]{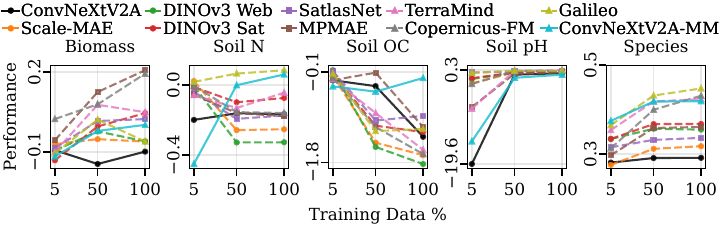}
    \caption{\textbf{Linear probing performance on geographic split.} As in \cref{fig:rq1_FT_geographic}, more training data usually improves performance on the geographic split but not always. Soil N and soil OC are counterexamples.}
    \label{fig:rq1_LP_geographic}
\end{figure}

\begin{figure}
    \centering
    \includegraphics[width=\linewidth]{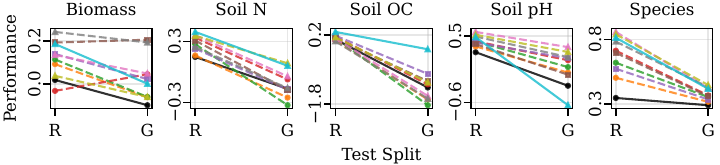}
    \caption{\textbf{Geographic generalization for linear probing.} Performance comparison on random (R) \vs geographic (G) test splits using all training data. As in Fig. 5 for finetuning, all the tasks other than biomass suffer large performance drops when switching from the random to the geographic test split. We generally see a less severe geographic generalization gap for linear probing than finetuning. This could be due to linear probing having a lower capacity for fitting to the source domain than finetuning.}
    \label{fig:rq2_LP}
\end{figure}

\begin{figure}
    \centering
    \includegraphics[width=\linewidth]{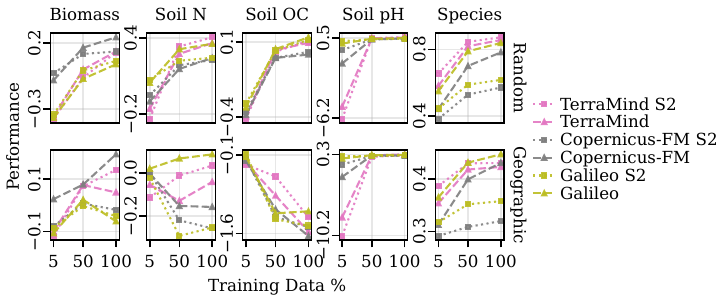}
    \caption{\textbf{Unimodal \vs multimodal input data for linear probing.} We see similar patterns as for finetuning in Fig. 6.}
    \label{fig:rq3_LP}
\end{figure}

\cref{fig:rq1_FT_geographic} is the equivalent of Fig. 4 but on the geographic test split. We additionally run linear probing experiments, in which the encoder is frozen and only the task decoder is trained. The \emph{linear probing} results are reported in \cref{fig:rq1_LP_random}, \cref{fig:rq1_LP_geographic}, \cref{fig:rq2_LP}, and \cref{fig:rq3_LP}, which correspond to the \emph{finetuning} results presented in Fig. 4, \cref{fig:rq1_FT_geographic}, Fig. 5, and Fig. 6, respectively. For readability, we provide brief interpretations of the results in the figure captions.

\FloatBarrier
\subsection{Model rankings}
\cref{tab:ft_ranked_models_by_task_100,tab:ft_ranked_models_by_task_50,tab:ft_ranked_models_by_task_5} show the models ranked overall and by task for different training subsets. \cref{tab:ft_ranks_by_task} shows model rankings on each task and overall. \cref{tab:ft_metrics_by_task} shows model test performance on each task.

\begin{table*}[ht]
\centering
\caption{\textbf{Model rankings by task and overall after finetuning on 100\% of the training data.} Models are ordered by lowest average rank over seeds for each task and split, or over tasks and seeds for the ``All tasks'' column. Lower is better.}
\label{tab:ft_ranked_models_by_task_100}
\resizebox{\linewidth}{!}{%
\begin{tabular}{lccccccc}
\toprule
\textbf{Split} & \textbf{Rank} & \textbf{All tasks} & \textbf{Biomass} & \textbf{Soil N} & \textbf{Soil OC} & \textbf{Soil pH} & \textbf{Species} \\
\midrule
\multirow{10}{*}{\textbf{Random}} & \textbf{1} & \textbf{\textcolor[RGB]{23,190,207}{ConvNeXtV2A-MM}} & \textbf{\textcolor[RGB]{140,86,75}{MPMAE}} & \textbf{\textcolor[RGB]{227,119,194}{TerraMind}} & \textbf{\textcolor[RGB]{127,127,127}{Copernicus-FM}} & \textbf{\textcolor[RGB]{227,119,194}{TerraMind}} & \textbf{\textcolor[RGB]{127,127,127}{Copernicus-FM}} \\
 & 2 & \textcolor[RGB]{127,127,127}{Copernicus-FM} & \textcolor[RGB]{23,190,207}{ConvNeXtV2A-MM} & \textcolor[RGB]{23,190,207}{ConvNeXtV2A-MM} & \textcolor[RGB]{188,189,34}{Galileo} & \textcolor[RGB]{23,190,207}{ConvNeXtV2A-MM} & \textcolor[RGB]{188,189,34}{Galileo} \\
 & 3 & \textcolor[RGB]{227,119,194}{TerraMind} & \textcolor[RGB]{127,127,127}{Copernicus-FM} & \textcolor[RGB]{127,127,127}{Copernicus-FM} & \textcolor[RGB]{23,190,207}{ConvNeXtV2A-MM} & \textcolor[RGB]{188,189,34}{Galileo} & \textcolor[RGB]{23,190,207}{ConvNeXtV2A-MM} \\
 & 4 & \textcolor[RGB]{188,189,34}{Galileo} & \textcolor[RGB]{227,119,194}{TerraMind} & \textcolor[RGB]{188,189,34}{Galileo} & \textcolor[RGB]{140,86,75}{MPMAE} & \textcolor[RGB]{127,127,127}{Copernicus-FM} & \textcolor[RGB]{227,119,194}{TerraMind} \\
 & 5 & \textcolor[RGB]{140,86,75}{MPMAE} & \textcolor[RGB]{188,189,34}{Galileo} & \textcolor[RGB]{140,86,75}{MPMAE} & \textcolor[RGB]{227,119,194}{TerraMind} & \textcolor[RGB]{140,86,75}{MPMAE} & \textcolor[RGB]{140,86,75}{MPMAE} \\
 & 6 & \textcolor[RGB]{148,103,189}{SatlasNet} & \textcolor[RGB]{214,39,40}{DINOv3 Sat} & \textcolor[RGB]{214,39,40}{DINOv3 Sat} & \textcolor[RGB]{148,103,189}{SatlasNet} & \textcolor[RGB]{44,160,44}{DINOv3 Web} & \textcolor[RGB]{148,103,189}{SatlasNet} \\
 & 7 & \textcolor[RGB]{214,39,40}{DINOv3 Sat} & \textcolor[RGB]{148,103,189}{SatlasNet} & \textcolor[RGB]{44,160,44}{DINOv3 Web} & \textcolor[RGB]{31,119,180}{ConvNeXtV2A} & \textcolor[RGB]{214,39,40}{DINOv3 Sat} & \textcolor[RGB]{214,39,40}{DINOv3 Sat} \\
 & 8 & \textcolor[RGB]{44,160,44}{DINOv3 Web} & \textcolor[RGB]{44,160,44}{DINOv3 Web} & \textcolor[RGB]{148,103,189}{SatlasNet} & \textcolor[RGB]{44,160,44}{DINOv3 Web} & \textcolor[RGB]{148,103,189}{SatlasNet} & \textcolor[RGB]{44,160,44}{DINOv3 Web} \\
 & 9 & \textcolor[RGB]{31,119,180}{ConvNeXtV2A} & \textcolor[RGB]{255,127,14}{Scale-MAE} & \textcolor[RGB]{31,119,180}{ConvNeXtV2A} & \textcolor[RGB]{255,127,14}{Scale-MAE} & \textcolor[RGB]{255,127,14}{Scale-MAE} & \textcolor[RGB]{255,127,14}{Scale-MAE} \\
 & 10 & \textcolor[RGB]{255,127,14}{Scale-MAE} & \textcolor[RGB]{31,119,180}{ConvNeXtV2A} & \textcolor[RGB]{255,127,14}{Scale-MAE} & \textcolor[RGB]{214,39,40}{DINOv3 Sat} & \textcolor[RGB]{31,119,180}{ConvNeXtV2A} & \textcolor[RGB]{31,119,180}{ConvNeXtV2A} \\
\midrule
\multirow{10}{*}{\textbf{Geographic}} & \textbf{1} & \textbf{\textcolor[RGB]{127,127,127}{Copernicus-FM}} & \textbf{\textcolor[RGB]{23,190,207}{ConvNeXtV2A-MM}} & \textbf{\textcolor[RGB]{23,190,207}{ConvNeXtV2A-MM}} & \textbf{\textcolor[RGB]{127,127,127}{Copernicus-FM}} & \textbf{\textcolor[RGB]{227,119,194}{TerraMind}} & \textbf{\textcolor[RGB]{127,127,127}{Copernicus-FM}} \\
 & 2 & \textcolor[RGB]{140,86,75}{MPMAE} & \textcolor[RGB]{140,86,75}{MPMAE} & \textcolor[RGB]{127,127,127}{Copernicus-FM} & \textcolor[RGB]{140,86,75}{MPMAE} & \textcolor[RGB]{140,86,75}{MPMAE} & \textcolor[RGB]{188,189,34}{Galileo} \\
 & 3 & \textcolor[RGB]{227,119,194}{TerraMind} & \textcolor[RGB]{227,119,194}{TerraMind} & \textcolor[RGB]{214,39,40}{DINOv3 Sat} & \textcolor[RGB]{148,103,189}{SatlasNet} & \textcolor[RGB]{148,103,189}{SatlasNet} & \textcolor[RGB]{23,190,207}{ConvNeXtV2A-MM} \\
 & 4 & \textcolor[RGB]{23,190,207}{ConvNeXtV2A-MM} & \textcolor[RGB]{127,127,127}{Copernicus-FM} & \textcolor[RGB]{188,189,34}{Galileo} & \textcolor[RGB]{227,119,194}{TerraMind} & \textcolor[RGB]{188,189,34}{Galileo} & \textcolor[RGB]{227,119,194}{TerraMind} \\
 & 5 & \textcolor[RGB]{188,189,34}{Galileo} & \textcolor[RGB]{214,39,40}{DINOv3 Sat} & \textcolor[RGB]{140,86,75}{MPMAE} & \textcolor[RGB]{31,119,180}{ConvNeXtV2A} & \textcolor[RGB]{44,160,44}{DINOv3 Web} & \textcolor[RGB]{140,86,75}{MPMAE} \\
 & 6 & \textcolor[RGB]{148,103,189}{SatlasNet} & \textcolor[RGB]{148,103,189}{SatlasNet} & \textcolor[RGB]{227,119,194}{TerraMind} & \textcolor[RGB]{23,190,207}{ConvNeXtV2A-MM} & \textcolor[RGB]{127,127,127}{Copernicus-FM} & \textcolor[RGB]{214,39,40}{DINOv3 Sat} \\
 & 7 & \textcolor[RGB]{214,39,40}{DINOv3 Sat} & \textcolor[RGB]{44,160,44}{DINOv3 Web} & \textcolor[RGB]{44,160,44}{DINOv3 Web} & \textcolor[RGB]{188,189,34}{Galileo} & \textcolor[RGB]{214,39,40}{DINOv3 Sat} & \textcolor[RGB]{44,160,44}{DINOv3 Web} \\
 & 8 & \textcolor[RGB]{44,160,44}{DINOv3 Web} & \textcolor[RGB]{188,189,34}{Galileo} & \textcolor[RGB]{31,119,180}{ConvNeXtV2A} & \textcolor[RGB]{44,160,44}{DINOv3 Web} & \textcolor[RGB]{31,119,180}{ConvNeXtV2A} & \textcolor[RGB]{148,103,189}{SatlasNet} \\
 & 9 & \textcolor[RGB]{31,119,180}{ConvNeXtV2A} & \textcolor[RGB]{31,119,180}{ConvNeXtV2A} & \textcolor[RGB]{148,103,189}{SatlasNet} & \textcolor[RGB]{255,127,14}{Scale-MAE} & \textcolor[RGB]{255,127,14}{Scale-MAE} & \textcolor[RGB]{255,127,14}{Scale-MAE} \\
 & 10 & \textcolor[RGB]{255,127,14}{Scale-MAE} & \textcolor[RGB]{255,127,14}{Scale-MAE} & \textcolor[RGB]{255,127,14}{Scale-MAE} & \textcolor[RGB]{214,39,40}{DINOv3 Sat} & \textcolor[RGB]{23,190,207}{ConvNeXtV2A-MM} & \textcolor[RGB]{31,119,180}{ConvNeXtV2A} \\
\bottomrule
\end{tabular}
}
\end{table*}

\begin{table*}[ht]
\centering
\caption{\textbf{Model rankings by task and overall after finetuning on 50\% of the training data.} Models are ordered by lowest average rank over seeds for each task and split, or over tasks and seeds for the ``All tasks'' column. Lower is better.}
\label{tab:ft_ranked_models_by_task_50}
\resizebox{\linewidth}{!}{%
\begin{tabular}{lccccccc}
\toprule
\textbf{Split} & \textbf{Rank} & \textbf{All tasks} & \textbf{Biomass} & \textbf{Soil N} & \textbf{Soil OC} & \textbf{Soil pH} & \textbf{Species} \\
\midrule
\multirow{10}{*}{\textbf{Random}} & \textbf{1} & \textbf{\textcolor[RGB]{127,127,127}{Copernicus-FM}} & \textbf{\textcolor[RGB]{127,127,127}{Copernicus-FM}} & \textbf{\textcolor[RGB]{227,119,194}{TerraMind}} & \textbf{\textcolor[RGB]{127,127,127}{Copernicus-FM}} & \textbf{\textcolor[RGB]{227,119,194}{TerraMind}} & \textbf{\textcolor[RGB]{127,127,127}{Copernicus-FM}} \\
 & 2 & \textcolor[RGB]{188,189,34}{Galileo} & \textcolor[RGB]{23,190,207}{ConvNeXtV2A-MM} & \textcolor[RGB]{127,127,127}{Copernicus-FM} & \textcolor[RGB]{188,189,34}{Galileo} & \textcolor[RGB]{23,190,207}{ConvNeXtV2A-MM} & \textcolor[RGB]{188,189,34}{Galileo} \\
 & 3 & \textcolor[RGB]{23,190,207}{ConvNeXtV2A-MM} & \textcolor[RGB]{140,86,75}{MPMAE} & \textcolor[RGB]{188,189,34}{Galileo} & \textcolor[RGB]{227,119,194}{TerraMind} & \textcolor[RGB]{188,189,34}{Galileo} & \textcolor[RGB]{23,190,207}{ConvNeXtV2A-MM} \\
 & 4 & \textcolor[RGB]{227,119,194}{TerraMind} & \textcolor[RGB]{188,189,34}{Galileo} & \textcolor[RGB]{23,190,207}{ConvNeXtV2A-MM} & \textcolor[RGB]{23,190,207}{ConvNeXtV2A-MM} & \textcolor[RGB]{127,127,127}{Copernicus-FM} & \textcolor[RGB]{227,119,194}{TerraMind} \\
 & 5 & \textcolor[RGB]{140,86,75}{MPMAE} & \textcolor[RGB]{227,119,194}{TerraMind} & \textcolor[RGB]{140,86,75}{MPMAE} & \textcolor[RGB]{148,103,189}{SatlasNet} & \textcolor[RGB]{140,86,75}{MPMAE} & \textcolor[RGB]{140,86,75}{MPMAE} \\
 & 6 & \textcolor[RGB]{148,103,189}{SatlasNet} & \textcolor[RGB]{148,103,189}{SatlasNet} & \textcolor[RGB]{44,160,44}{DINOv3 Web} & \textcolor[RGB]{140,86,75}{MPMAE} & \textcolor[RGB]{44,160,44}{DINOv3 Web} & \textcolor[RGB]{148,103,189}{SatlasNet} \\
 & 7 & \textcolor[RGB]{44,160,44}{DINOv3 Web} & \textcolor[RGB]{214,39,40}{DINOv3 Sat} & \textcolor[RGB]{214,39,40}{DINOv3 Sat} & \textcolor[RGB]{31,119,180}{ConvNeXtV2A} & \textcolor[RGB]{214,39,40}{DINOv3 Sat} & \textcolor[RGB]{214,39,40}{DINOv3 Sat} \\
 & 8 & \textcolor[RGB]{214,39,40}{DINOv3 Sat} & \textcolor[RGB]{44,160,44}{DINOv3 Web} & \textcolor[RGB]{148,103,189}{SatlasNet} & \textcolor[RGB]{44,160,44}{DINOv3 Web} & \textcolor[RGB]{148,103,189}{SatlasNet} & \textcolor[RGB]{44,160,44}{DINOv3 Web} \\
 & 9 & \textcolor[RGB]{31,119,180}{ConvNeXtV2A} & \textcolor[RGB]{31,119,180}{ConvNeXtV2A} & \textcolor[RGB]{31,119,180}{ConvNeXtV2A} & \textcolor[RGB]{214,39,40}{DINOv3 Sat} & \textcolor[RGB]{31,119,180}{ConvNeXtV2A} & \textcolor[RGB]{255,127,14}{Scale-MAE} \\
 & 10 & \textcolor[RGB]{255,127,14}{Scale-MAE} & \textcolor[RGB]{255,127,14}{Scale-MAE} & \textcolor[RGB]{255,127,14}{Scale-MAE} & \textcolor[RGB]{255,127,14}{Scale-MAE} & \textcolor[RGB]{255,127,14}{Scale-MAE} & \textcolor[RGB]{31,119,180}{ConvNeXtV2A} \\
\midrule
\multirow{10}{*}{\textbf{Geographic}} & \textbf{1} & \textbf{\textcolor[RGB]{127,127,127}{Copernicus-FM}} & \textbf{\textcolor[RGB]{23,190,207}{ConvNeXtV2A-MM}} & \textbf{\textcolor[RGB]{188,189,34}{Galileo}} & \textbf{\textcolor[RGB]{140,86,75}{MPMAE}} & \textbf{\textcolor[RGB]{148,103,189}{SatlasNet}} & \textbf{\textcolor[RGB]{127,127,127}{Copernicus-FM}} \\
 & 2 & \textcolor[RGB]{140,86,75}{MPMAE} & \textcolor[RGB]{127,127,127}{Copernicus-FM} & \textcolor[RGB]{127,127,127}{Copernicus-FM} & \textcolor[RGB]{127,127,127}{Copernicus-FM} & \textcolor[RGB]{227,119,194}{TerraMind} & \textcolor[RGB]{188,189,34}{Galileo} \\
 & 3 & \textcolor[RGB]{23,190,207}{ConvNeXtV2A-MM} & \textcolor[RGB]{140,86,75}{MPMAE} & \textcolor[RGB]{140,86,75}{MPMAE} & \textcolor[RGB]{148,103,189}{SatlasNet} & \textcolor[RGB]{188,189,34}{Galileo} & \textcolor[RGB]{23,190,207}{ConvNeXtV2A-MM} \\
 & 4 & \textcolor[RGB]{188,189,34}{Galileo} & \textcolor[RGB]{227,119,194}{TerraMind} & \textcolor[RGB]{23,190,207}{ConvNeXtV2A-MM} & \textcolor[RGB]{23,190,207}{ConvNeXtV2A-MM} & \textcolor[RGB]{140,86,75}{MPMAE} & \textcolor[RGB]{227,119,194}{TerraMind} \\
 & 5 & \textcolor[RGB]{227,119,194}{TerraMind} & \textcolor[RGB]{214,39,40}{DINOv3 Sat} & \textcolor[RGB]{214,39,40}{DINOv3 Sat} & \textcolor[RGB]{31,119,180}{ConvNeXtV2A} & \textcolor[RGB]{44,160,44}{DINOv3 Web} & \textcolor[RGB]{140,86,75}{MPMAE} \\
 & 6 & \textcolor[RGB]{148,103,189}{SatlasNet} & \textcolor[RGB]{188,189,34}{Galileo} & \textcolor[RGB]{227,119,194}{TerraMind} & \textcolor[RGB]{227,119,194}{TerraMind} & \textcolor[RGB]{214,39,40}{DINOv3 Sat} & \textcolor[RGB]{44,160,44}{DINOv3 Web} \\
 & 7 & \textcolor[RGB]{214,39,40}{DINOv3 Sat} & \textcolor[RGB]{148,103,189}{SatlasNet} & \textcolor[RGB]{31,119,180}{ConvNeXtV2A} & \textcolor[RGB]{188,189,34}{Galileo} & \textcolor[RGB]{127,127,127}{Copernicus-FM} & \textcolor[RGB]{214,39,40}{DINOv3 Sat} \\
 & 8 & \textcolor[RGB]{44,160,44}{DINOv3 Web} & \textcolor[RGB]{44,160,44}{DINOv3 Web} & \textcolor[RGB]{148,103,189}{SatlasNet} & \textcolor[RGB]{255,127,14}{Scale-MAE} & \textcolor[RGB]{31,119,180}{ConvNeXtV2A} & \textcolor[RGB]{148,103,189}{SatlasNet} \\
 & 9 & \textcolor[RGB]{31,119,180}{ConvNeXtV2A} & \textcolor[RGB]{31,119,180}{ConvNeXtV2A} & \textcolor[RGB]{44,160,44}{DINOv3 Web} & \textcolor[RGB]{214,39,40}{DINOv3 Sat} & \textcolor[RGB]{255,127,14}{Scale-MAE} & \textcolor[RGB]{255,127,14}{Scale-MAE} \\
 & 10 & \textcolor[RGB]{255,127,14}{Scale-MAE} & \textcolor[RGB]{255,127,14}{Scale-MAE} & \textcolor[RGB]{255,127,14}{Scale-MAE} & \textcolor[RGB]{44,160,44}{DINOv3 Web} & \textcolor[RGB]{23,190,207}{ConvNeXtV2A-MM} & \textcolor[RGB]{31,119,180}{ConvNeXtV2A} \\
\bottomrule
\end{tabular}
}
\end{table*}

\begin{table*}[ht]
\centering
\caption{\textbf{Model rankings by task and overall after finetuning on 5\% of the training data.} Models are ordered by lowest average rank over seeds for each task and split, or over tasks and seeds for the ``All tasks'' column. Lower is better.}
\label{tab:ft_ranked_models_by_task_5}
\resizebox{\linewidth}{!}{%
\begin{tabular}{lccccccc}
\toprule
\textbf{Split} & \textbf{Rank} & \textbf{All tasks} & \textbf{Biomass} & \textbf{Soil N} & \textbf{Soil OC} & \textbf{Soil pH} & \textbf{Species} \\
\midrule
\multirow{10}{*}{\textbf{Random}} & \textbf{1} & \textbf{\textcolor[RGB]{127,127,127}{Copernicus-FM}} & \textbf{\textcolor[RGB]{140,86,75}{MPMAE}} & \textbf{\textcolor[RGB]{23,190,207}{ConvNeXtV2A-MM}} & \textbf{\textcolor[RGB]{255,127,14}{Scale-MAE}} & \textbf{\textcolor[RGB]{23,190,207}{ConvNeXtV2A-MM}} & \textbf{\textcolor[RGB]{188,189,34}{Galileo}} \\
 & 2 & \textcolor[RGB]{188,189,34}{Galileo} & \textcolor[RGB]{127,127,127}{Copernicus-FM} & \textcolor[RGB]{127,127,127}{Copernicus-FM} & \textcolor[RGB]{44,160,44}{DINOv3 Web} & \textcolor[RGB]{188,189,34}{Galileo} & \textcolor[RGB]{127,127,127}{Copernicus-FM} \\
 & 3 & \textcolor[RGB]{227,119,194}{TerraMind} & \textcolor[RGB]{148,103,189}{SatlasNet} & \textcolor[RGB]{188,189,34}{Galileo} & \textcolor[RGB]{148,103,189}{SatlasNet} & \textcolor[RGB]{227,119,194}{TerraMind} & \textcolor[RGB]{23,190,207}{ConvNeXtV2A-MM} \\
 & 4 & \textcolor[RGB]{23,190,207}{ConvNeXtV2A-MM} & \textcolor[RGB]{255,127,14}{Scale-MAE} & \textcolor[RGB]{227,119,194}{TerraMind} & \textcolor[RGB]{214,39,40}{DINOv3 Sat} & \textcolor[RGB]{127,127,127}{Copernicus-FM} & \textcolor[RGB]{227,119,194}{TerraMind} \\
 & 5 & \textcolor[RGB]{140,86,75}{MPMAE} & \textcolor[RGB]{214,39,40}{DINOv3 Sat} & \textcolor[RGB]{140,86,75}{MPMAE} & \textcolor[RGB]{188,189,34}{Galileo} & \textcolor[RGB]{44,160,44}{DINOv3 Web} & \textcolor[RGB]{140,86,75}{MPMAE} \\
 & 6 & \textcolor[RGB]{148,103,189}{SatlasNet} & \textcolor[RGB]{44,160,44}{DINOv3 Web} & \textcolor[RGB]{214,39,40}{DINOv3 Sat} & \textcolor[RGB]{227,119,194}{TerraMind} & \textcolor[RGB]{140,86,75}{MPMAE} & \textcolor[RGB]{148,103,189}{SatlasNet} \\
 & 7 & \textcolor[RGB]{214,39,40}{DINOv3 Sat} & \textcolor[RGB]{227,119,194}{TerraMind} & \textcolor[RGB]{31,119,180}{ConvNeXtV2A} & \textcolor[RGB]{127,127,127}{Copernicus-FM} & \textcolor[RGB]{148,103,189}{SatlasNet} & \textcolor[RGB]{214,39,40}{DINOv3 Sat} \\
 & 8 & \textcolor[RGB]{44,160,44}{DINOv3 Web} & \textcolor[RGB]{188,189,34}{Galileo} & \textcolor[RGB]{148,103,189}{SatlasNet} & \textcolor[RGB]{31,119,180}{ConvNeXtV2A} & \textcolor[RGB]{214,39,40}{DINOv3 Sat} & \textcolor[RGB]{44,160,44}{DINOv3 Web} \\
 & 9 & \textcolor[RGB]{255,127,14}{Scale-MAE} & \textcolor[RGB]{23,190,207}{ConvNeXtV2A-MM} & \textcolor[RGB]{44,160,44}{DINOv3 Web} & \textcolor[RGB]{140,86,75}{MPMAE} & \textcolor[RGB]{31,119,180}{ConvNeXtV2A} & \textcolor[RGB]{255,127,14}{Scale-MAE} \\
 & 10 & \textcolor[RGB]{31,119,180}{ConvNeXtV2A} & \textcolor[RGB]{31,119,180}{ConvNeXtV2A} & \textcolor[RGB]{255,127,14}{Scale-MAE} & \textcolor[RGB]{23,190,207}{ConvNeXtV2A-MM} & \textcolor[RGB]{255,127,14}{Scale-MAE} & \textcolor[RGB]{31,119,180}{ConvNeXtV2A} \\
\midrule
\multirow{10}{*}{\textbf{Geographic}} & \textbf{1} & \textbf{\textcolor[RGB]{188,189,34}{Galileo}} & \textbf{\textcolor[RGB]{140,86,75}{MPMAE}} & \textbf{\textcolor[RGB]{23,190,207}{ConvNeXtV2A-MM}} & \textbf{\textcolor[RGB]{188,189,34}{Galileo}} & \textbf{\textcolor[RGB]{227,119,194}{TerraMind}} & \textbf{\textcolor[RGB]{127,127,127}{Copernicus-FM}} \\
 & 2 & \textcolor[RGB]{127,127,127}{Copernicus-FM} & \textcolor[RGB]{127,127,127}{Copernicus-FM} & \textcolor[RGB]{188,189,34}{Galileo} & \textcolor[RGB]{227,119,194}{TerraMind} & \textcolor[RGB]{188,189,34}{Galileo} & \textcolor[RGB]{188,189,34}{Galileo} \\
 & 3 & \textcolor[RGB]{227,119,194}{TerraMind} & \textcolor[RGB]{148,103,189}{SatlasNet} & \textcolor[RGB]{214,39,40}{DINOv3 Sat} & \textcolor[RGB]{127,127,127}{Copernicus-FM} & \textcolor[RGB]{127,127,127}{Copernicus-FM} & \textcolor[RGB]{23,190,207}{ConvNeXtV2A-MM} \\
 & 4 & \textcolor[RGB]{140,86,75}{MPMAE} & \textcolor[RGB]{227,119,194}{TerraMind} & \textcolor[RGB]{227,119,194}{TerraMind} & \textcolor[RGB]{214,39,40}{DINOv3 Sat} & \textcolor[RGB]{148,103,189}{SatlasNet} & \textcolor[RGB]{227,119,194}{TerraMind} \\
 & 5 & \textcolor[RGB]{214,39,40}{DINOv3 Sat} & \textcolor[RGB]{214,39,40}{DINOv3 Sat} & \textcolor[RGB]{127,127,127}{Copernicus-FM} & \textcolor[RGB]{140,86,75}{MPMAE} & \textcolor[RGB]{44,160,44}{DINOv3 Web} & \textcolor[RGB]{140,86,75}{MPMAE} \\
 & 6 & \textcolor[RGB]{23,190,207}{ConvNeXtV2A-MM} & \textcolor[RGB]{44,160,44}{DINOv3 Web} & \textcolor[RGB]{140,86,75}{MPMAE} & \textcolor[RGB]{31,119,180}{ConvNeXtV2A} & \textcolor[RGB]{214,39,40}{DINOv3 Sat} & \textcolor[RGB]{44,160,44}{DINOv3 Web} \\
 & 7 & \textcolor[RGB]{44,160,44}{DINOv3 Web} & \textcolor[RGB]{188,189,34}{Galileo} & \textcolor[RGB]{44,160,44}{DINOv3 Web} & \textcolor[RGB]{23,190,207}{ConvNeXtV2A-MM} & \textcolor[RGB]{255,127,14}{Scale-MAE} & \textcolor[RGB]{214,39,40}{DINOv3 Sat} \\
 & 8 & \textcolor[RGB]{148,103,189}{SatlasNet} & \textcolor[RGB]{23,190,207}{ConvNeXtV2A-MM} & \textcolor[RGB]{31,119,180}{ConvNeXtV2A} & \textcolor[RGB]{44,160,44}{DINOv3 Web} & \textcolor[RGB]{140,86,75}{MPMAE} & \textcolor[RGB]{148,103,189}{SatlasNet} \\
 & 9 & \textcolor[RGB]{31,119,180}{ConvNeXtV2A} & \textcolor[RGB]{255,127,14}{Scale-MAE} & \textcolor[RGB]{148,103,189}{SatlasNet} & \textcolor[RGB]{255,127,14}{Scale-MAE} & \textcolor[RGB]{31,119,180}{ConvNeXtV2A} & \textcolor[RGB]{31,119,180}{ConvNeXtV2A} \\
 & 10 & \textcolor[RGB]{255,127,14}{Scale-MAE} & \textcolor[RGB]{31,119,180}{ConvNeXtV2A} & \textcolor[RGB]{255,127,14}{Scale-MAE} & \textcolor[RGB]{148,103,189}{SatlasNet} & \textcolor[RGB]{23,190,207}{ConvNeXtV2A-MM} & \textcolor[RGB]{255,127,14}{Scale-MAE} \\
\bottomrule
\end{tabular}
}
\end{table*}

\begin{table*}[ht]
\centering
\caption{\textbf{Average model ranks for finetuning on all training data.} Ranks are mean $\pm$ standard error averaged over seeds, or over tasks and seeds for the ``All tasks'' column. Lower is better.}
\label{tab:ft_ranks_by_task}
\resizebox{\linewidth}{!}{%
\begin{tabular}{llcccccc}
\toprule
\textbf{Split} & \textbf{Model} & \textbf{All tasks} & \textbf{Biomass} & \textbf{Soil N} & \textbf{Soil OC} & \textbf{Soil pH} & \textbf{Species} \\
\midrule
\multirow{10}{*}{\textbf{Random}} & ConvNeXtV2A & $9.1 \pm 0.3$ & $9.7 \pm 0.3$ & $9.0 \pm 0.0$ & $7.0 \pm 0.0$ & $9.7 \pm 0.3$ & $10.0 \pm 0.0$ \\
 & Scale-MAE & $9.3 \pm 0.2$ & $9.3 \pm 0.3$ & $10.0 \pm 0.0$ & $9.0 \pm 0.6$ & $9.3 \pm 0.3$ & $9.0 \pm 0.0$ \\
 & DINOv3 Web & $7.3 \pm 0.3$ & $7.7 \pm 0.3$ & $6.3 \pm 0.3$ & $8.7 \pm 0.3$ & $6.0 \pm 0.0$ & $8.0 \pm 0.0$ \\
 & DINOv3 Sat & $7.1 \pm 0.4$ & $6.0 \pm 0.0$ & $6.0 \pm 0.6$ & $9.3 \pm 0.7$ & $7.0 \pm 0.0$ & $7.0 \pm 0.0$ \\
 & SatlasNet & $7.1 \pm 0.2$ & $7.3 \pm 0.3$ & $8.0 \pm 0.0$ & $6.0 \pm 0.0$ & $8.0 \pm 0.0$ & $6.0 \pm 0.0$ \\
 & MPMAE & $4.1 \pm 0.4$ & $\mathbf{1.3 \pm 0.3}$ & $5.3 \pm 0.9$ & $4.0 \pm 0.0$ & $5.0 \pm 0.0$ & $5.0 \pm 0.0$ \\
 & TerraMind & $2.9 \pm 0.5$ & $3.7 \pm 1.3$ & $\mathbf{1.0 \pm 0.0}$ & $5.0 \pm 0.0$ & $\mathbf{1.0 \pm 0.0}$ & $4.0 \pm 0.0$ \\
 & Copernicus-FM & $\mathbf{2.5 \pm 0.3}$ & $3.0 \pm 0.6$ & $3.0 \pm 0.6$ & $\mathbf{1.7 \pm 0.3}$ & $4.0 \pm 0.0$ & $\mathbf{1.0 \pm 0.0}$ \\
 & Galileo & $3.0 \pm 0.3$ & $4.3 \pm 0.3$ & $4.0 \pm 0.6$ & $\mathbf{1.7 \pm 0.7}$ & $3.0 \pm 0.0$ & $2.0 \pm 0.0$ \\
 & ConvNeXtV2A-MM & $\mathbf{2.5 \pm 0.1}$ & $2.7 \pm 0.3$ & $2.3 \pm 0.3$ & $2.7 \pm 0.3$ & $2.0 \pm 0.0$ & $3.0 \pm 0.0$ \\
\midrule
\multirow{10}{*}{\textbf{Geographic}} & ConvNeXtV2A & $7.9 \pm 0.5$ & $9.0 \pm 0.0$ & $7.7 \pm 1.3$ & $5.0 \pm 0.0$ & $8.0 \pm 0.0$ & $10.0 \pm 0.0$ \\
 & Scale-MAE & $9.4 \pm 0.2$ & $10.0 \pm 0.0$ & $10.0 \pm 0.0$ & $9.0 \pm 0.6$ & $9.0 \pm 0.0$ & $9.0 \pm 0.0$ \\
 & DINOv3 Web & $6.8 \pm 0.4$ & $7.0 \pm 0.6$ & $6.7 \pm 1.2$ & $8.7 \pm 0.7$ & $4.7 \pm 0.3$ & $7.0 \pm 0.0$ \\
 & DINOv3 Sat & $6.1 \pm 0.6$ & $5.0 \pm 0.0$ & $3.0 \pm 0.6$ & $9.3 \pm 0.3$ & $6.3 \pm 0.3$ & $6.7 \pm 0.7$ \\
 & SatlasNet & $5.5 \pm 0.6$ & $6.3 \pm 0.3$ & $8.0 \pm 0.0$ & $3.0 \pm 0.6$ & $2.7 \pm 0.7$ & $7.3 \pm 0.7$ \\
 & MPMAE & $3.5 \pm 0.4$ & $2.7 \pm 0.3$ & $5.0 \pm 1.0$ & $2.3 \pm 0.3$ & $2.7 \pm 0.3$ & $5.0 \pm 0.0$ \\
 & TerraMind & $3.6 \pm 0.5$ & $3.0 \pm 0.6$ & $6.3 \pm 0.7$ & $3.7 \pm 0.3$ & $\mathbf{1.0 \pm 0.0}$ & $4.0 \pm 0.0$ \\
 & Copernicus-FM & $\mathbf{3.0 \pm 0.6}$ & $3.3 \pm 0.7$ & $3.0 \pm 2.0$ & $\mathbf{1.0 \pm 0.0}$ & $6.3 \pm 0.7$ & $\mathbf{1.3 \pm 0.3}$ \\
 & Galileo & $4.8 \pm 0.6$ & $7.7 \pm 0.3$ & $3.7 \pm 0.3$ & $6.7 \pm 0.3$ & $4.3 \pm 0.9$ & $1.7 \pm 0.3$ \\
 & ConvNeXtV2A-MM & $4.4 \pm 0.9$ & $\mathbf{1.0 \pm 0.0}$ & $\mathbf{1.7 \pm 0.3}$ & $6.3 \pm 0.3$ & $10.0 \pm 0.0$ & $3.0 \pm 0.0$ \\
\bottomrule
\end{tabular}
}
\end{table*}

\begin{table*}[ht]
\centering
\caption{\textbf{Average model test performance after finetuning on all training data.} Values are mean $\pm$ standard error over seeds (R$^2$ for regression tasks; mAP for species). Higher is better.}
\label{tab:ft_metrics_by_task}
\resizebox{\linewidth}{!}{%
\begin{tabular}{llccccc}
\toprule
\textbf{Split} & \textbf{Model} & \textbf{Biomass} & \textbf{Soil N} & \textbf{Soil OC} & \textbf{Soil pH} & \textbf{Species} \\
\midrule
\multirow{10}{*}{\textbf{Random}} & ConvNeXtV2A & $0.16 \pm 0.00$ & $0.21 \pm 0.01$ & $0.20 \pm 0.01$ & $0.39 \pm 0.01$ & $0.47 \pm 0.00$ \\
 & Scale-MAE & $0.17 \pm 0.00$ & $0.18 \pm 0.01$ & $0.11 \pm 0.01$ & $0.39 \pm 0.00$ & $0.67 \pm 0.00$ \\
 & DINOv3 Web & $0.38 \pm 0.00$ & $0.38 \pm 0.01$ & $0.11 \pm 0.00$ & $0.55 \pm 0.00$ & $0.82 \pm 0.00$ \\
 & DINOv3 Sat & $0.40 \pm 0.00$ & $0.38 \pm 0.00$ & $0.08 \pm 0.02$ & $0.55 \pm 0.00$ & $0.83 \pm 0.00$ \\
 & SatlasNet & $0.38 \pm 0.00$ & $0.31 \pm 0.00$ & $0.31 \pm 0.00$ & $0.51 \pm 0.01$ & $0.84 \pm 0.00$ \\
 & MPMAE & $\mathbf{0.45 \pm 0.00}$ & $0.42 \pm 0.02$ & $0.40 \pm 0.00$ & $0.57 \pm 0.01$ & $0.90 \pm 0.00$ \\
 & TerraMind & $0.43 \pm 0.01$ & $\mathbf{0.50 \pm 0.01}$ & $0.33 \pm 0.00$ & $\mathbf{0.67 \pm 0.00}$ & $0.95 \pm 0.00$ \\
 & Copernicus-FM & $0.44 \pm 0.00$ & $0.46 \pm 0.01$ & $0.47 \pm 0.01$ & $0.62 \pm 0.00$ & $\mathbf{0.99 \pm 0.00}$ \\
 & Galileo & $0.43 \pm 0.00$ & $0.44 \pm 0.00$ & $\mathbf{0.48 \pm 0.02}$ & $0.63 \pm 0.00$ & $\mathbf{0.99 \pm 0.00}$ \\
 & ConvNeXtV2A-MM & $0.44 \pm 0.00$ & $0.48 \pm 0.00$ & $0.44 \pm 0.01$ & $0.65 \pm 0.00$ & $\mathbf{0.99 \pm 0.00}$ \\
\midrule
\multirow{10}{*}{\textbf{Geographic}} & ConvNeXtV2A & $0.17 \pm 0.01$ & $-0.14 \pm 0.05$ & $-0.50 \pm 0.12$ & $-0.02 \pm 0.01$ & $0.32 \pm 0.00$ \\
 & Scale-MAE & $0.10 \pm 0.02$ & $-0.39 \pm 0.06$ & $-2.21 \pm 0.18$ & $-0.07 \pm 0.02$ & $0.33 \pm 0.00$ \\
 & DINOv3 Web & $0.32 \pm 0.03$ & $-0.11 \pm 0.04$ & $-2.20 \pm 0.26$ & $0.18 \pm 0.00$ & $0.36 \pm 0.00$ \\
 & DINOv3 Sat & $0.40 \pm 0.01$ & $0.00 \pm 0.00$ & $-2.19 \pm 0.05$ & $0.16 \pm 0.00$ & $0.36 \pm 0.00$ \\
 & SatlasNet & $0.36 \pm 0.01$ & $-0.13 \pm 0.02$ & $-0.15 \pm 0.09$ & $0.24 \pm 0.03$ & $0.36 \pm 0.00$ \\
 & MPMAE & $0.47 \pm 0.01$ & $-0.06 \pm 0.02$ & $-0.08 \pm 0.05$ & $0.23 \pm 0.02$ & $0.39 \pm 0.00$ \\
 & TerraMind & $0.47 \pm 0.01$ & $-0.10 \pm 0.01$ & $-0.25 \pm 0.03$ & $\mathbf{0.35 \pm 0.01}$ & $0.41 \pm 0.00$ \\
 & Copernicus-FM & $0.46 \pm 0.02$ & $-0.01 \pm 0.07$ & $\mathbf{0.07 \pm 0.01}$ & $0.16 \pm 0.01$ & $\mathbf{0.44 \pm 0.01}$ \\
 & Galileo & $0.28 \pm 0.02$ & $-0.04 \pm 0.02$ & $-1.32 \pm 0.07$ & $0.20 \pm 0.02$ & $0.43 \pm 0.01$ \\
 & ConvNeXtV2A-MM & $\mathbf{0.50 \pm 0.01}$ & $\mathbf{0.02 \pm 0.01}$ & $-1.14 \pm 0.37$ & $-0.18 \pm 0.04$ & $0.42 \pm 0.00$ \\
\bottomrule
\end{tabular}
}
\end{table*}

\FloatBarrier
\subsection{Test-time training performance with relative improvement metric}
\begin{figure}
    \centering
    \includegraphics[width=\linewidth]{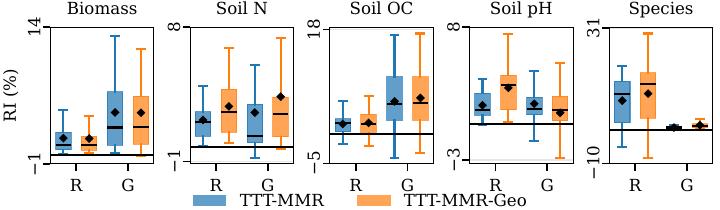}
    \caption{\textbf{Multimodal test-time training relative improvement (RI) per task.} RI represents the relative improvement ($\frac{\text{TTT}-\text{JT}}{1-\text{JT}}$) (multiplied by 100 to express a percentage). Patterns are generally consistent with Fig. 7, which showed the raw deltas. With the RI metric, however, we see that TTT leads to the greatest performance gains on the random split for species rather than the geographic split of soil OC, where R$^2$ values are mostly negative before and after TTT.}
    \label{fig:TTT-plot-RI}
\end{figure}

\cref{fig:TTT-plot-RI} is analogous to Fig. 7 except using the relative improvement metric instead of the raw delta.

\FloatBarrier
\begin{table*}
\caption{\textbf{Multimodal test-time training relative improvement (RI) per model.} Average improvement over JT of \textsc{TTT-MMR} and \textsc{TTT-MMR-Geo}. Values show the relative improvement ($\frac{\text{TTT}-\text{JT}}{1-\text{JT}}$) as mean $\pm$ standard error (multiplied by 100 to represent a percentage) averaged over tasks and seeds. Higher is better. These RI values support the claim in the main paper that the RGB-only models tend to undergo larger performance improvements as a result of TTT than the multimodal models. MPMAE benefits less from TTT than the RGB-only models and SatlasNet but more than the multimodal models. The RI metric shows that Galileo and ConvNeXtV2A-MM, the models with the most input modalities, do not always benefit from TTT on the random test split. This result does not appear when averaging ranks over tasks in Tab. 5 because it is primarily soil OC where TTT can slightly worsen performance.}
\centering
\resizebox{\linewidth}{!}{%
\begin{tabular}{llcccccccccc}
\toprule
\textbf{Split} & \textbf{Method} & \textbf{ConvNeXtV2A} & \textbf{Scale-MAE} & \textbf{DINOv3 Web} & \textbf{DINOv3 Sat} & \textbf{SatlasNet} & \textbf{MPMAE} & \textbf{TerraMind} & \textbf{Copernicus-FM} & \textbf{Galileo} & \textbf{ConvNeXtV2A-MM} \\
\midrule
\multirow{2}{*}{\textbf{Random}} & TTT-MMR & $3.5 \pm 0.7$ & $5.2 \pm 1.4$ & $5.5 \pm 1.6$ & $5.0 \pm 1.2$ & $5.7 \pm 1.9$ & $3.9 \pm 1.1$ & $2.7 \pm 1.0$ & $0.4 \pm 0.1$ & $\mathbf{-0.4 \pm 0.7}$ & $\mathbf{0.3 \pm 0.2}$ \\
 & TTT-MMR-Geo & $\mathbf{8.1 \pm 1.1}$ & $\mathbf{5.9 \pm 1.6}$ & $\mathbf{6.2 \pm 1.9}$ & $\mathbf{5.6 \pm 1.4}$ & $\mathbf{9.0 \pm 2.8}$ & $\mathbf{4.3 \pm 1.0}$ & $\mathbf{3.1 \pm 1.2}$ & $\mathbf{0.5 \pm 0.2}$ & $-1.3 \pm 1.0$ & $-0.4 \pm 0.5$ \\
\midrule
\multirow{2}{*}{\textbf{Geographic}} & TTT-MMR & $9.0 \pm 1.3$ & $2.7 \pm 0.7$ & $3.5 \pm 0.7$ & $3.4 \pm 1.1$ & $3.9 \pm 2.0$ & $1.8 \pm 0.6$ & $0.8 \pm 0.3$ & $0.6 \pm 0.1$ & $4.1 \pm 1.5$ & $\mathbf{0.5 \pm 1.3}$ \\
 & TTT-MMR-Geo & $\mathbf{9.1 \pm 1.4}$ & $\mathbf{2.9 \pm 0.7}$ & $\mathbf{3.6 \pm 0.6}$ & $\mathbf{3.6 \pm 1.2}$ & $\mathbf{5.1 \pm 1.9}$ & $\mathbf{2.7 \pm 0.7}$ & $\mathbf{1.0 \pm 0.3}$ & $\mathbf{0.9 \pm 0.2}$ & $\mathbf{4.4 \pm 1.6}$ & $0.1 \pm 1.7$ \\
\bottomrule
\end{tabular}}
\label{tab:ttt_by_model}
\end{table*}

\subsection{TTT-MMR iteration counts}
As the number of TTT iterations performed is determined using the validation set, in \cref{tab:ttt_iteration_numbers} we report the average across seeds for each task and model.

\begin{table*}[ht]
\centering
\caption{Average iteration number used in TTT-MMR for each task and model, averaged across seeds 41, 42, and 43.}
\begin{tabular}{lccccc}
\toprule
 & Biomass & Soil N & Soil OC & Soil pH & Species \\
\midrule
ConvNeXtV2A & 4.00 & 3.67 & 3.33 & 3.00 & 4.00 \\
Scale-MAE & 4.67 & 4.00 & 3.67 & 4.00 & 5.00 \\
DINOv3 Web & 4.33 & 3.67 & 4.00 & 4.00 & 5.00 \\
DINOv3 Sat & 4.33 & 4.00 & 4.33 & 4.00 & 5.00 \\
SatlasNet & 4.00 & 3.00 & 3.00 & 3.00 & 4.00 \\
MPMAE & 3.00 & 3.00 & 3.00 & 3.00 & 4.00 \\
TerraMind & 4.00 & 3.33 & 4.00 & 3.33 & 5.00 \\
Copernicus-FM & 4.00 & 3.00 & 3.00 & 3.33 & 4.00 \\
Galileo & 3.67 & 3.00 & 3.00 & 3.33 & 2.00 \\
ConvNeXtV2A-MM & 3.00 & 3.00 & 3.00 & 3.00 & 2.67 \\
\bottomrule
\end{tabular}
\label{tab:ttt_iteration_numbers}
\end{table*}

\FloatBarrier
\subsection{Modality importance in TTT-MMR}
\begin{figure}
    \centering
    \begin{subfigure}{\textwidth}
    \centering
        \includegraphics[width=\linewidth]{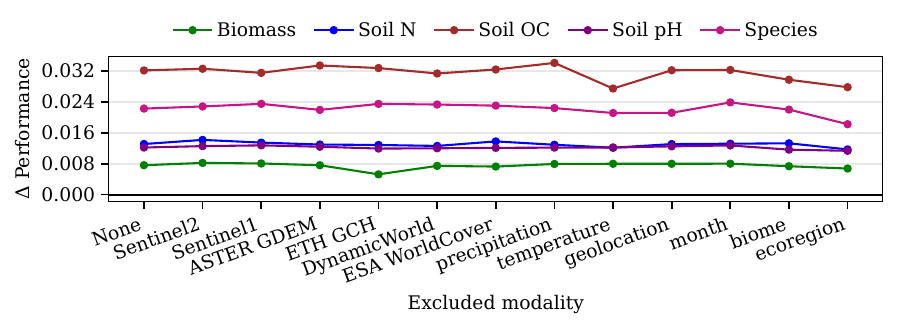}
        \caption{Random split}
    \end{subfigure}
    \vfill
    \begin{subfigure}{\textwidth}
    \centering
        \includegraphics[width=\linewidth]{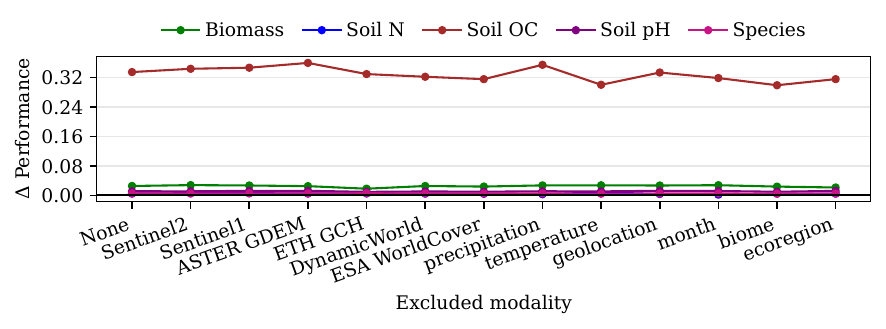}
        \caption{Geographic split}
    \end{subfigure}
    \caption{\textbf{Modality importance for TTT.} Improvement of TTT-MMR over JT for DINOv3 Sat on the random (a) and geographic (b) splits. We hold out each modality at a time \vs ``None,'' i.e., using \emph{all} modalities.}
    \label{fig:leave_one_out}
\end{figure}

We reran TTT-MMR leaving one modality out at a time for one seed and find our method is rather robust and does not rely on only one particular modality per task, as shown in \cref{fig:leave_one_out}. TTT-MMR's gains are not dominated by one modality, but as expected, canopy height is particularly useful for the biomass task.

\FloatBarrier
\subsection{TTT benefit with best model}
We picked the best model for each task and split and found \textsc{TTT-MMR-Geo} still improves in all cases, as shown in \cref{tab:check_ttt_need}.

\begin{table}
\caption{Improvement of \textsc{TTT-MMR-Geo} over joint training (mean $\pm$ standard error across 3 seeds) for the best model for each task and test split at 100\% training data.}
\label{tab:check_ttt_need}
\centering
\begin{tabular}{l|cc|cc}
\toprule
 & \multicolumn{2}{c|}{\textbf{Random}} & \multicolumn{2}{c}{\textbf{Geographic}} \\
\textbf{Task} & Model & $\Delta$ Performance & Model & $\Delta$ Performance \\
\midrule
\textbf{Biomass} & MPMAE & $0.0098 \pm 0.0023$ & MPMAE & $0.0082 \pm 0.0063$ \\
\textbf{Soil N} & TerraMind & $0.0022 \pm 0.0011$ & Copernicus-FM & $0.0123 \pm 0.0099$ \\
\textbf{Soil OC} & Copernicus-FM & $0.00000 \pm 0.00057$ & Copernicus-FM & $0.0012 \pm 0.00079$ \\
\textbf{Soil pH} & TerraMind & $0.0029 \pm 0.00010$ & TerraMind & $0.00068 \pm 0.00098$ \\
\textbf{Species} & Copernicus-FM & $0.00010 \pm 0.00001$ & Copernicus-FM & $0.0046 \pm 0.00028$ \\
\bottomrule
\end{tabular}
\end{table}

\subsection{Modality reconstruction performance for JT}
The absolute modality reconstruction performance for JT is shown in \cref{tab:reconstruction_jt_raw_biomass,tab:reconstruction_jt_raw_soilnitrogen,tab:reconstruction_jt_raw_soilorganiccarbon,tab:reconstruction_jt_raw_soilpH,tab:reconstruction_jt_raw_species}.

\begin{table*}[ht]
\centering
\caption{\textbf{Biomass per-modality reconstruction quality after JT (averaged across seeds 41, 42, 43).} R$^2$ for continuous modalities, accuracy for \texttt{DynamicWorld}, \texttt{ESA\_WorldCover}, \texttt{biome}, \texttt{ecoregion}. }\resizebox{\linewidth}{!}{%
\begin{tabular}{l|cccccccccccc|cccccccccccc}
\toprule
& \multicolumn{12}{c|}{\textbf{Random}} & \multicolumn{12}{c}{\textbf{Geographic}} \\
\textbf{Model} & Sentinel2 & Sentinel1 & \shortstack{ASTER \\ GDEM} & \shortstack{ETH \\ GCH} & DynamicWorld & \shortstack{ESA \\ WorldCover} & precipitation & temperature & \shortstack{geolocation \\ encoding} & \shortstack{month \\ encoding} & biome & ecoregion & Sentinel2 & Sentinel1 & \shortstack{ASTER \\ GDEM} & \shortstack{ETH \\ GCH} & DynamicWorld & \shortstack{ESA \\ WorldCover} & precipitation & temperature & \shortstack{geolocation \\ encoding} & \shortstack{month \\ encoding} & biome & ecoregion \\
\midrule
ConvNeXtV2A & 0.2193 & 0.1541 & 0.0407 & 0.2638 & 0.2166 & 0.1394 & 0.1196 & 0.0123 & -0.1004 & 0.0023 & 0.1550 & 0.0057 & 0.2208 & 0.2166 & -0.7355 & 0.0820 & 0.2252 & 0.1544 & 0.2362 & -1.2793 & -30.3725 & -0.2815 & 0.0930 & 0.0000 \\
Scale-MAE & 0.2575 & 0.1485 & 0.0745 & 0.2604 & 0.2218 & 0.1419 & 0.1277 & 0.0409 & -0.0037 & 0.0052 & 0.1704 & 0.0064 & 0.2674 & 0.2392 & -0.8564 & 0.0276 & 0.2191 & 0.1786 & 0.2177 & -1.5965 & -34.2534 & -0.3042 & 0.0754 & 0.0000 \\
DINOv3 Web & 0.3213 & 0.2382 & 0.3632 & 0.5154 & 0.3648 & 0.2536 & 0.2141 & 0.2668 & 0.1234 & 0.0783 & 0.3297 & 0.0248 & 0.3413 & 0.3086 & -0.9943 & 0.2171 & 0.2998 & 0.2569 & 0.2533 & -1.0140 & -33.8121 & -0.2210 & 0.1136 & 0.0000 \\
DINOv3 Sat & 0.3907 & 0.2828 & 0.4114 & 0.5388 & 0.4417 & 0.3409 & 0.2756 & 0.3804 & 0.2804 & 0.1242 & 0.4400 & 0.0416 & 0.3664 & 0.3244 & -0.5886 & 0.3360 & 0.3790 & 0.3132 & 0.2868 & -0.7216 & -32.3494 & -0.1632 & 0.1560 & 0.0000 \\
SatlasNet & 0.3298 & 0.2086 & 0.2304 & 0.5281 & 0.2860 & 0.1855 & 0.1888 & 0.1340 & 0.1057 & 0.0481 & 0.2732 & 0.0098 & 0.2997 & 0.2834 & -0.3321 & 0.3413 & 0.2706 & 0.2071 & 0.2151 & -1.0881 & -29.4853 & -0.2664 & 0.1414 & 0.0000 \\
MPMAE & 0.3069 & 0.0833 & 0.1789 & 0.3651 & 0.4131 & 0.3452 & 0.1499 & 0.3160 & -1.4759 & -0.1128 & 0.4438 & 0.0698 & 0.4045 & 0.1374 & -1.1817 & 0.2130 & 0.3570 & 0.2924 & 0.0576 & -0.6538 & -43.7792 & -0.4924 & 0.2045 & 0.0000 \\
TerraMind & 0.7173 & 0.5416 & 0.7781 & 0.6602 & 0.5365 & 0.4547 & 0.3686 & 0.6604 & 0.4191 & 0.2260 & 0.5473 & 0.0754 & 0.8680 & 0.5943 & 0.5391 & 0.6739 & 0.4260 & 0.4393 & 0.3905 & 0.1435 & -27.2861 & -0.1513 & 0.2278 & 0.0000 \\
Copernicus-FM & 0.3707 & -0.0498 & 0.2194 & 0.2671 & 0.4017 & 0.3540 & 0.0884 & 0.2266 & -3.9140 & -0.3172 & 0.4509 & 0.0492 & 0.5942 & 0.0884 & -1.0327 & 0.1801 & 0.3356 & 0.3223 & 0.1625 & -0.9592 & -75.4976 & -0.5758 & 0.1758 & 0.0021 \\
Galileo & 0.2011 & 0.2019 & 0.4648 & 0.5313 & 0.3474 & 0.1753 & 0.1616 & 0.2778 & 0.3466 & 0.3311 & 0.3071 & 0.0182 & 0.1126 & 0.1515 & 0.1905 & 0.3771 & 0.2623 & 0.1732 & 0.0890 & -0.8025 & -18.1174 & 0.1984 & 0.1230 & 0.0000 \\
ConvNeXtV2A-MM & 0.2921 & 0.2279 & 0.3815 & 0.5745 & 0.3087 & 0.2158 & 0.2764 & 0.3156 & 0.2335 & 0.2127 & 0.3358 & 0.0152 & 0.2389 & 0.2358 & 0.1244 & 0.5005 & 0.2748 & 0.2479 & 0.1923 & -0.6101 & -23.3217 & 0.0015 & 0.2072 & 0.0000 \\
\bottomrule
\end{tabular}
}
\label{tab:reconstruction_jt_raw_biomass}
\end{table*}
\vspace{2\baselineskip}
\begin{table*}[ht]
\centering
\caption{\textbf{Soil nitrogen per-modality reconstruction quality after JT (averaged across seeds 41, 42, 43).} R$^2$ for continuous modalities, accuracy for \texttt{DynamicWorld}, \texttt{ESA\_WorldCover}, \texttt{biome}, \texttt{ecoregion}. }\resizebox{\linewidth}{!}{%
\begin{tabular}{l|cccccccccccc|cccccccccccc}
\toprule
& \multicolumn{12}{c|}{\textbf{Random}} & \multicolumn{12}{c}{\textbf{Geographic}} \\
\textbf{Model} & Sentinel2 & Sentinel1 & \shortstack{ASTER \\ GDEM} & \shortstack{ETH \\ GCH} & DynamicWorld & \shortstack{ESA \\ WorldCover} & precipitation & temperature & \shortstack{geolocation \\ encoding} & \shortstack{month \\ encoding} & biome & ecoregion & Sentinel2 & Sentinel1 & \shortstack{ASTER \\ GDEM} & \shortstack{ETH \\ GCH} & DynamicWorld & \shortstack{ESA \\ WorldCover} & precipitation & temperature & \shortstack{geolocation \\ encoding} & \shortstack{month \\ encoding} & biome & ecoregion \\
\midrule
ConvNeXtV2A & 0.1524 & 0.2549 & 0.0011 & 0.4218 & 0.1712 & 0.1381 & 0.2627 & 0.1479 & -0.8211 & -0.1731 & 0.1428 & 0.0196 & 0.2738 & 0.2592 & -0.1945 & 0.1985 & 0.1813 & 0.1292 & 0.2588 & -1.3321 & -169.4327 & -0.2757 & 0.1038 & 0.0000 \\
Scale-MAE & 0.5370 & 0.3531 & 0.1590 & 0.4820 & 0.3516 & 0.2519 & 0.2463 & 0.2882 & 0.1133 & 0.0943 & 0.3344 & 0.0872 & 0.2875 & 0.2791 & -0.4845 & -0.0417 & 0.2677 & 0.2188 & 0.0997 & -1.4750 & -136.2411 & -0.5687 & 0.0976 & 0.0000 \\
DINOv3 Web & 0.5953 & 0.4428 & 0.4381 & 0.6667 & 0.4383 & 0.2926 & 0.5168 & 0.6254 & 0.4551 & 0.3264 & 0.4447 & 0.1512 & 0.3393 & 0.4346 & -0.2253 & 0.1346 & 0.3532 & 0.2614 & 0.2550 & -0.4915 & -114.2132 & -0.4070 & 0.1755 & 0.0000 \\
DINOv3 Sat & 0.5766 & 0.4324 & 0.4232 & 0.6653 & 0.4180 & 0.2832 & 0.5213 & 0.6340 & 0.3673 & 0.3472 & 0.4308 & 0.1393 & 0.3764 & 0.3720 & -0.1552 & 0.4170 & 0.3785 & 0.2762 & 0.2639 & -0.1975 & -119.9445 & -0.5131 & 0.1613 & 0.0000 \\
SatlasNet & 0.6039 & 0.3884 & 0.2648 & 0.6283 & 0.3852 & 0.2954 & 0.4196 & 0.5006 & 0.2330 & 0.1920 & 0.3800 & 0.1116 & 0.6090 & 0.2145 & -0.0685 & 0.4824 & 0.3899 & 0.2982 & 0.1854 & -0.1746 & -124.2319 & -0.4933 & 0.1663 & 0.0000 \\
MPMAE & 0.1880 & 0.1844 & 0.1277 & 0.4478 & 0.2840 & 0.2074 & 0.3760 & 0.5030 & -2.7042 & -0.6983 & 0.3833 & 0.0814 & 0.2262 & -0.1404 & -0.4018 & 0.2854 & 0.2354 & 0.1659 & 0.1825 & -0.1150 & -202.5426 & -0.7810 & 0.1855 & 0.0000 \\
TerraMind & 0.8910 & 0.6778 & 0.8212 & 0.8231 & 0.6315 & 0.5318 & 0.7435 & 0.8856 & 0.7708 & 0.5633 & 0.7039 & 0.4156 & 0.8190 & 0.5652 & 0.7079 & 0.6149 & 0.5549 & 0.4283 & 0.2472 & 0.4304 & -93.9853 & -0.5515 & 0.2360 & 0.0000 \\
Copernicus-FM & 0.7044 & 0.5007 & 0.6344 & 0.6615 & 0.4381 & 0.3346 & 0.5921 & 0.7285 & -0.5312 & 0.2501 & 0.5138 & 0.2278 & 0.6109 & 0.4035 & 0.5048 & 0.4806 & 0.4434 & 0.3273 & 0.3061 & -0.2906 & -156.2495 & 0.1253 & 0.1324 & 0.0000 \\
Galileo & 0.2297 & 0.2993 & 0.3952 & 0.5776 & 0.2768 & 0.1592 & 0.3383 & 0.6169 & 0.3076 & 0.4226 & 0.2984 & 0.0377 & 0.2549 & 0.1622 & -0.1547 & 0.4071 & 0.2924 & 0.1912 & 0.1566 & -1.1687 & -131.1033 & 0.2466 & 0.1081 & 0.0000 \\
ConvNeXtV2A-MM & 0.2801 & 0.4560 & 0.4768 & 0.6773 & 0.3048 & 0.1728 & 0.6071 & 0.7497 & 0.3031 & 0.4484 & 0.4229 & 0.0807 & 0.3960 & 0.4102 & -0.0534 & 0.4649 & 0.2591 & 0.1951 & 0.6033 & 0.3322 & -128.6356 & -0.1680 & 0.3001 & 0.0000 \\
\bottomrule
\end{tabular}
}
\label{tab:reconstruction_jt_raw_soilnitrogen}
\end{table*}
\vspace{2\baselineskip}
\begin{table*}[ht]
\centering
\caption{\textbf{Soil organic carbon per-modality reconstruction quality after JT (averaged across seeds 41, 42, 43).} R$^2$ for continuous modalities, accuracy for \texttt{DynamicWorld}, \texttt{ESA\_WorldCover}, \texttt{biome}, \texttt{ecoregion}. }\resizebox{\linewidth}{!}{%
\begin{tabular}{l|cccccccccccc|cccccccccccc}
\toprule
& \multicolumn{12}{c|}{\textbf{Random}} & \multicolumn{12}{c}{\textbf{Geographic}} \\
\textbf{Model} & Sentinel2 & Sentinel1 & \shortstack{ASTER \\ GDEM} & \shortstack{ETH \\ GCH} & DynamicWorld & \shortstack{ESA \\ WorldCover} & precipitation & temperature & \shortstack{geolocation \\ encoding} & \shortstack{month \\ encoding} & biome & ecoregion & Sentinel2 & Sentinel1 & \shortstack{ASTER \\ GDEM} & \shortstack{ETH \\ GCH} & DynamicWorld & \shortstack{ESA \\ WorldCover} & precipitation & temperature & \shortstack{geolocation \\ encoding} & \shortstack{month \\ encoding} & biome & ecoregion \\
\midrule
ConvNeXtV2A & 0.1365 & 0.1430 & 0.0218 & 0.3130 & 0.1724 & 0.1351 & 0.1608 & 0.2380 & -0.4887 & 0.0182 & 0.1422 & 0.0125 & 0.1857 & 0.0659 & -0.0034 & 0.1244 & 0.1411 & 0.1258 & 0.0703 & -0.9086 & -118.2410 & -0.0958 & 0.0853 & 0.0000 \\
Scale-MAE & 0.2154 & 0.2437 & 0.0238 & 0.3410 & 0.1779 & 0.1316 & 0.1902 & 0.1204 & 0.0821 & 0.0517 & 0.1078 & 0.0143 & 0.2497 & 0.2503 & -0.0077 & 0.0056 & 0.1794 & 0.1607 & 0.1832 & -1.5469 & -113.6746 & -0.1634 & 0.0633 & 0.0000 \\
DINOv3 Web & 0.2192 & 0.2556 & 0.0340 & 0.3601 & 0.1791 & 0.1332 & 0.1925 & 0.1296 & 0.1026 & 0.0711 & 0.1059 & 0.0143 & 0.2540 & 0.2296 & 0.0224 & 0.0338 & 0.1687 & 0.1601 & 0.1327 & -1.7859 & -114.6285 & -0.1818 & 0.0645 & 0.0000 \\
DINOv3 Sat & 0.2510 & 0.2683 & 0.0600 & 0.3569 & 0.2024 & 0.1408 & 0.2150 & 0.2567 & 0.1992 & 0.1173 & 0.1290 & 0.0165 & 0.3059 & 0.2473 & -0.0423 & -0.2208 & 0.1837 & 0.1578 & 0.1990 & -1.3572 & -114.9629 & -0.1501 & 0.0621 & 0.0000 \\
SatlasNet & 0.2163 & 0.2248 & 0.0474 & 0.4193 & 0.2018 & 0.1376 & 0.1488 & 0.3503 & 0.1755 & 0.0798 & 0.1569 & 0.0147 & 0.3391 & -0.0223 & -0.0006 & 0.2343 & 0.2169 & 0.1518 & 0.0659 & -0.4670 & -118.3986 & -0.0521 & 0.1134 & 0.0000 \\
MPMAE & 0.2010 & 0.1673 & 0.0681 & 0.4286 & 0.3752 & 0.2610 & 0.3363 & 0.5342 & -1.0590 & -0.3602 & 0.3059 & 0.1047 & 0.4046 & -0.1757 & -0.2120 & 0.2734 & 0.3206 & 0.2013 & 0.1234 & -0.3958 & -154.4180 & -0.9608 & 0.1291 & 0.0000 \\
TerraMind & 0.3864 & 0.4966 & 0.3791 & 0.5295 & 0.2967 & 0.2040 & 0.4053 & 0.6018 & 0.3895 & 0.2675 & 0.2697 & 0.0420 & 0.7005 & 0.4563 & 0.3678 & 0.1657 & 0.2628 & 0.1776 & 0.2184 & -0.5282 & -110.5220 & -0.6268 & 0.1281 & 0.0000 \\
Copernicus-FM & 0.4027 & 0.3281 & 0.3337 & 0.3500 & 0.3675 & 0.2796 & 0.2539 & 0.4975 & -3.4952 & -1.0586 & 0.3246 & 0.0812 & 0.5260 & 0.2498 & 0.1936 & -0.0919 & 0.2755 & 0.2635 & 0.2704 & -0.4243 & -152.2533 & -0.7341 & 0.1108 & 0.0000 \\
Galileo & 0.1536 & 0.2382 & 0.2335 & 0.4970 & 0.2763 & 0.1437 & 0.3108 & 0.6043 & 0.4488 & 0.3093 & 0.2220 & 0.0186 & 0.2112 & 0.1296 & 0.1234 & 0.2166 & 0.2916 & 0.1871 & 0.0037 & -1.7811 & -114.8523 & 0.1544 & 0.0560 & 0.0000 \\
ConvNeXtV2A-MM & 0.0980 & 0.1439 & 0.1300 & 0.3886 & 0.2013 & 0.1331 & 0.1015 & 0.4905 & 0.1503 & 0.1096 & 0.1962 & 0.0199 & 0.0878 & -0.0404 & 0.1197 & 0.3678 & 0.1678 & 0.1410 & -0.0111 & -0.6175 & -116.4112 & -0.0548 & 0.0919 & 0.0000 \\
\bottomrule
\end{tabular}
}
\label{tab:reconstruction_jt_raw_soilorganiccarbon}
\end{table*}
\vspace{2\baselineskip}
\begin{table*}[ht]
\centering
\caption{\textbf{Soil pH per-modality reconstruction quality after JT (averaged across seeds 41, 42, 43).} R$^2$ for continuous modalities, accuracy for \texttt{DynamicWorld}, \texttt{ESA\_WorldCover}, \texttt{biome}, \texttt{ecoregion}. }\resizebox{\linewidth}{!}{%
\begin{tabular}{l|cccccccccccc|cccccccccccc}
\toprule
& \multicolumn{12}{c|}{\textbf{Random}} & \multicolumn{12}{c}{\textbf{Geographic}} \\
\textbf{Model} & Sentinel2 & Sentinel1 & \shortstack{ASTER \\ GDEM} & \shortstack{ETH \\ GCH} & DynamicWorld & \shortstack{ESA \\ WorldCover} & precipitation & temperature & \shortstack{geolocation \\ encoding} & \shortstack{month \\ encoding} & biome & ecoregion & Sentinel2 & Sentinel1 & \shortstack{ASTER \\ GDEM} & \shortstack{ETH \\ GCH} & DynamicWorld & \shortstack{ESA \\ WorldCover} & precipitation & temperature & \shortstack{geolocation \\ encoding} & \shortstack{month \\ encoding} & biome & ecoregion \\
\midrule
ConvNeXtV2A & 0.4964 & 0.2939 & 0.0727 & 0.4714 & 0.2227 & 0.1475 & 0.2845 & 0.2590 & -0.0466 & 0.0614 & 0.1648 & 0.0145 & 0.3739 & 0.2844 & -0.1590 & 0.1802 & 0.1988 & 0.1975 & 0.3234 & -1.0678 & -120.1653 & -0.3718 & 0.0691 & 0.0000 \\
Scale-MAE & 0.6133 & 0.3852 & 0.2291 & 0.5742 & 0.4198 & 0.2667 & 0.3651 & 0.4664 & 0.3957 & 0.1393 & 0.3093 & 0.0911 & 0.3543 & 0.3202 & -0.4426 & -0.1928 & 0.3104 & 0.2447 & 0.1804 & -0.7119 & -146.6355 & -0.5057 & 0.1111 & 0.0000 \\
DINOv3 Web & 0.7787 & 0.4763 & 0.4713 & 0.7270 & 0.5741 & 0.3998 & 0.5336 & 0.7329 & 0.7172 & 0.3676 & 0.4395 & 0.2392 & 0.4317 & 0.3650 & -0.4107 & 0.3799 & 0.5010 & 0.4357 & 0.2446 & 0.0355 & -124.8783 & -0.5094 & 0.1563 & 0.0000 \\
DINOv3 Sat & 0.7713 & 0.4562 & 0.4447 & 0.7270 & 0.5472 & 0.3712 & 0.5367 & 0.7038 & 0.7063 & 0.3224 & 0.4243 & 0.2088 & 0.3560 & 0.3269 & -0.3590 & 0.3571 & 0.4545 & 0.3862 & 0.1907 & 0.0814 & -133.8328 & -0.5166 & 0.1462 & 0.0000 \\
SatlasNet & 0.8568 & 0.4799 & 0.4588 & 0.7340 & 0.5716 & 0.4122 & 0.6082 & 0.7707 & 0.7290 & 0.4746 & 0.5161 & 0.2583 & 0.8236 & 0.3242 & -0.3424 & 0.4902 & 0.4785 & 0.3881 & 0.2728 & 0.2047 & -122.1771 & -0.4334 & 0.1787 & 0.0000 \\
MPMAE & 0.6153 & 0.3576 & 0.3098 & 0.6419 & 0.4052 & 0.2705 & 0.5285 & 0.6434 & -0.3589 & 0.0303 & 0.3675 & 0.0978 & 0.3835 & 0.1736 & -0.1769 & 0.4594 & 0.3432 & 0.2901 & 0.2821 & 0.0057 & -165.4489 & -0.5563 & 0.1913 & 0.0000 \\
TerraMind & 0.9096 & 0.6872 & 0.8528 & 0.8364 & 0.6774 & 0.5432 & 0.7611 & 0.9224 & 0.9017 & 0.6158 & 0.6538 & 0.4675 & 0.8632 & 0.6231 & 0.8097 & 0.6617 & 0.6252 & 0.5059 & 0.3184 & 0.5324 & -96.7463 & -0.4767 & 0.2662 & 0.0000 \\
Copernicus-FM & 0.9145 & 0.6744 & 0.8456 & 0.8212 & 0.6370 & 0.4902 & 0.8036 & 0.9411 & 0.7108 & 0.7783 & 0.7269 & 0.6154 & 0.8866 & 0.6247 & 0.8172 & 0.5987 & 0.4754 & 0.4127 & 0.4770 & 0.3759 & -53.4121 & 0.6305 & 0.2045 & 0.0000 \\
Galileo & 0.6497 & 0.4469 & 0.5956 & 0.7366 & 0.4710 & 0.2813 & 0.6810 & 0.8548 & 0.7775 & 0.5990 & 0.4451 & 0.1045 & 0.4745 & 0.3286 & 0.3220 & 0.4371 & 0.4118 & 0.2613 & 0.4797 & -0.0811 & -148.6093 & 0.5286 & 0.2024 & 0.0000 \\
ConvNeXtV2A-MM & 0.7327 & 0.5808 & 0.7434 & 0.8250 & 0.5464 & 0.3818 & 0.9209 & 0.9383 & 0.7042 & 0.7441 & 0.6804 & 0.4089 & 0.6591 & 0.4524 & 0.5800 & 0.7145 & 0.5121 & 0.4405 & 0.8769 & 0.7521 & -49.5717 & 0.5180 & 0.4617 & 0.0000 \\
\bottomrule
\end{tabular}
}
\label{tab:reconstruction_jt_raw_soilpH}
\end{table*}
\vspace{2\baselineskip}
\begin{table*}[ht]
\centering
\caption{\textbf{Species per-modality reconstruction quality after JT (averaged across seeds 41, 42, 43).} R$^2$ for continuous modalities, accuracy for \texttt{DynamicWorld}, \texttt{ESA\_WorldCover}, \texttt{biome}, \texttt{ecoregion}. }\resizebox{\linewidth}{!}{%
\begin{tabular}{l|cccccccccccc|cccccccccccc}
\toprule
& \multicolumn{12}{c|}{\textbf{Random}} & \multicolumn{12}{c}{\textbf{Geographic}} \\
\textbf{Model} & Sentinel2 & Sentinel1 & \shortstack{ASTER \\ GDEM} & \shortstack{ETH \\ GCH} & DynamicWorld & \shortstack{ESA \\ WorldCover} & precipitation & temperature & \shortstack{geolocation \\ encoding} & \shortstack{month \\ encoding} & biome & ecoregion & Sentinel2 & Sentinel1 & \shortstack{ASTER \\ GDEM} & \shortstack{ETH \\ GCH} & DynamicWorld & \shortstack{ESA \\ WorldCover} & precipitation & temperature & \shortstack{geolocation \\ encoding} & \shortstack{month \\ encoding} & biome & ecoregion \\
\midrule
ConvNeXtV2A & 0.8445 & 0.3161 & 0.2192 & 0.5992 & 0.4483 & 0.3297 & 0.3770 & 0.5166 & 0.3188 & -0.0307 & 0.2981 & 0.0565 & 0.8670 & 0.2846 & -0.0225 & 0.4063 & 0.3708 & 0.3300 & 0.3446 & -0.3975 & -4.6243 & -0.1299 & 0.1187 & 0.0067 \\
Scale-MAE & 0.7852 & 0.4306 & 0.5115 & 0.7227 & 0.5909 & 0.4667 & 0.4783 & 0.6534 & 0.5282 & -0.0395 & 0.4245 & 0.1477 & 0.6201 & 0.3667 & 0.0491 & 0.1942 & 0.4910 & 0.3942 & 0.3678 & -0.1556 & -8.1188 & -0.2758 & 0.1600 & 0.0057 \\
DINOv3 Web & 0.8542 & 0.5440 & 0.6672 & 0.7986 & 0.6600 & 0.5514 & 0.6021 & 0.8022 & 0.7568 & 0.1676 & 0.5578 & 0.2319 & 0.7158 & 0.4801 & 0.2211 & 0.3070 & 0.5538 & 0.4684 & 0.4084 & 0.1350 & -9.6225 & -0.1473 & 0.2294 & 0.0067 \\
DINOv3 Sat & 0.8543 & 0.5574 & 0.6730 & 0.8022 & 0.6596 & 0.5566 & 0.6082 & 0.7974 & 0.7324 & 0.1592 & 0.5707 & 0.2383 & 0.7174 & 0.4916 & 0.2563 & 0.3646 & 0.5467 & 0.4640 & 0.4215 & 0.1282 & -9.2538 & -0.1544 & 0.1798 & 0.0067 \\
SatlasNet & 0.9054 & 0.5104 & 0.6448 & 0.8013 & 0.6747 & 0.5333 & 0.6352 & 0.8370 & 0.7632 & 0.2306 & 0.5822 & 0.2583 & 0.9263 & 0.4430 & 0.1891 & 0.3337 & 0.5423 & 0.4583 & 0.3451 & 0.0794 & -9.6678 & -0.2967 & 0.2026 & 0.0069 \\
MPMAE & 0.9421 & 0.5349 & 0.7112 & 0.8465 & 0.7117 & 0.5528 & 0.6551 & 0.8801 & 0.7929 & 0.2803 & 0.6150 & 0.2953 & 0.9519 & 0.4478 & 0.2626 & 0.5488 & 0.5611 & 0.4433 & 0.3472 & 0.3853 & -8.2678 & -0.2183 & 0.1932 & 0.0042 \\
TerraMind & 0.9424 & 0.7318 & 0.9095 & 0.8725 & 0.7564 & 0.6255 & 0.7295 & 0.9390 & 0.8759 & 0.4534 & 0.6849 & 0.4237 & 0.9448 & 0.7277 & 0.8544 & 0.5146 & 0.6132 & 0.5224 & 0.4904 & 0.4999 & -7.2830 & -0.2025 & 0.2264 & 0.0081 \\
Copernicus-FM & 0.9520 & 0.7295 & 0.9126 & 0.8635 & 0.7182 & 0.5934 & 0.8383 & 0.9715 & 0.9612 & 0.9340 & 0.7658 & 0.5724 & 0.9667 & 0.7297 & 0.8677 & 0.6108 & 0.5423 & 0.4865 & 0.4361 & 0.5388 & -1.6601 & 0.3388 & 0.2661 & 0.0134 \\
Galileo & 0.9462 & 0.7025 & 0.9138 & 0.8620 & 0.8956 & 0.6520 & 0.9217 & 0.9729 & 0.9807 & 0.9759 & 0.7526 & 0.5524 & 0.9366 & 0.6884 & 0.8615 & 0.6193 & 0.8212 & 0.5516 & 0.7485 & 0.4287 & -1.9608 & 0.9537 & 0.2078 & 0.0163 \\
ConvNeXtV2A-MM & 0.9229 & 0.7114 & 0.8794 & 0.9066 & 0.8060 & 0.7373 & 0.9899 & 0.9946 & 0.9652 & 0.9134 & 0.9702 & 0.8924 & 0.9035 & 0.6724 & 0.7970 & 0.8061 & 0.7297 & 0.7322 & 0.9297 & 0.8832 & -0.5333 & 0.7995 & 0.5570 & 0.0221 \\
\bottomrule
\end{tabular}
}
\label{tab:reconstruction_jt_raw_species}
\end{table*}
\vspace{2\baselineskip}

\FloatBarrier
\subsection{Modality reconstruction for TTT-MMR \vs JT}
As expected, TTT-MMR almost always reconstructs the task modalities better than JT, as can be seen for all five tasks in \cref{tab:reconstruction_jt_ttt_minus_jt_biomass,tab:reconstruction_jt_ttt_minus_jt_soilnitrogen,tab:reconstruction_jt_ttt_minus_jt_soilorganiccarbon,tab:reconstruction_jt_ttt_minus_jt_soilpH,tab:reconstruction_jt_ttt_minus_jt_species}.

\begin{table*}[ht]
\centering
\caption{\textbf{Biomass per-modality reconstruction improvement for TTT-MMR over joint training (TTT-MMR $-$ JT).} R$^2$ for continuous modalities, accuracy for \texttt{DynamicWorld}, \texttt{ESA\_WorldCover}, \texttt{biome}, \texttt{ecoregion}. }
\resizebox{\linewidth}{!}{%
\begin{tabular}{l|cccccccccccc|cccccccccccc}
\toprule
& \multicolumn{12}{c|}{\textbf{Random} ($\Delta$ TTT-MMR$-$JT)} & \multicolumn{12}{c}{\textbf{Geographic} ($\Delta$ TTT-MMR$-$JT)} \\
\textbf{Model} & Sentinel2 & Sentinel1 & \shortstack{ASTER \\ GDEM} & \shortstack{ETH \\ GCH} & DynamicWorld & \shortstack{ESA \\ WorldCover} & precipitation & temperature & \shortstack{geolocation \\ encoding} & \shortstack{month \\ encoding} & biome & ecoregion & Sentinel2 & Sentinel1 & \shortstack{ASTER \\ GDEM} & \shortstack{ETH \\ GCH} & DynamicWorld & \shortstack{ESA \\ WorldCover} & precipitation & temperature & \shortstack{geolocation \\ encoding} & \shortstack{month \\ encoding} & biome & ecoregion \\
\midrule
ConvNeXtV2A & 0.0230 & 0.0372 & 0.0377 & 0.0779 & 0.0131 & 0.0039 & 0.0446 & 0.0171 & 0.0117 & 0.0304 & 0.0222 & 0.0002 & 0.0156 & 0.0300 & 0.0891 & 0.1152 & 0.0104 & -0.0132 & 0.0381 & 0.0501 & 1.1228 & 0.0394 & 0.0125 & 0.0000 \\
Scale-MAE & 0.0142 & 0.0239 & 0.0433 & 0.0665 & 0.0105 & 0.0076 & 0.0243 & 0.0316 & 0.0296 & 0.0316 & 0.0153 & 0.0013 & 0.0176 & 0.0206 & 0.2260 & 0.1588 & 0.0104 & 0.0089 & 0.0132 & 0.1340 & 1.4146 & 0.0561 & 0.0061 & 0.0000 \\
DINOv3 Web & 0.0189 & 0.0192 & 0.0468 & 0.0305 & 0.0156 & 0.0138 & 0.0343 & 0.0600 & 0.0496 & 0.0437 & 0.0284 & 0.0026 & 0.0184 & 0.0229 & 0.2161 & 0.0685 & 0.0093 & 0.0118 & 0.0270 & 0.1580 & 1.4671 & 0.0497 & 0.0164 & 0.0000 \\
DINOv3 Sat & 0.0134 & 0.0179 & 0.0311 & 0.0243 & 0.0137 & 0.0168 & 0.0287 & 0.0492 & 0.0390 & 0.0343 & 0.0251 & 0.0040 & 0.0160 & 0.0186 & 0.1102 & 0.0678 & 0.0088 & 0.0106 & 0.0237 & 0.0894 & 0.9696 & 0.0331 & 0.0106 & 0.0000 \\
SatlasNet & 0.0245 & 0.0290 & 0.0401 & 0.0472 & 0.0157 & 0.0139 & 0.0371 & 0.0359 & 0.0271 & 0.0271 & 0.0320 & 0.0004 & 0.0047 & 0.0096 & 0.1515 & 0.1661 & 0.0104 & 0.0157 & 0.0080 & 0.0865 & 0.9863 & 0.0497 & 0.0073 & 0.0000 \\
MPMAE & 0.0472 & 0.0708 & 0.1127 & 0.0960 & 0.0301 & 0.0270 & 0.1243 & 0.1266 & 0.5666 & 0.2195 & 0.0691 & 0.0094 & 0.0435 & 0.0839 & 0.3665 & 0.1480 & 0.0274 & 0.0317 & 0.1356 & 0.1695 & 7.3520 & 0.1961 & 0.0293 & 0.0000 \\
TerraMind & 0.0054 & 0.0061 & 0.0060 & 0.0105 & 0.0088 & 0.0127 & 0.0165 & 0.0252 & 0.0329 & 0.0285 & 0.0193 & 0.0037 & 0.0034 & 0.0086 & 0.0185 & 0.0132 & 0.0057 & 0.0118 & 0.0138 & 0.0488 & 1.0039 & 0.0289 & 0.0102 & 0.0000 \\
Copernicus-FM & 0.0211 & 0.0496 & 0.0587 & 0.0647 & 0.0210 & 0.0182 & 0.0918 & 0.0596 & 0.5332 & 0.1978 & 0.0417 & 0.0137 & 0.0181 & 0.0582 & 0.2265 & 0.1411 & 0.0082 & 0.0118 & 0.0898 & 0.2509 & 17.3486 & 0.3112 & 0.0101 & -0.0001 \\
Galileo & 0.0095 & 0.0073 & 0.0128 & 0.0117 & 0.0077 & 0.0036 & 0.0141 & 0.0189 & 0.0130 & 0.0111 & 0.0100 & 0.0017 & 0.0167 & 0.0181 & 0.0198 & 0.0288 & 0.0099 & 0.0099 & 0.0142 & 0.0268 & 0.2609 & 0.0115 & 0.0088 & 0.0000 \\
ConvNeXtV2A-MM & 0.0431 & 0.0350 & 0.0602 & 0.0358 & 0.0214 & 0.0111 & 0.0567 & 0.0752 & 0.0476 & 0.0505 & 0.0599 & -0.0023 & 0.0367 & 0.0367 & 0.0650 & 0.0805 & 0.0116 & 0.0349 & 0.0482 & 0.1865 & 2.9905 & 0.0747 & 0.0304 & 0.0000 \\
\bottomrule
\end{tabular}
}
\label{tab:reconstruction_jt_ttt_minus_jt_biomass}
\end{table*}
\vspace{2\baselineskip}
\begin{table*}[ht]
\centering
\caption{\textbf{Soil nitrogen per-modality reconstruction improvement for TTT-MMR over joint training (TTT-MMR $-$ JT).} R$^2$ for continuous modalities, accuracy for \texttt{DynamicWorld}, \texttt{ESA\_WorldCover}, \texttt{biome}, \texttt{ecoregion}. }
\resizebox{\linewidth}{!}{%
\begin{tabular}{l|cccccccccccc|cccccccccccc}
\toprule
& \multicolumn{12}{c|}{\textbf{Random} ($\Delta$ TTT-MMR$-$JT)} & \multicolumn{12}{c}{\textbf{Geographic} ($\Delta$ TTT-MMR$-$JT)} \\
\textbf{Model} & Sentinel2 & Sentinel1 & \shortstack{ASTER \\ GDEM} & \shortstack{ETH \\ GCH} & DynamicWorld & \shortstack{ESA \\ WorldCover} & precipitation & temperature & \shortstack{geolocation \\ encoding} & \shortstack{month \\ encoding} & biome & ecoregion & Sentinel2 & Sentinel1 & \shortstack{ASTER \\ GDEM} & \shortstack{ETH \\ GCH} & DynamicWorld & \shortstack{ESA \\ WorldCover} & precipitation & temperature & \shortstack{geolocation \\ encoding} & \shortstack{month \\ encoding} & biome & ecoregion \\
\midrule
ConvNeXtV2A & 0.0457 & 0.0368 & 0.0608 & 0.0546 & 0.0034 & 0.0025 & 0.0397 & 0.0515 & 0.2171 & 0.2408 & 0.0040 & 0.0014 & 0.0519 & 0.0389 & 0.0973 & 0.1287 & -0.0009 & 0.0026 & 0.0172 & 0.1631 & 30.2010 & 0.1231 & 0.0052 & 0.0000 \\
Scale-MAE & 0.0726 & 0.0532 & 0.1840 & 0.1180 & 0.0516 & 0.0298 & 0.1486 & 0.2307 & 0.2100 & 0.1637 & 0.0846 & 0.0451 & 0.0948 & 0.0412 & 0.1837 & 0.1887 & 0.0327 & 0.0205 & 0.0823 & 0.5047 & 12.9439 & 0.1149 & -0.0060 & 0.0000 \\
DINOv3 Web & 0.0357 & 0.0246 & 0.0550 & 0.0392 & 0.0181 & 0.0255 & 0.0816 & 0.0823 & 0.0806 & 0.0651 & 0.0423 & 0.0264 & 0.0546 & 0.0119 & 0.0679 & 0.1299 & 0.0228 & 0.0110 & 0.0497 & 0.1394 & 5.3636 & 0.0765 & 0.0128 & 0.0000 \\
DINOv3 Sat & 0.0217 & 0.0188 & 0.0501 & 0.0318 & 0.0154 & 0.0148 & 0.0493 & 0.0641 & 0.0678 & 0.0514 & 0.0264 & 0.0163 & 0.0464 & 0.0137 & 0.0473 & 0.0597 & 0.0183 & 0.0499 & 0.0311 & 0.0828 & 3.5433 & 0.0749 & 0.0072 & 0.0000 \\
SatlasNet & 0.0076 & 0.0285 & 0.0525 & 0.0677 & 0.0063 & 0.0022 & 0.0444 & 0.0655 & 0.0555 & 0.0551 & 0.0213 & 0.0006 & 0.0180 & 0.0189 & 0.0593 & 0.1248 & 0.0088 & 0.0040 & 0.0494 & 0.1657 & 0.4730 & 0.0797 & -0.0230 & 0.0000 \\
MPMAE & 0.0357 & 0.0866 & 0.1013 & 0.1174 & 0.0194 & 0.0132 & 0.1245 & 0.0792 & 1.0051 & 0.4342 & 0.0240 & 0.0030 & 0.0930 & 0.1070 & 0.1230 & 0.1048 & 0.0077 & 0.0065 & 0.0797 & 0.0999 & 39.4374 & 0.3303 & 0.0144 & 0.0000 \\
TerraMind & 0.0037 & 0.0044 & 0.0046 & 0.0083 & 0.0077 & 0.0054 & 0.0272 & 0.0117 & 0.0236 & 0.0274 & 0.0259 & 0.0089 & 0.0083 & 0.0141 & 0.0055 & 0.0244 & 0.0088 & 0.0052 & 0.0319 & 0.0259 & 4.1809 & 0.0528 & 0.0072 & 0.0000 \\
Copernicus-FM & 0.0228 & 0.0238 & 0.0282 & 0.0301 & 0.0087 & 0.0090 & 0.0393 & 0.0256 & 0.3416 & 0.1586 & 0.0130 & 0.0052 & 0.0380 & 0.0452 & 0.0586 & 0.0476 & 0.0071 & 0.0051 & 0.0387 & 0.1388 & 33.8576 & 0.1269 & 0.0106 & 0.0000 \\
Galileo & 0.0055 & 0.0087 & 0.0078 & 0.0153 & 0.0020 & 0.0128 & 0.0152 & 0.0093 & 0.0178 & 0.0078 & 0.0054 & 0.0012 & 0.0092 & -0.0007 & 0.0128 & 0.0255 & 0.0034 & 0.0027 & 0.0040 & 0.0301 & 1.3096 & -0.0029 & 0.0000 & 0.0000 \\
ConvNeXtV2A-MM & 0.0276 & 0.0221 & 0.0535 & 0.0249 & 0.0140 & 0.0124 & 0.0408 & 0.0266 & 0.0482 & 0.0492 & 0.0193 & 0.0041 & 0.0367 & 0.0170 & 0.0862 & 0.0542 & 0.0188 & 0.0084 & 0.0200 & 0.0388 & 6.9265 & 0.0560 & 0.0103 & 0.0000 \\
\bottomrule
\end{tabular}
}
\label{tab:reconstruction_jt_ttt_minus_jt_soilnitrogen}
\end{table*}
\vspace{2\baselineskip}
\begin{table*}[ht]
\centering
\caption{\textbf{Soil organic carbon per-modality reconstruction improvement for TTT-MMR over joint training (TTT-MMR $-$ JT).} R$^2$ for continuous modalities, accuracy for \texttt{DynamicWorld}, \texttt{ESA\_WorldCover}, \texttt{biome}, \texttt{ecoregion}. }
\resizebox{\linewidth}{!}{%
\begin{tabular}{l|cccccccccccc|cccccccccccc}
\toprule
& \multicolumn{12}{c|}{\textbf{Random} ($\Delta$ TTT-MMR$-$JT)} & \multicolumn{12}{c}{\textbf{Geographic} ($\Delta$ TTT-MMR$-$JT)} \\
\textbf{Model} & Sentinel2 & Sentinel1 & \shortstack{ASTER \\ GDEM} & \shortstack{ETH \\ GCH} & DynamicWorld & \shortstack{ESA \\ WorldCover} & precipitation & temperature & \shortstack{geolocation \\ encoding} & \shortstack{month \\ encoding} & biome & ecoregion & Sentinel2 & Sentinel1 & \shortstack{ASTER \\ GDEM} & \shortstack{ETH \\ GCH} & DynamicWorld & \shortstack{ESA \\ WorldCover} & precipitation & temperature & \shortstack{geolocation \\ encoding} & \shortstack{month \\ encoding} & biome & ecoregion \\
\midrule
ConvNeXtV2A & 0.0354 & 0.0550 & 0.0351 & 0.0855 & 0.0089 & -0.0057 & 0.0613 & 0.0883 & 0.3129 & 0.0842 & 0.0117 & 0.0013 & 0.0524 & 0.0471 & 0.0313 & 0.0973 & 0.0113 & 0.0096 & 0.0661 & 0.1279 & 5.1823 & 0.0621 & 0.0175 & 0.0000 \\
Scale-MAE & 0.0036 & 0.0070 & 0.0026 & 0.0137 & 0.0013 & 0.0009 & 0.0062 & 0.0085 & 0.0093 & 0.0076 & 0.0003 & 0.0000 & 0.0064 & 0.0023 & 0.0077 & 0.0340 & 0.0010 & 0.0010 & 0.0030 & 0.0397 & -0.0237 & 0.0095 & 0.0017 & 0.0000 \\
DINOv3 Web & 0.0122 & 0.0177 & 0.0109 & 0.0374 & 0.0049 & 0.0032 & 0.0161 & 0.0731 & 0.0400 & 0.0182 & 0.0037 & 0.0007 & 0.0138 & -0.0028 & 0.0092 & 0.0655 & 0.0049 & 0.0030 & 0.0081 & 0.2019 & -0.5811 & 0.0264 & 0.0030 & 0.0000 \\
DINOv3 Sat & 0.0060 & 0.0119 & 0.0056 & 0.0345 & 0.0024 & 0.0017 & 0.0136 & 0.0281 & 0.0128 & 0.0070 & 0.0043 & 0.0010 & 0.0073 & 0.0022 & 0.0115 & 0.0746 & 0.0029 & 0.0023 & 0.0048 & 0.1150 & 0.2623 & 0.0107 & 0.0036 & 0.0000 \\
SatlasNet & 0.0043 & 0.0397 & 0.0115 & 0.0547 & 0.0077 & 0.0036 & 0.0325 & 0.0356 & 0.0294 & 0.0244 & 0.0087 & 0.0012 & 0.0003 & 0.0310 & 0.0190 & 0.1536 & 0.0063 & 0.0195 & 0.0187 & 0.0026 & -0.0493 & 0.0363 & 0.0214 & 0.0000 \\
MPMAE & 0.0372 & 0.0573 & 0.0843 & 0.0654 & 0.0210 & 0.0231 & 0.1095 & 0.0983 & 0.3711 & 0.2146 & 0.0242 & 0.0187 & 0.0616 & 0.1051 & 0.1147 & 0.1117 & 0.0155 & 0.0545 & 0.0983 & 0.2498 & 19.1913 & 0.2878 & 0.0104 & 0.0000 \\
TerraMind & 0.0053 & 0.0066 & 0.0100 & 0.0125 & 0.0027 & 0.0031 & 0.0168 & 0.0246 & 0.0255 & 0.0198 & 0.0038 & 0.0010 & 0.0051 & 0.0140 & 0.0105 & 0.0282 & 0.0062 & 0.0024 & 0.0149 & 0.1059 & 0.8447 & 0.0441 & 0.0009 & 0.0000 \\
Copernicus-FM & 0.0565 & 0.1059 & 0.1872 & 0.1936 & 0.0399 & 0.0269 & 0.1498 & 0.0928 & 0.8792 & 0.3423 & 0.0377 & 0.0192 & 0.0771 & 0.1247 & 0.3688 & 0.5551 & 0.0376 & 0.0274 & 0.0926 & 0.7489 & 96.2705 & 0.2196 & 0.0072 & 0.0000 \\
Galileo & 0.0064 & 0.0094 & 0.0097 & 0.0166 & 0.0017 & 0.0017 & 0.0151 & 0.0229 & 0.0210 & 0.0151 & 0.0040 & 0.0020 & 0.0079 & 0.0078 & 0.0051 & 0.0358 & 0.0033 & 0.0026 & 0.0272 & 0.1757 & 0.6237 & 0.0258 & 0.0055 & 0.0000 \\
ConvNeXtV2A-MM & 0.0172 & 0.0235 & 0.0209 & 0.0503 & 0.0025 & 0.0026 & 0.0425 & 0.0157 & 0.0212 & 0.0269 & 0.0108 & 0.0003 & 0.0349 & 0.0606 & 0.0132 & 0.0475 & 0.0069 & 0.0057 & 0.0454 & 0.0229 & 8.9837 & 0.0342 & 0.0014 & 0.0000 \\
\bottomrule
\end{tabular}
}
\label{tab:reconstruction_jt_ttt_minus_jt_soilorganiccarbon}
\end{table*}
\vspace{2\baselineskip}
\begin{table*}[ht]
\centering
\caption{\textbf{Soil pH per-modality reconstruction improvement for TTT-MMR over joint training (TTT-MMR $-$ JT).} R$^2$ for continuous modalities, accuracy for \texttt{DynamicWorld}, \texttt{ESA\_WorldCover}, \texttt{biome}, \texttt{ecoregion}. }
\resizebox{\linewidth}{!}{%
\begin{tabular}{l|cccccccccccc|cccccccccccc}
\toprule
& \multicolumn{12}{c|}{\textbf{Random} ($\Delta$ TTT-MMR$-$JT)} & \multicolumn{12}{c}{\textbf{Geographic} ($\Delta$ TTT-MMR$-$JT)} \\
\textbf{Model} & Sentinel2 & Sentinel1 & \shortstack{ASTER \\ GDEM} & \shortstack{ETH \\ GCH} & DynamicWorld & \shortstack{ESA \\ WorldCover} & precipitation & temperature & \shortstack{geolocation \\ encoding} & \shortstack{month \\ encoding} & biome & ecoregion & Sentinel2 & Sentinel1 & \shortstack{ASTER \\ GDEM} & \shortstack{ETH \\ GCH} & DynamicWorld & \shortstack{ESA \\ WorldCover} & precipitation & temperature & \shortstack{geolocation \\ encoding} & \shortstack{month \\ encoding} & biome & ecoregion \\
\midrule
ConvNeXtV2A & 0.0260 & 0.0329 & 0.0525 & 0.0511 & 0.0276 & 0.0229 & 0.0536 & 0.0513 & 0.1084 & 0.0930 & 0.0252 & 0.0013 & 0.0731 & 0.0069 & 0.1199 & 0.2475 & 0.0241 & 0.0443 & 0.0115 & 0.2865 & 8.2419 & 0.1411 & 0.0069 & 0.0000 \\
Scale-MAE & 0.0775 & 0.0465 & 0.1251 & 0.0789 & 0.0399 & 0.0271 & 0.1072 & 0.1497 & 0.1721 & 0.1646 & 0.0691 & 0.0261 & 0.0818 & 0.0459 & 0.1351 & 0.2264 & 0.0765 & 0.0319 & 0.0662 & 0.3573 & 9.9610 & 0.1492 & 0.0347 & 0.0000 \\
DINOv3 Web & 0.0318 & 0.0339 & 0.0871 & 0.0462 & 0.0332 & 0.0283 & 0.0921 & 0.0847 & 0.0842 & 0.1375 & 0.0789 & 0.0512 & 0.0667 & 0.0410 & 0.1397 & 0.1268 & 0.0187 & 0.0197 & 0.0854 & 0.2205 & 11.0170 & 0.1574 & 0.0175 & 0.0000 \\
DINOv3 Sat & 0.0270 & 0.0329 & 0.0780 & 0.0363 & 0.0248 & 0.0227 & 0.0729 & 0.0756 & 0.0764 & 0.1235 & 0.0878 & 0.0464 & 0.0593 & 0.0401 & 0.1001 & 0.0938 & 0.0205 & 0.0155 & 0.0719 & 0.1375 & 10.4549 & 0.1210 & 0.0249 & 0.0000 \\
SatlasNet & 0.0005 & 0.0116 & 0.0630 & 0.0233 & 0.0115 & 0.0180 & 0.0237 & 0.0442 & 0.0512 & 0.0536 & 0.0585 & 0.0373 & -0.0057 & 0.0585 & 0.1720 & 0.1145 & 0.0016 & 0.0214 & 0.0841 & 0.0656 & 14.7853 & 0.1264 & 0.0133 & 0.0000 \\
MPMAE & 0.0281 & 0.0299 & 0.0564 & 0.0363 & 0.0329 & 0.0264 & 0.0586 & 0.0637 & 0.2060 & 0.1635 & 0.0421 & 0.0090 & 0.0445 & 0.0630 & 0.0731 & 0.0564 & 0.0306 & 0.0128 & 0.0677 & 0.1376 & 15.6839 & 0.1761 & 0.0125 & 0.0000 \\
TerraMind & 0.0036 & 0.0047 & 0.0035 & 0.0115 & 0.0104 & 0.0176 & 0.0337 & 0.0130 & 0.0187 & 0.0667 & 0.0233 & 0.0455 & 0.0069 & 0.0090 & 0.0091 & 0.0272 & 0.0082 & 0.0672 & 0.0549 & 0.0346 & 3.9667 & 0.0966 & 0.0069 & 0.0000 \\
Copernicus-FM & 0.0019 & 0.0074 & 0.0054 & 0.0153 & 0.0158 & 0.0119 & 0.0292 & 0.0104 & 0.0932 & 0.0935 & 0.1213 & 0.0240 & 0.0042 & 0.0136 & 0.0105 & 0.0521 & 0.0100 & 0.0117 & 0.0443 & 0.0513 & 8.8952 & 0.0606 & 0.0070 & 0.0000 \\
Galileo & 0.0128 & 0.0095 & 0.0057 & 0.0073 & -0.0007 & 0.0033 & 0.0105 & 0.0089 & 0.0083 & 0.0123 & 0.0049 & 0.0062 & 0.0184 & 0.0118 & 0.0052 & 0.0295 & 0.0060 & 0.0071 & 0.0198 & 0.0536 & 5.3517 & 0.0061 & 0.0031 & 0.0000 \\
ConvNeXtV2A-MM & 0.0204 & 0.0107 & 0.0143 & 0.0111 & 0.0121 & 0.0045 & 0.0133 & 0.0142 & 0.0414 & 0.0547 & 0.0149 & 0.0112 & 0.0371 & 0.0289 & 0.0568 & 0.0305 & 0.0237 & 0.0215 & 0.0238 & 0.0664 & 9.1575 & 0.0763 & -0.0012 & 0.0000 \\
\bottomrule
\end{tabular}
}
\label{tab:reconstruction_jt_ttt_minus_jt_soilpH}
\end{table*}
\vspace{2\baselineskip}
\begin{table*}[ht]
\centering
\caption{\textbf{Species per-modality reconstruction improvement for TTT-MMR over joint training (TTT-MMR $-$ JT).} R$^2$ for continuous modalities, accuracy for \texttt{DynamicWorld}, \texttt{ESA\_WorldCover}, \texttt{biome}, \texttt{ecoregion}. }
\resizebox{\linewidth}{!}{%
\begin{tabular}{l|cccccccccccc|cccccccccccc}
\toprule
& \multicolumn{12}{c|}{\textbf{Random} ($\Delta$ TTT-MMR$-$JT)} & \multicolumn{12}{c}{\textbf{Geographic} ($\Delta$ TTT-MMR$-$JT)} \\
\textbf{Model} & Sentinel2 & Sentinel1 & \shortstack{ASTER \\ GDEM} & \shortstack{ETH \\ GCH} & DynamicWorld & \shortstack{ESA \\ WorldCover} & precipitation & temperature & \shortstack{geolocation \\ encoding} & \shortstack{month \\ encoding} & biome & ecoregion & Sentinel2 & Sentinel1 & \shortstack{ASTER \\ GDEM} & \shortstack{ETH \\ GCH} & DynamicWorld & \shortstack{ESA \\ WorldCover} & precipitation & temperature & \shortstack{geolocation \\ encoding} & \shortstack{month \\ encoding} & biome & ecoregion \\
\midrule
ConvNeXtV2A & -0.0127 & 0.0186 & 0.1157 & -0.0135 & 0.0045 & 0.0132 & 0.0469 & 0.0591 & 0.0953 & 0.1218 & 0.0243 & 0.0051 & -0.0052 & 0.0740 & 0.0953 & 0.0150 & 0.0204 & 0.0204 & 0.0851 & 0.3405 & 0.3319 & 0.1392 & 0.1127 & 0.0005 \\
Scale-MAE & 0.0421 & 0.0386 & 0.0767 & 0.0447 & 0.0285 & 0.0304 & 0.0847 & 0.0958 & 0.1234 & 0.1794 & 0.0762 & 0.0395 & 0.0571 & 0.0466 & 0.1151 & 0.1305 & 0.0205 & 0.0175 & 0.0503 & 0.2046 & 0.8505 & 0.1217 & 0.0120 & 0.0004 \\
DINOv3 Web & 0.0197 & 0.0245 & 0.0456 & 0.0250 & 0.0272 & 0.0171 & 0.0512 & 0.0566 & 0.0697 & 0.1484 & 0.0705 & 0.0502 & 0.0375 & 0.0307 & 0.0792 & 0.0837 & 0.0178 & 0.0162 & 0.0434 & 0.1294 & 0.8778 & 0.1053 & 0.0257 & 0.0006 \\
DINOv3 Sat & 0.0178 & 0.0229 & 0.0424 & 0.0215 & 0.0169 & 0.0157 & 0.0449 & 0.0510 & 0.0604 & 0.1440 & 0.0461 & 0.0403 & 0.0366 & 0.0286 & 0.0648 & 0.0661 & 0.0141 & 0.0149 & 0.0379 & 0.1240 & 0.7018 & 0.1124 & 0.0246 & 0.0005 \\
SatlasNet & 0.0030 & 0.0233 & 0.0578 & 0.0212 & 0.0203 & 0.0349 & 0.0666 & 0.0512 & 0.0660 & 0.1534 & 0.0592 & 0.0604 & 0.0042 & 0.0415 & 0.0986 & 0.1998 & 0.0085 & 0.0196 & 0.1001 & 0.2204 & 1.2370 & 0.1629 & 0.0161 & 0.0005 \\
MPMAE & 0.0019 & 0.0273 & 0.0286 & 0.0205 & 0.0263 & 0.0394 & 0.1189 & 0.0321 & 0.0585 & 0.1633 & 0.1146 & 0.1141 & 0.0021 & 0.0675 & 0.1247 & 0.1331 & 0.0282 & 0.0956 & 0.1458 & 0.1602 & 1.8773 & 0.2321 & 0.0476 & 0.0037 \\
TerraMind & 0.0031 & 0.0055 & 0.0027 & 0.0095 & 0.0086 & 0.0097 & 0.0298 & 0.0104 & 0.0290 & 0.1010 & 0.0412 & 0.0367 & 0.0043 & 0.0078 & 0.0065 & 0.0584 & 0.0097 & 0.0660 & 0.0343 & 0.0500 & 0.5073 & 0.1137 & 0.0205 & 0.0002 \\
Copernicus-FM & 0.0005 & 0.0044 & 0.0015 & 0.0088 & 0.0096 & 0.0170 & 0.0143 & 0.0043 & 0.0147 & 0.0291 & 0.0174 & 0.0241 & 0.0007 & 0.0057 & 0.0035 & 0.0358 & 0.0085 & 0.0111 & 0.0258 & 0.0460 & 0.2836 & 0.0596 & 0.0321 & 0.0001 \\
Galileo & 0.0008 & 0.0055 & 0.0008 & 0.0030 & 0.0002 & 0.0028 & 0.0044 & 0.0015 & 0.0017 & 0.0044 & 0.0125 & 0.0186 & 0.0024 & 0.0059 & 0.0021 & 0.0108 & 0.0009 & 0.0037 & 0.0057 & 0.0144 & 0.1045 & 0.0038 & -0.0181 & -0.0001 \\
ConvNeXtV2A-MM & 0.0018 & 0.0018 & 0.0018 & 0.0023 & 0.0011 & 0.0089 & 0.0026 & 0.0017 & 0.0120 & 0.0346 & 0.0004 & 0.0002 & 0.0173 & 0.0188 & 0.0157 & 0.0278 & 0.0048 & 0.0056 & 0.0156 & 0.0193 & 0.1881 & 0.0382 & 0.0655 & -0.0003 \\
\bottomrule
\end{tabular}
}
\label{tab:reconstruction_jt_ttt_minus_jt_species}
\end{table*}
\vspace{2\baselineskip}

\FloatBarrier
\subsection{Modality reconstruction visualization}
\begin{figure*}
    \centering
    \includegraphics[width=0.7\linewidth]{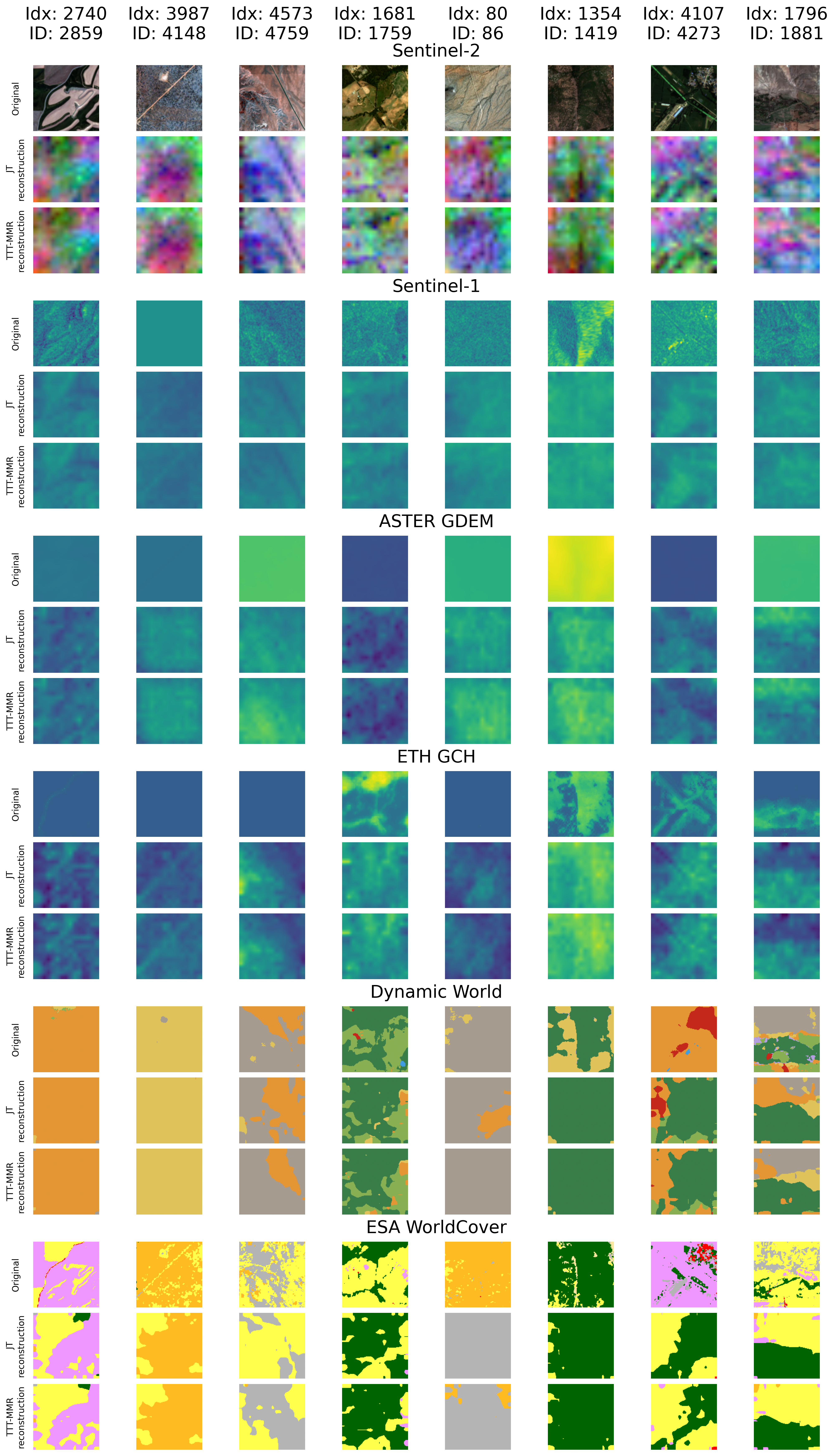}
    \caption{\textbf{Pixel-level modality reconstructions.} MPMAE reconstructions after JT and TTT-MMR as compared to the original. Tiles are from the soil nitrogen random test set.}
    \label{fig:modality_reconstructions}
\end{figure*}

\cref{fig:modality_reconstructions} compares the original pixel-level modalities to their reconstructed forms following JT and TTT-MMR. In particular, we visualize a batch of tiles in the soil nitrogen random test set when run with the MPMAE pretrained model.

\FloatBarrier
\section{Raw results}
For completeness, we provide the raw numbers underlying all the results figures in the main paper and appendix. We bold the result for the best model (vertical comparison).

\subsection{Finetuning}
The ``Random'' section of \cref{tab:FT-biomass,tab:FT-soil_nitrogen,tab:FT-soil_organic_carbon,tab:FT-soil_pH,tab:FT-species} shows the numbers used to create Fig. 4. The ``100\%'' columns in \cref{tab:FT-biomass,tab:FT-soil_nitrogen,tab:FT-soil_organic_carbon,tab:FT-soil_pH,tab:FT-species} were used to generate Fig. 5. We made Fig. 6. based off the ``TerraMind S2'', ``TerraMind'', ``Copernicus-FM S2'', ``Copernicus-FM'', ``Galileo S2,'' and ``Galileo'' rows in \cref{tab:FT-biomass,tab:FT-soil_nitrogen,tab:FT-soil_organic_carbon,tab:FT-soil_pH,tab:FT-species}. The ``Geographic'' section of \cref{tab:FT-biomass,tab:FT-soil_nitrogen,tab:FT-soil_organic_carbon,tab:FT-soil_pH,tab:FT-species} shows the numbers used to create \cref{fig:rq1_FT_geographic}.

\FloatBarrier
\begin{table*}
\centering
\caption{Biomass finetuning test R$^2$}
\resizebox{\linewidth}{!}{%
\begin{tabular}{l|rrr|rrr|rrr|rrr|rrr|rrr}
\toprule
 & \multicolumn{9}{c|}{\textbf{Random}} & \multicolumn{9}{c}{\textbf{Geographic}} \\
 & \multicolumn{3}{c|}{\textbf{Seed 41}} & \multicolumn{3}{c|}{\textbf{Seed 42}} & \multicolumn{3}{c|}{\textbf{Seed 43}} & \multicolumn{3}{c|}{\textbf{Seed 41}} & \multicolumn{3}{c|}{\textbf{Seed 42}} & \multicolumn{3}{c}{\textbf{Seed 43}} \\
 & \textbf{5\%} & \textbf{50\%} & \textbf{100\%} & \textbf{5\%} & \textbf{50\%} & \textbf{100\%} & \textbf{5\%} & \textbf{50\%} & \textbf{100\%} & \textbf{5\%} & \textbf{50\%} & \textbf{100\%} & \textbf{5\%} & \textbf{50\%} & \textbf{100\%} & \textbf{5\%} & \textbf{50\%} & \textbf{100\%} \\
\midrule
\midrule
ConvNeXtV2A & -0.15 & 0.13 & 0.16 & -0.15 & 0.13 & 0.16 & -0.15 & 0.13 & 0.15 & 0.04 & 0.18 & 0.16 & 0.05 & 0.18 & 0.19 & 0.04 & 0.17 & 0.15 \\
Scale-MAE & 0.03 & 0.13 & 0.16 & 0.02 & 0.13 & 0.16 & 0.01 & 0.12 & 0.17 & 0.06 & -0.01 & 0.13 & 0.04 & 0.05 & 0.10 & 0.06 & 0.00 & 0.08 \\
DINOv3 Web & 0.01 & 0.32 & 0.38 & 0.01 & 0.31 & 0.38 & 0.01 & 0.32 & 0.38 & 0.07 & 0.29 & 0.26 & 0.08 & 0.23 & 0.35 & 0.08 & 0.26 & 0.36 \\
DINOv3 Sat & 0.02 & 0.33 & 0.40 & 0.00 & 0.34 & 0.39 & 0.01 & 0.35 & 0.40 & 0.11 & 0.33 & 0.39 & 0.11 & 0.32 & 0.41 & 0.12 & 0.35 & 0.40 \\
SatlasNet & 0.22 & 0.36 & 0.39 & 0.21 & 0.35 & 0.38 & 0.22 & 0.35 & 0.38 & 0.22 & 0.30 & 0.37 & 0.24 & 0.30 & 0.37 & 0.23 & 0.30 & 0.34 \\
MPMAE & \textbf{0.36} & 0.42 & \textbf{0.45} & \textbf{0.36} & 0.40 & \textbf{0.45} & \textbf{0.36} & 0.41 & \textbf{0.45} & \textbf{0.37} & 0.46 & 0.48 & \textbf{0.42} & 0.43 & 0.47 & \textbf{0.39} & 0.45 & 0.46 \\
TerraMind S2 & -0.04 & 0.34 & 0.41 & -0.03 & 0.35 & 0.42 & -0.03 & 0.35 & 0.41 & 0.12 & 0.42 & 0.44 & 0.14 & 0.37 & 0.46 & 0.13 & 0.40 & 0.44 \\
TerraMind & -0.05 & 0.34 & 0.43 & -0.04 & 0.37 & \textbf{0.45} & -0.04 & 0.39 & 0.42 & 0.12 & 0.40 & 0.48 & 0.14 & 0.44 & 0.49 & 0.14 & 0.46 & 0.45 \\
Copernicus-FM S2 & 0.21 & 0.37 & 0.41 & 0.19 & 0.36 & 0.41 & 0.21 & 0.37 & 0.41 & 0.28 & 0.35 & 0.40 & 0.24 & 0.40 & 0.38 & 0.28 & 0.39 & 0.43 \\
Copernicus-FM & 0.34 & \textbf{0.44} & 0.44 & 0.33 & \textbf{0.43} & 0.44 & 0.34 & \textbf{0.43} & 0.44 & 0.36 & 0.45 & 0.43 & 0.38 & 0.44 & 0.45 & 0.33 & 0.46 & \textbf{0.49} \\
Galileo S2 & -0.16 & 0.17 & 0.23 & -0.15 & 0.18 & 0.22 & -0.15 & 0.17 & 0.22 & -0.00 & 0.21 & 0.25 & 0.00 & 0.22 & 0.26 & 0.00 & 0.19 & 0.26 \\
Galileo & -0.14 & 0.39 & 0.43 & -0.14 & 0.40 & 0.43 & -0.14 & 0.40 & 0.42 & 0.07 & 0.38 & 0.32 & 0.05 & 0.17 & 0.24 & 0.07 & 0.37 & 0.27 \\
ConvNeXtV2A-MM & -0.14 & 0.41 & 0.44 & -0.14 & 0.40 & 0.44 & -0.14 & 0.41 & 0.44 & 0.06 & \textbf{0.48} & \textbf{0.49} & 0.07 & \textbf{0.47} & \textbf{0.51} & 0.07 & \textbf{0.47} & \textbf{0.49} \\
\bottomrule
\end{tabular}

}
\label{tab:FT-biomass}
\end{table*}

\begin{table*}
\centering
\caption{Soil nitrogen finetuning test R$^2$}
\resizebox{\linewidth}{!}{%
\begin{tabular}{l|rrr|rrr|rrr|rrr|rrr|rrr}
\toprule
 & \multicolumn{9}{c|}{\textbf{Random}} & \multicolumn{9}{c}{\textbf{Geographic}} \\
 & \multicolumn{3}{c|}{\textbf{Seed 41}} & \multicolumn{3}{c|}{\textbf{Seed 42}} & \multicolumn{3}{c|}{\textbf{Seed 43}} & \multicolumn{3}{c|}{\textbf{Seed 41}} & \multicolumn{3}{c|}{\textbf{Seed 42}} & \multicolumn{3}{c}{\textbf{Seed 43}} \\
 & \textbf{5\%} & \textbf{50\%} & \textbf{100\%} & \textbf{5\%} & \textbf{50\%} & \textbf{100\%} & \textbf{5\%} & \textbf{50\%} & \textbf{100\%} & \textbf{5\%} & \textbf{50\%} & \textbf{100\%} & \textbf{5\%} & \textbf{50\%} & \textbf{100\%} & \textbf{5\%} & \textbf{50\%} & \textbf{100\%} \\
\midrule
\midrule
ConvNeXtV2A & 0.18 & 0.22 & 0.22 & 0.18 & 0.20 & 0.20 & 0.18 & 0.21 & 0.22 & -0.11 & -0.19 & -0.07 & -0.11 & -0.15 & -0.12 & -0.10 & -0.04 & -0.24 \\
Scale-MAE & 0.10 & 0.18 & 0.16 & 0.08 & 0.19 & 0.19 & 0.10 & 0.17 & 0.20 & -0.12 & -0.19 & -0.49 & -0.13 & -0.40 & -0.40 & -0.11 & -0.52 & -0.28 \\
DINOv3 Web & 0.14 & 0.34 & 0.39 & 0.18 & 0.35 & 0.36 & 0.19 & 0.33 & 0.38 & -0.07 & -0.22 & -0.17 & -0.30 & -0.25 & -0.05 & -0.02 & -0.30 & -0.12 \\
DINOv3 Sat & 0.23 & 0.34 & 0.37 & 0.21 & 0.34 & 0.38 & 0.22 & 0.33 & 0.38 & 0.01 & -0.09 & 0.01 & 0.03 & -0.02 & 0.00 & 0.02 & -0.04 & \textbf{-0.00} \\
SatlasNet & 0.17 & 0.25 & 0.31 & 0.19 & 0.30 & 0.31 & 0.18 & 0.27 & 0.32 & -0.19 & -0.18 & -0.10 & -0.12 & -0.12 & -0.12 & -0.09 & -0.17 & -0.18 \\
MPMAE & 0.29 & 0.40 & 0.42 & 0.30 & 0.41 & 0.45 & 0.29 & 0.41 & 0.38 & -0.05 & -0.04 & -0.08 & -0.03 & 0.05 & -0.08 & -0.03 & 0.01 & -0.03 \\
TerraMind S2 & \textbf{0.40} & 0.43 & 0.48 & 0.36 & \textbf{0.46} & 0.47 & 0.37 & 0.42 & \textbf{0.50} & -0.04 & -0.02 & 0.01 & -0.06 & -0.01 & -0.04 & -0.04 & -0.08 & -0.04 \\
TerraMind & 0.33 & \textbf{0.46} & \textbf{0.50} & 0.37 & 0.44 & \textbf{0.51} & 0.37 & \textbf{0.48} & \textbf{0.50} & -0.03 & -0.10 & -0.08 & -0.05 & -0.05 & -0.11 & 0.01 & -0.11 & -0.09 \\
Copernicus-FM S2 & 0.25 & 0.34 & 0.31 & 0.25 & 0.33 & 0.34 & 0.25 & 0.34 & 0.35 & -0.09 & -0.05 & -0.08 & -0.13 & -0.02 & -0.11 & -0.11 & -0.18 & -0.09 \\
Copernicus-FM & 0.38 & 0.44 & 0.44 & \textbf{0.39} & \textbf{0.46} & 0.46 & 0.39 & 0.46 & 0.48 & -0.05 & 0.02 & \textbf{0.04} & -0.02 & 0.01 & \textbf{0.08} & -0.01 & \textbf{0.06} & -0.13 \\
Galileo S2 & 0.26 & 0.27 & 0.34 & 0.19 & 0.32 & 0.32 & 0.24 & 0.32 & 0.28 & -0.08 & -0.08 & 0.03 & -0.19 & 0.02 & 0.00 & -0.26 & -0.02 & -0.01 \\
Galileo & 0.38 & 0.45 & 0.44 & 0.38 & 0.45 & 0.43 & \textbf{0.40} & 0.45 & 0.43 & \textbf{0.14} & \textbf{0.03} & -0.05 & 0.12 & \textbf{0.07} & 0.01 & \textbf{0.11} & 0.03 & -0.07 \\
ConvNeXtV2A-MM & 0.39 & 0.45 & 0.47 & \textbf{0.39} & 0.44 & 0.48 & 0.38 & 0.46 & 0.47 & 0.13 & \textbf{0.03} & 0.02 & \textbf{0.14} & -0.02 & 0.03 & \textbf{0.11} & -0.00 & \textbf{0.00} \\
\bottomrule
\end{tabular}
}
\label{tab:FT-soil_nitrogen}
\end{table*}

\begin{table*}
\centering
\caption{Soil organic carbon finetuning test R$^2$}
\resizebox{\linewidth}{!}{%
\begin{tabular}{l|rrr|rrr|rrr|rrr|rrr|rrr}
\toprule
 & \multicolumn{9}{c|}{\textbf{Random}} & \multicolumn{9}{c}{\textbf{Geographic}} \\
 & \multicolumn{3}{c|}{\textbf{Seed 41}} & \multicolumn{3}{c|}{\textbf{Seed 42}} & \multicolumn{3}{c|}{\textbf{Seed 43}} & \multicolumn{3}{c|}{\textbf{Seed 41}} & \multicolumn{3}{c|}{\textbf{Seed 42}} & \multicolumn{3}{c}{\textbf{Seed 43}} \\
 & \textbf{5\%} & \textbf{50\%} & \textbf{100\%} & \textbf{5\%} & \textbf{50\%} & \textbf{100\%} & \textbf{5\%} & \textbf{50\%} & \textbf{100\%} & \textbf{5\%} & \textbf{50\%} & \textbf{100\%} & \textbf{5\%} & \textbf{50\%} & \textbf{100\%} & \textbf{5\%} & \textbf{50\%} & \textbf{100\%} \\
\midrule
\midrule
ConvNeXtV2A & -0.26 & 0.14 & 0.21 & -0.26 & 0.14 & 0.18 & -0.26 & 0.14 & 0.19 & -0.01 & -0.50 & -0.72 & -0.02 & -0.37 & -0.31 & -0.02 & -0.17 & -0.48 \\
Scale-MAE & \textbf{-0.09} & 0.09 & 0.13 & \textbf{-0.09} & 0.06 & 0.10 & -0.10 & 0.09 & 0.10 & -0.05 & -1.55 & -2.21 & -0.13 & -2.11 & -1.90 & -0.04 & -1.43 & -2.52 \\
DINOv3 Web & -0.10 & 0.10 & 0.11 & \textbf{-0.09} & 0.08 & 0.12 & \textbf{-0.09} & 0.09 & 0.10 & 0.02 & -1.74 & -1.92 & -0.12 & -1.58 & -2.72 & -0.12 & -1.77 & -1.97 \\
DINOv3 Sat & -0.10 & 0.08 & 0.05 & -0.10 & 0.09 & 0.08 & -0.11 & 0.06 & 0.12 & 0.08 & -2.54 & -2.22 & 0.05 & -1.56 & -2.08 & 0.07 & -1.58 & -2.26 \\
SatlasNet & -0.10 & 0.18 & 0.30 & -0.10 & 0.30 & 0.31 & -0.10 & 0.28 & 0.31 & -0.12 & -0.32 & -0.03 & -0.12 & -0.09 & -0.09 & -0.12 & -0.21 & -0.32 \\
MPMAE & -0.26 & 0.22 & 0.40 & -0.26 & 0.23 & 0.40 & -0.26 & 0.24 & 0.40 & -0.01 & \textbf{0.04} & 0.01 & -0.01 & \textbf{0.14} & -0.12 & -0.01 & 0.02 & -0.13 \\
TerraMind S2 & -0.13 & 0.34 & 0.31 & -0.14 & 0.31 & 0.31 & -0.14 & 0.33 & 0.28 & 0.06 & -0.01 & 0.04 & 0.06 & -0.03 & 0.02 & 0.07 & \textbf{0.04} & 0.02 \\
TerraMind & -0.14 & 0.30 & 0.33 & -0.14 & 0.34 & 0.34 & -0.15 & 0.33 & 0.33 & 0.10 & -0.68 & -0.20 & 0.07 & -0.35 & -0.25 & 0.07 & -0.52 & -0.32 \\
Copernicus-FM S2 & -0.15 & 0.28 & 0.30 & -0.15 & 0.25 & 0.31 & -0.14 & 0.31 & 0.27 & 0.05 & -0.30 & -0.23 & 0.07 & -0.00 & -0.05 & 0.06 & -0.06 & -0.10 \\
Copernicus-FM & -0.14 & \textbf{0.41} & 0.45 & -0.14 & \textbf{0.44} & \textbf{0.50} & -0.14 & 0.39 & 0.47 & 0.07 & -0.10 & \textbf{0.09} & 0.06 & 0.04 & \textbf{0.08} & 0.07 & 0.02 & \textbf{0.05} \\
Galileo S2 & -0.14 & 0.25 & 0.25 & -0.15 & 0.24 & 0.28 & -0.15 & 0.23 & 0.28 & 0.07 & -0.48 & -0.21 & 0.07 & -0.28 & -0.11 & 0.08 & -0.22 & -0.50 \\
Galileo & -0.13 & 0.40 & \textbf{0.49} & -0.13 & 0.43 & 0.45 & -0.13 & \textbf{0.43} & \textbf{0.51} & \textbf{0.11} & -1.54 & -1.46 & \textbf{0.14} & -1.54 & -1.22 & \textbf{0.14} & -1.91 & -1.30 \\
ConvNeXtV2A-MM & -0.26 & 0.25 & 0.43 & -0.26 & 0.25 & 0.47 & -0.26 & 0.25 & 0.43 & -0.04 & -0.18 & -1.80 & -0.03 & -0.32 & -1.11 & -0.03 & -0.40 & -0.51 \\
\bottomrule
\end{tabular}

}
\label{tab:FT-soil_organic_carbon}
\end{table*}

\begin{table*}
\centering
\caption{Soil pH finetuning test R$^2$}
\resizebox{\linewidth}{!}{%
\begin{tabular}{l|rrr|rrr|rrr|rrr|rrr|rrr}
\toprule
 & \multicolumn{9}{c|}{\textbf{Random}} & \multicolumn{9}{c}{\textbf{Geographic}} \\
 & \multicolumn{3}{c|}{\textbf{Seed 41}} & \multicolumn{3}{c|}{\textbf{Seed 42}} & \multicolumn{3}{c|}{\textbf{Seed 43}} & \multicolumn{3}{c|}{\textbf{Seed 41}} & \multicolumn{3}{c|}{\textbf{Seed 42}} & \multicolumn{3}{c}{\textbf{Seed 43}} \\
 & \textbf{5\%} & \textbf{50\%} & \textbf{100\%} & \textbf{5\%} & \textbf{50\%} & \textbf{100\%} & \textbf{5\%} & \textbf{50\%} & \textbf{100\%} & \textbf{5\%} & \textbf{50\%} & \textbf{100\%} & \textbf{5\%} & \textbf{50\%} & \textbf{100\%} & \textbf{5\%} & \textbf{50\%} & \textbf{100\%} \\
\midrule
\midrule
ConvNeXtV2A & 0.37 & 0.38 & 0.39 & 0.37 & 0.38 & 0.42 & 0.37 & 0.39 & 0.37 & -0.29 & -0.08 & -0.03 & -0.16 & -0.03 & -0.01 & -0.22 & -0.03 & -0.04 \\
Scale-MAE & 0.28 & 0.37 & 0.40 & 0.29 & 0.38 & 0.39 & 0.29 & 0.36 & 0.39 & -0.23 & -0.21 & -0.07 & -0.02 & -0.12 & -0.04 & -0.03 & -0.15 & -0.10 \\
DINOv3 Web & 0.40 & 0.51 & 0.56 & 0.41 & 0.50 & 0.55 & 0.41 & 0.50 & 0.55 & -0.03 & 0.13 & 0.18 & -0.04 & 0.13 & 0.18 & -0.05 & 0.11 & 0.18 \\
DINOv3 Sat & 0.38 & 0.49 & 0.54 & 0.37 & 0.49 & 0.55 & 0.38 & 0.50 & 0.54 & -0.04 & 0.08 & 0.16 & -0.06 & 0.09 & 0.17 & -0.04 & 0.09 & 0.16 \\
SatlasNet & 0.40 & 0.47 & 0.52 & 0.41 & 0.45 & 0.50 & 0.42 & 0.48 & 0.50 & -0.20 & 0.22 & 0.29 & 0.14 & 0.24 & 0.20 & 0.12 & \textbf{0.31} & 0.23 \\
MPMAE & 0.40 & 0.55 & 0.57 & 0.41 & 0.53 & 0.56 & 0.39 & 0.54 & 0.58 & -0.15 & 0.16 & 0.21 & -0.08 & 0.13 & 0.27 & -0.11 & 0.19 & 0.22 \\
TerraMind S2 & 0.50 & 0.61 & 0.66 & 0.50 & 0.61 & 0.65 & 0.48 & 0.61 & 0.64 & 0.22 & 0.22 & 0.35 & \textbf{0.27} & \textbf{0.26} & 0.32 & 0.05 & 0.25 & 0.32 \\
TerraMind & 0.52 & \textbf{0.62} & \textbf{0.67} & 0.51 & \textbf{0.62} & \textbf{0.67} & 0.51 & \textbf{0.63} & \textbf{0.67} & \textbf{0.24} & 0.24 & \textbf{0.36} & 0.18 & 0.23 & \textbf{0.36} & \textbf{0.30} & 0.27 & \textbf{0.34} \\
Copernicus-FM S2 & 0.46 & 0.51 & 0.55 & 0.46 & 0.50 & 0.56 & 0.47 & 0.50 & 0.55 & 0.10 & 0.17 & 0.26 & 0.11 & 0.17 & 0.26 & 0.11 & 0.14 & 0.26 \\
Copernicus-FM & 0.50 & 0.56 & 0.63 & 0.50 & 0.57 & 0.62 & 0.51 & 0.57 & 0.62 & 0.16 & 0.05 & 0.17 & 0.14 & 0.08 & 0.15 & 0.19 & 0.05 & 0.15 \\
Galileo S2 & 0.43 & 0.49 & 0.53 & 0.43 & 0.48 & 0.53 & 0.44 & 0.49 & 0.49 & 0.09 & \textbf{0.26} & 0.32 & 0.07 & 0.21 & 0.29 & 0.11 & 0.24 & 0.21 \\
Galileo & 0.52 & 0.60 & 0.64 & 0.52 & 0.61 & 0.63 & 0.54 & 0.61 & 0.63 & 0.12 & 0.19 & 0.17 & 0.19 & 0.23 & 0.24 & 0.12 & 0.24 & 0.18 \\
ConvNeXtV2A-MM & \textbf{0.55} & \textbf{0.62} & 0.64 & \textbf{0.56} & 0.60 & 0.65 & \textbf{0.55} & 0.61 & 0.65 & -2.60 & -0.43 & -0.15 & -2.32 & -0.20 & -0.12 & -2.33 & -0.52 & -0.27 \\
\bottomrule
\end{tabular}

}
\label{tab:FT-soil_pH}
\end{table*}

\begin{table*}
\centering
\caption{Species finetuning test mAP}
\resizebox{\linewidth}{!}{%
\begin{tabular}{l|rrr|rrr|rrr|rrr|rrr|rrr}
\toprule
 & \multicolumn{9}{c|}{\textbf{Random}} & \multicolumn{9}{c}{\textbf{Geographic}} \\
 & \multicolumn{3}{c|}{\textbf{Seed 41}} & \multicolumn{3}{c|}{\textbf{Seed 42}} & \multicolumn{3}{c|}{\textbf{Seed 43}} & \multicolumn{3}{c|}{\textbf{Seed 41}} & \multicolumn{3}{c|}{\textbf{Seed 42}} & \multicolumn{3}{c}{\textbf{Seed 43}} \\
 & \textbf{5\%} & \textbf{50\%} & \textbf{100\%} & \textbf{5\%} & \textbf{50\%} & \textbf{100\%} & \textbf{5\%} & \textbf{50\%} & \textbf{100\%} & \textbf{5\%} & \textbf{50\%} & \textbf{100\%} & \textbf{5\%} & \textbf{50\%} & \textbf{100\%} & \textbf{5\%} & \textbf{50\%} & \textbf{100\%} \\
\midrule
\midrule
ConvNeXtV2A & 0.37 & 0.45 & 0.47 & 0.37 & 0.43 & 0.48 & 0.38 & 0.45 & 0.48 & 0.30 & 0.31 & 0.32 & 0.30 & 0.31 & 0.32 & 0.30 & 0.30 & 0.32 \\
Scale-MAE & 0.43 & 0.61 & 0.67 & 0.43 & 0.60 & 0.67 & 0.44 & 0.61 & 0.67 & 0.29 & 0.32 & 0.33 & 0.29 & 0.32 & 0.33 & 0.29 & 0.32 & 0.34 \\
DINOv3 Web & 0.56 & 0.77 & 0.82 & 0.56 & 0.76 & 0.82 & 0.56 & 0.77 & 0.82 & 0.34 & 0.37 & 0.36 & 0.35 & 0.37 & 0.36 & 0.35 & 0.36 & 0.36 \\
DINOv3 Sat & 0.58 & 0.77 & 0.82 & 0.58 & 0.77 & 0.83 & 0.58 & 0.78 & 0.83 & 0.34 & 0.36 & 0.36 & 0.34 & 0.36 & 0.37 & 0.34 & 0.36 & 0.36 \\
SatlasNet & 0.61 & 0.78 & 0.84 & 0.62 & 0.79 & 0.84 & 0.62 & 0.78 & 0.84 & 0.34 & 0.35 & 0.36 & 0.34 & 0.35 & 0.36 & 0.34 & 0.35 & 0.36 \\
MPMAE & 0.70 & 0.87 & 0.90 & 0.71 & 0.87 & 0.90 & 0.70 & 0.87 & 0.90 & 0.38 & 0.38 & 0.39 & 0.38 & 0.39 & 0.39 & 0.38 & 0.38 & 0.38 \\
TerraMind S2 & 0.80 & 0.91 & 0.93 & 0.80 & 0.91 & 0.93 & 0.80 & 0.91 & 0.93 & 0.41 & 0.42 & 0.41 & 0.41 & 0.41 & 0.41 & 0.41 & 0.42 & 0.41 \\
TerraMind & 0.83 & 0.93 & 0.95 & 0.83 & 0.93 & 0.95 & 0.82 & 0.93 & 0.95 & 0.41 & 0.41 & 0.41 & 0.41 & 0.41 & 0.41 & 0.41 & 0.41 & 0.40 \\
Copernicus-FM S2 & 0.65 & 0.83 & 0.88 & 0.66 & 0.84 & 0.88 & 0.66 & 0.84 & 0.87 & 0.34 & 0.38 & 0.37 & 0.34 & 0.37 & 0.38 & 0.34 & 0.37 & 0.39 \\
Copernicus-FM & 0.92 & \textbf{0.99} & \textbf{0.99} & 0.91 & \textbf{0.99} & \textbf{0.99} & 0.91 & \textbf{0.99} & \textbf{0.99} & \textbf{0.47} & \textbf{0.46} & \textbf{0.45} & 0.47 & \textbf{0.45} & 0.43 & \textbf{0.47} & \textbf{0.45} & \textbf{0.44} \\
Galileo S2 & 0.63 & 0.80 & 0.85 & 0.63 & 0.80 & 0.85 & 0.63 & 0.80 & 0.85 & 0.35 & 0.37 & 0.37 & 0.35 & 0.37 & 0.37 & 0.35 & 0.37 & 0.38 \\
Galileo & \textbf{0.94} & 0.98 & \textbf{0.99} & \textbf{0.94} & 0.98 & \textbf{0.99} & \textbf{0.94} & 0.98 & \textbf{0.99} & 0.46 & 0.44 & 0.43 & \textbf{0.48} & 0.44 & \textbf{0.44} & 0.46 & 0.43 & 0.42 \\
ConvNeXtV2A-MM & 0.89 & 0.98 & \textbf{0.99} & 0.89 & 0.98 & \textbf{0.99} & 0.89 & 0.98 & \textbf{0.99} & 0.45 & 0.43 & 0.41 & 0.45 & 0.43 & 0.42 & 0.46 & 0.43 & 0.42 \\
\bottomrule
\end{tabular}

}
\label{tab:FT-species}
\end{table*}

\FloatBarrier
\subsection{Joint training and test-time training}
The ``JT'', ``TTT-MMR'', and ``TTT-MMR-Geo'' columns in \crefrange{tab:biomass_R2_seed41}{tab:species_mAP_seed43} were used to create Fig. 7. Fig. 8 was generated only with the seed 42 results. We additionally underline the result for the best method (horizontal comparison).

\begin{table}
\centering
\caption{Biomass test R$^2$ by architecture, adaptation mode, and split (Seed 41)}
\label{tab:biomass_R2_seed41}
\resizebox{\linewidth}{!}{%
\begin{tabular}{l|cccc|cccc}
\toprule
 & \multicolumn{4}{c|}{\textbf{Random}} & \multicolumn{4}{c}{\textbf{Geographic}} \\
\textbf{Model} & FT & JT & TTT-MMR & TTT-MMR-Geo & FT & JT & TTT-MMR & TTT-MMR-Geo \\
\midrule\midrule
ConvNeXtV2A & 0.16 & 0.16 & \underline{0.24} & \underline{0.24} & 0.16 & 0.13 & \underline{0.28} & \underline{0.28} \\
Scale-MAE & 0.16 & 0.16 & \underline{0.19} & 0.18 & 0.13 & 0.15 & \underline{0.20} & \underline{0.20} \\
DINOv3 Web & 0.38 & 0.38 & \underline{0.39} & \underline{0.39} & 0.26 & 0.33 & 0.37 & \underline{0.38} \\
DINOv3 Sat & \underline{0.40} & \underline{0.40} & \underline{0.40} & \underline{0.40} & 0.39 & 0.41 & \underline{0.42} & \underline{0.42} \\
SatlasNet & 0.39 & 0.38 & \underline{0.40} & \underline{0.40} & 0.37 & 0.35 & \underline{0.42} & \underline{0.42} \\
MPMAE & \textbf{0.45} & 0.44 & \underline{\textbf{0.46}} & \underline{\textbf{0.46}} & 0.48 & 0.48 & \underline{0.49} & 0.48 \\
TerraMind & 0.43 & 0.44 & \underline{0.45} & \underline{0.45} & 0.48 & 0.48 & \underline{0.49} & \underline{0.49} \\
Copernicus-FM & 0.44 & \textbf{0.45} & \underline{\textbf{0.46}} & \underline{\textbf{0.46}} & 0.43 & \underline{\textbf{0.49}} & \underline{0.49} & \underline{0.49} \\
Galileo & 0.43 & 0.43 & \underline{0.44} & 0.43 & \underline{0.32} & 0.13 & 0.15 & 0.15 \\
ConvNeXtV2A-MM & 0.44 & \underline{\textbf{0.45}} & \underline{0.45} & \underline{0.45} & \textbf{0.49} & \textbf{0.49} & \underline{\textbf{0.50}} & \underline{\textbf{0.50}} \\
\bottomrule
\end{tabular}

}
\end{table}

\begin{table}
\centering
\caption{Biomass test R$^2$ by architecture, adaptation mode, and split (Seed 42)}
\label{tab:biomass_R2_seed42}
\resizebox{\linewidth}{!}{%
\begin{tabular}{l|cccc|cccc}
\toprule
 & \multicolumn{4}{c|}{\textbf{Random}} & \multicolumn{4}{c}{\textbf{Geographic}} \\
\textbf{Model} & FT & JT & TTT-MMR & TTT-MMR-Geo & FT & JT & TTT-MMR & TTT-MMR-Geo \\
\midrule\midrule
ConvNeXtV2A & 0.16 & 0.17 & 0.21 & \underline{0.22} & 0.19 & 0.19 & 0.28 & \underline{0.29} \\
Scale-MAE & 0.16 & 0.22 & \underline{0.26} & 0.25 & 0.10 & 0.10 & \underline{0.20} & \underline{0.20} \\
DINOv3 Web & \underline{0.38} & \underline{0.38} & \underline{0.38} & \underline{0.38} & 0.35 & 0.36 & \underline{0.39} & 0.38 \\
DINOv3 Sat & \underline{0.39} & 0.37 & 0.38 & 0.38 & \underline{0.41} & 0.35 & 0.37 & 0.37 \\
SatlasNet & 0.38 & 0.38 & \underline{0.40} & 0.39 & 0.37 & 0.34 & \underline{0.43} & 0.42 \\
MPMAE & \textbf{0.45} & \textbf{0.46} & \underline{\textbf{0.47}} & \underline{\textbf{0.47}} & 0.47 & 0.49 & \underline{0.50} & \underline{0.50} \\
TerraMind & \underline{\textbf{0.45}} & \underline{0.45} & \underline{0.45} & \underline{0.45} & \underline{0.49} & \underline{0.49} & \underline{0.49} & \underline{0.49} \\
Copernicus-FM & 0.44 & 0.44 & \underline{0.45} & \underline{0.45} & 0.45 & 0.45 & \underline{0.46} & \underline{0.46} \\
Galileo & \underline{0.43} & 0.42 & \underline{0.43} & \underline{0.43} & \underline{0.24} & 0.20 & 0.21 & 0.21 \\
ConvNeXtV2A-MM & 0.44 & 0.44 & 0.44 & \underline{0.45} & \underline{\textbf{0.51}} & \underline{\textbf{0.51}} & \underline{\textbf{0.51}} & \underline{\textbf{0.51}} \\
\bottomrule
\end{tabular}

}
\end{table}

\begin{table}
\centering
\caption{Biomass test R$^2$ by architecture, adaptation mode, and split (Seed 43)}
\label{tab:biomass_R2_seed43}
\resizebox{\linewidth}{!}{%
\begin{tabular}{l|cccc|cccc}
\toprule
 & \multicolumn{4}{c|}{\textbf{Random}} & \multicolumn{4}{c}{\textbf{Geographic}} \\
\textbf{Model} & FT & JT & TTT-MMR & TTT-MMR-Geo & FT & JT & TTT-MMR & TTT-MMR-Geo \\
\midrule\midrule
ConvNeXtV2A & 0.15 & 0.16 & 0.19 & \underline{0.20} & 0.15 & 0.12 & 0.23 & \underline{0.24} \\
Scale-MAE & 0.17 & 0.18 & \underline{0.19} & \underline{0.19} & 0.08 & 0.11 & \underline{0.14} & \underline{0.14} \\
DINOv3 Web & \underline{0.38} & 0.37 & \underline{0.38} & \underline{0.38} & \underline{0.36} & 0.32 & \underline{0.36} & 0.35 \\
DINOv3 Sat & \underline{0.40} & 0.39 & 0.39 & 0.39 & \underline{0.40} & 0.37 & 0.39 & 0.39 \\
SatlasNet & 0.38 & 0.39 & \underline{0.41} & \underline{0.41} & 0.34 & 0.37 & \underline{0.44} & 0.43 \\
MPMAE & \textbf{0.45} & \textbf{0.46} & \underline{\textbf{0.47}} & \textbf{0.46} & 0.46 & \underline{0.48} & \underline{0.48} & \underline{0.48} \\
TerraMind & 0.42 & 0.44 & \underline{0.45} & \underline{0.45} & 0.45 & \underline{\textbf{0.51}} & \underline{\textbf{0.51}} & \underline{\textbf{0.51}} \\
Copernicus-FM & \underline{0.44} & 0.43 & \underline{0.44} & \underline{0.44} & \underline{\textbf{0.49}} & 0.47 & 0.48 & 0.48 \\
Galileo & 0.42 & \underline{0.43} & \underline{0.43} & \underline{0.43} & 0.27 & 0.36 & \underline{0.37} & \underline{0.37} \\
ConvNeXtV2A-MM & \underline{0.44} & \underline{0.44} & \underline{0.44} & \underline{0.44} & \textbf{0.49} & 0.49 & \underline{0.50} & \underline{0.50} \\
\bottomrule
\end{tabular}

}
\end{table}

\begin{table}
\centering
\caption{Soil nitrogen test R$^2$ by architecture, adaptation mode, and split (Seed 41)}
\label{tab:soil_nitrogen_R2_seed41}
\resizebox{\linewidth}{!}{%
\begin{tabular}{l|cccc|cccc}
\toprule
 & \multicolumn{4}{c|}{\textbf{Random}} & \multicolumn{4}{c}{\textbf{Geographic}} \\
\textbf{Model} & FT & JT & TTT-MMR & TTT-MMR-Geo & FT & JT & TTT-MMR & TTT-MMR-Geo \\
\midrule\midrule
ConvNeXtV2A & 0.22 & 0.22 & 0.24 & \underline{0.26} & -0.07 & -0.07 & \underline{-0.01} & -0.02 \\
Scale-MAE & 0.16 & 0.18 & \underline{0.20} & \underline{0.20} & -0.49 & \underline{-0.42} & \underline{-0.42} & \underline{-0.42} \\
DINOv3 Web & 0.39 & 0.38 & 0.40 & \underline{0.41} & -0.17 & -0.23 & \underline{-0.16} & \underline{-0.16} \\
DINOv3 Sat & 0.37 & 0.37 & 0.38 & \underline{0.39} & \underline{0.01} & -0.03 & -0.02 & -0.02 \\
SatlasNet & 0.31 & 0.30 & 0.33 & \underline{0.35} & -0.10 & -0.09 & -0.10 & \underline{-0.04} \\
MPMAE & 0.42 & 0.44 & \underline{0.45} & \underline{0.45} & -0.08 & \underline{-0.02} & -0.04 & -0.03 \\
TerraMind & \underline{\textbf{0.50}} & \textbf{0.49} & \textbf{0.49} & \textbf{0.49} & \underline{-0.08} & -0.16 & -0.16 & -0.15 \\
Copernicus-FM & 0.44 & \underline{0.46} & \underline{0.46} & \underline{0.46} & \textbf{0.04} & \underline{\textbf{0.05}} & \underline{\textbf{0.05}} & \underline{\textbf{0.05}} \\
Galileo & \underline{0.44} & 0.43 & 0.43 & 0.42 & \underline{-0.05} & -0.43 & -0.26 & -0.23 \\
ConvNeXtV2A-MM & \underline{0.47} & \underline{0.47} & \underline{0.47} & \underline{0.47} & \underline{0.02} & -0.30 & -0.21 & -0.20 \\
\bottomrule
\end{tabular}

}
\end{table}

\begin{table}
\centering
\caption{Soil nitrogen test R$^2$ by architecture, adaptation mode, and split (Seed 42)}
\label{tab:soil_nitrogen_R2_seed42}
\resizebox{\linewidth}{!}{%
\begin{tabular}{l|cccc|cccc}
\toprule
 & \multicolumn{4}{c|}{\textbf{Random}} & \multicolumn{4}{c}{\textbf{Geographic}} \\
\textbf{Model} & FT & JT & TTT-MMR & TTT-MMR-Geo & FT & JT & TTT-MMR & TTT-MMR-Geo \\
\midrule\midrule
ConvNeXtV2A & 0.20 & 0.20 & 0.22 & \underline{0.23} & -0.12 & -0.12 & \underline{-0.07} & \underline{-0.07} \\
Scale-MAE & \underline{0.19} & 0.17 & \underline{0.19} & \underline{0.19} & \underline{-0.40} & -0.50 & -0.50 & -0.49 \\
DINOv3 Web & 0.36 & 0.38 & \underline{0.39} & \underline{0.39} & -0.05 & -0.04 & -0.03 & \underline{-0.02} \\
DINOv3 Sat & 0.38 & 0.38 & \underline{0.39} & \underline{0.39} & 0.00 & \textbf{0.01} & \textbf{0.01} & \underline{0.02} \\
SatlasNet & 0.31 & 0.30 & 0.36 & \underline{0.37} & -0.12 & -0.23 & -0.06 & \underline{\textbf{0.05}} \\
MPMAE & \underline{0.45} & 0.41 & 0.43 & 0.43 & -0.08 & -0.03 & -0.02 & \underline{-0.01} \\
TerraMind & \underline{\textbf{0.51}} & \underline{\textbf{0.51}} & \underline{\textbf{0.51}} & \underline{\textbf{0.51}} & \underline{-0.11} & -0.12 & -0.12 & -0.12 \\
Copernicus-FM & 0.46 & 0.48 & 0.48 & \underline{0.49} & \underline{\textbf{0.08}} & -0.05 & -0.04 & -0.02 \\
Galileo & 0.43 & 0.44 & 0.45 & \underline{0.46} & \underline{0.01} & -0.07 & -0.06 & -0.06 \\
ConvNeXtV2A-MM & \underline{0.48} & \underline{0.48} & \underline{0.48} & \underline{0.48} & \underline{0.03} & -0.09 & -0.08 & -0.07 \\
\bottomrule
\end{tabular}

}
\end{table}

\begin{table}
\centering
\caption{Soil nitrogen test R$^2$ by architecture, adaptation mode, and split (Seed 43)}
\label{tab:soil_nitrogen_R2_seed43}
\resizebox{\linewidth}{!}{%
\begin{tabular}{l|cccc|cccc}
\toprule
 & \multicolumn{4}{c|}{\textbf{Random}} & \multicolumn{4}{c}{\textbf{Geographic}} \\
\textbf{Model} & FT & JT & TTT-MMR & TTT-MMR-Geo & FT & JT & TTT-MMR & TTT-MMR-Geo \\
\midrule\midrule
ConvNeXtV2A & 0.22 & 0.22 & 0.24 & \underline{0.27} & -0.24 & -0.19 & \underline{-0.08} & -0.09 \\
Scale-MAE & \underline{0.20} & 0.15 & 0.17 & 0.18 & \underline{-0.28} & -0.31 & -0.30 & -0.29 \\
DINOv3 Web & 0.38 & 0.39 & \underline{0.40} & \underline{0.40} & -0.12 & -0.09 & \underline{-0.07} & \underline{-0.07} \\
DINOv3 Sat & 0.38 & 0.39 & 0.41 & \underline{0.42} & \underline{\textbf{-0.00}} & -0.03 & -0.02 & -0.02 \\
SatlasNet & 0.32 & 0.31 & 0.32 & \underline{0.35} & -0.18 & -0.08 & -0.07 & \underline{-0.01} \\
MPMAE & 0.38 & 0.40 & \underline{0.42} & \underline{0.42} & \underline{-0.03} & -0.05 & -0.05 & -0.05 \\
TerraMind & \underline{\textbf{0.50}} & \textbf{0.49} & \underline{\textbf{0.50}} & \underline{\textbf{0.50}} & -0.09 & -0.08 & \underline{-0.07} & \underline{-0.07} \\
Copernicus-FM & \underline{0.48} & 0.46 & 0.46 & 0.46 & -0.13 & \underline{\textbf{0.04}} & \underline{\textbf{0.04}} & \underline{\textbf{0.04}} \\
Galileo & 0.43 & \underline{0.45} & \underline{0.45} & \underline{0.45} & -0.07 & -0.07 & -0.06 & \underline{-0.05} \\
ConvNeXtV2A-MM & 0.47 & 0.47 & 0.47 & \underline{0.48} & \textbf{0.00} & 0.02 & 0.03 & \underline{\textbf{0.04}} \\
\bottomrule
\end{tabular}

}
\end{table}

\begin{table}
\centering
\caption{Soil organic carbon test R$^2$ by architecture, adaptation mode, and split (Seed 41)}
\label{tab:soil_organic_carbon_R2_seed41}
\resizebox{\linewidth}{!}{%
\begin{tabular}{l|cccc|cccc}
\toprule
 & \multicolumn{4}{c|}{\textbf{Random}} & \multicolumn{4}{c}{\textbf{Geographic}} \\
\textbf{Model} & FT & JT & TTT-MMR & TTT-MMR-Geo & FT & JT & TTT-MMR & TTT-MMR-Geo \\
\midrule\midrule
ConvNeXtV2A & 0.21 & 0.22 & 0.24 & \underline{0.25} & -0.72 & -0.62 & \underline{-0.45} & -0.50 \\
Scale-MAE & \underline{0.13} & 0.09 & 0.11 & 0.11 & \underline{-2.21} & -3.20 & -3.07 & -3.04 \\
DINOv3 Web & 0.11 & 0.12 & \underline{0.15} & \underline{0.15} & \underline{-1.92} & -2.07 & -1.95 & \underline{-1.92} \\
DINOv3 Sat & 0.05 & 0.10 & 0.15 & \underline{0.16} & -2.22 & -2.07 & -1.61 & \underline{-1.52} \\
SatlasNet & 0.30 & \underline{0.33} & 0.31 & \underline{0.33} & \underline{-0.03} & -0.08 & -0.25 & -0.17 \\
MPMAE & \underline{0.40} & 0.38 & 0.39 & \underline{0.40} & \underline{0.01} & 0.00 & \underline{0.01} & \underline{0.01} \\
TerraMind & \underline{0.33} & 0.32 & \underline{0.33} & \underline{0.33} & -0.20 & -0.11 & -0.09 & \underline{-0.08} \\
Copernicus-FM & \underline{0.45} & \underline{0.45} & \underline{0.45} & \underline{0.45} & \underline{\textbf{0.09}} & \textbf{0.03} & \textbf{0.03} & \textbf{0.03} \\
Galileo & \underline{\textbf{0.49}} & 0.44 & 0.44 & 0.43 & \underline{-1.46} & -3.37 & -2.96 & -2.91 \\
ConvNeXtV2A-MM & 0.43 & \underline{\textbf{0.49}} & \textbf{0.48} & \textbf{0.48} & -1.80 & \underline{-0.36} & -0.47 & -0.50 \\
\bottomrule
\end{tabular}

}
\end{table}

\begin{table}
\centering
\caption{Soil organic carbon test R$^2$ by architecture, adaptation mode, and split (Seed 42)}
\label{tab:soil_organic_carbon_R2_seed42}
\resizebox{\linewidth}{!}{%
\begin{tabular}{l|cccc|cccc}
\toprule
 & \multicolumn{4}{c|}{\textbf{Random}} & \multicolumn{4}{c}{\textbf{Geographic}} \\
\textbf{Model} & FT & JT & TTT-MMR & TTT-MMR-Geo & FT & JT & TTT-MMR & TTT-MMR-Geo \\
\midrule\midrule
ConvNeXtV2A & 0.18 & 0.18 & \underline{0.25} & \underline{0.25} & -0.31 & -0.42 & \underline{-0.18} & -0.22 \\
Scale-MAE & 0.10 & 0.10 & \underline{0.11} & \underline{0.11} & \underline{-1.90} & -2.13 & -2.02 & -2.00 \\
DINOv3 Web & 0.12 & 0.12 & \underline{0.17} & \underline{0.17} & \underline{-2.72} & -3.22 & -2.80 & -2.81 \\
DINOv3 Sat & 0.08 & 0.11 & \underline{0.14} & \underline{0.14} & -2.08 & -1.88 & \underline{-1.55} & -1.56 \\
SatlasNet & 0.31 & 0.32 & \underline{0.34} & \underline{0.34} & \underline{-0.09} & -0.40 & -0.21 & -0.25 \\
MPMAE & 0.40 & 0.40 & \underline{0.41} & \underline{0.41} & -0.12 & -0.00 & 0.04 & \underline{0.05} \\
TerraMind & 0.34 & 0.34 & \underline{0.36} & \underline{0.36} & \underline{-0.25} & -0.46 & -0.41 & -0.40 \\
Copernicus-FM & \underline{\textbf{0.50}} & 0.46 & 0.46 & 0.46 & \underline{\textbf{0.08}} & \textbf{0.06} & \textbf{0.06} & \textbf{0.06} \\
Galileo & 0.45 & \underline{\textbf{0.48}} & \underline{\textbf{0.48}} & 0.47 & \underline{-1.22} & -1.92 & -1.65 & -1.64 \\
ConvNeXtV2A-MM & 0.47 & \underline{\textbf{0.48}} & 0.47 & \underline{\textbf{0.48}} & -1.11 & -0.62 & \underline{-0.52} & -0.53 \\
\bottomrule
\end{tabular}

}
\end{table}

\begin{table}
\centering
\caption{Soil organic carbon test R$^2$ by architecture, adaptation mode, and split (Seed 43)}
\label{tab:soil_organic_carbon_R2_seed43}
\resizebox{\linewidth}{!}{%
\begin{tabular}{l|cccc|cccc}
\toprule
 & \multicolumn{4}{c|}{\textbf{Random}} & \multicolumn{4}{c}{\textbf{Geographic}} \\
\textbf{Model} & FT & JT & TTT-MMR & TTT-MMR-Geo & FT & JT & TTT-MMR & TTT-MMR-Geo \\
\midrule\midrule
ConvNeXtV2A & 0.19 & 0.20 & 0.22 & \underline{0.24} & -0.48 & -0.75 & -0.55 & \underline{-0.38} \\
Scale-MAE & \underline{0.10} & 0.08 & 0.09 & \underline{0.10} & -2.52 & -1.60 & -1.56 & \underline{-1.53} \\
DINOv3 Web & 0.10 & 0.13 & \underline{0.16} & \underline{0.16} & -1.97 & -2.02 & -1.85 & \underline{-1.82} \\
DINOv3 Sat & 0.12 & 0.09 & \underline{0.14} & 0.13 & -2.26 & -2.19 & \underline{-2.05} & -2.06 \\
SatlasNet & 0.31 & 0.31 & 0.32 & \underline{0.34} & -0.32 & -0.25 & \underline{-0.19} & -0.22 \\
MPMAE & 0.40 & 0.40 & \underline{0.41} & \underline{0.41} & -0.13 & \textbf{0.08} & \textbf{0.15} & \underline{\textbf{0.17}} \\
TerraMind & 0.33 & 0.34 & \underline{0.35} & \underline{0.35} & \underline{-0.32} & -1.06 & -0.99 & -0.98 \\
Copernicus-FM & \underline{0.47} & \underline{0.47} & 0.46 & 0.46 & \textbf{0.05} & \underline{\textbf{0.08}} & \underline{0.08} & \underline{0.08} \\
Galileo & \underline{\textbf{0.51}} & \textbf{0.48} & 0.47 & 0.47 & -1.30 & -1.39 & -0.92 & \underline{-0.90} \\
ConvNeXtV2A-MM & 0.43 & 0.47 & \underline{\textbf{0.48}} & \underline{\textbf{0.48}} & \underline{-0.51} & -0.84 & -0.62 & -0.59 \\
\bottomrule
\end{tabular}

}
\end{table}

\begin{table}
\centering
\caption{Soil pH test R$^2$ by architecture, adaptation mode, and split (Seed 41)}
\label{tab:soil_pH_R2_seed41}
\resizebox{\linewidth}{!}{%
\begin{tabular}{l|cccc|cccc}
\toprule
 & \multicolumn{4}{c|}{\textbf{Random}} & \multicolumn{4}{c}{\textbf{Geographic}} \\
\textbf{Model} & FT & JT & TTT-MMR & TTT-MMR-Geo & FT & JT & TTT-MMR & TTT-MMR-Geo \\
\midrule\midrule
ConvNeXtV2A & 0.39 & 0.41 & 0.42 & \underline{0.46} & -0.03 & 0.04 & \underline{0.14} & 0.13 \\
Scale-MAE & 0.40 & 0.43 & 0.45 & \underline{0.46} & -0.07 & -0.01 & \underline{0.01} & \underline{0.01} \\
DINOv3 Web & 0.56 & 0.58 & \underline{0.59} & \underline{0.59} & 0.18 & 0.23 & \underline{0.25} & \underline{0.25} \\
DINOv3 Sat & 0.54 & 0.56 & \underline{0.58} & \underline{0.58} & 0.16 & 0.20 & \underline{0.21} & \underline{0.21} \\
SatlasNet & 0.52 & 0.54 & 0.54 & \underline{0.56} & \underline{0.29} & 0.20 & 0.21 & 0.19 \\
MPMAE & 0.57 & 0.59 & 0.59 & \underline{0.60} & 0.21 & 0.26 & 0.27 & \underline{0.29} \\
TerraMind & \textbf{0.67} & \textbf{0.67} & \underline{\textbf{0.68}} & \underline{\textbf{0.68}} & \underline{\textbf{0.36}} & \textbf{0.35} & \textbf{0.35} & \textbf{0.35} \\
Copernicus-FM & 0.63 & \underline{0.65} & \underline{0.65} & \underline{0.65} & 0.17 & 0.19 & 0.19 & \underline{0.20} \\
Galileo & \underline{0.64} & \underline{0.64} & \underline{0.64} & \underline{0.64} & 0.17 & 0.18 & \underline{0.19} & 0.18 \\
ConvNeXtV2A-MM & \underline{0.64} & 0.63 & 0.63 & 0.63 & \underline{-0.15} & -0.42 & -0.52 & -0.59 \\
\bottomrule
\end{tabular}

}
\end{table}

\begin{table}
\centering
\caption{Soil pH test R$^2$ by architecture, adaptation mode, and split (Seed 42)}
\label{tab:soil_pH_R2_seed42}
\resizebox{\linewidth}{!}{%
\begin{tabular}{l|cccc|cccc}
\toprule
 & \multicolumn{4}{c|}{\textbf{Random}} & \multicolumn{4}{c}{\textbf{Geographic}} \\
\textbf{Model} & FT & JT & TTT-MMR & TTT-MMR-Geo & FT & JT & TTT-MMR & TTT-MMR-Geo \\
\midrule\midrule
ConvNeXtV2A & 0.42 & 0.42 & 0.41 & \underline{0.46} & -0.01 & 0.05 & \underline{0.16} & 0.12 \\
Scale-MAE & 0.39 & 0.42 & 0.44 & \underline{0.45} & -0.04 & -0.00 & 0.02 & \underline{0.03} \\
DINOv3 Web & 0.55 & 0.58 & \underline{0.59} & \underline{0.59} & 0.18 & 0.26 & \underline{0.27} & \underline{0.27} \\
DINOv3 Sat & 0.55 & 0.57 & \underline{0.58} & \underline{0.58} & 0.17 & 0.22 & \underline{0.23} & \underline{0.23} \\
SatlasNet & 0.50 & 0.55 & 0.55 & \underline{0.56} & 0.20 & 0.20 & \underline{0.22} & \underline{0.22} \\
MPMAE & 0.56 & 0.59 & 0.60 & \underline{0.61} & 0.27 & 0.26 & \underline{0.28} & 0.27 \\
TerraMind & \textbf{0.67} & \textbf{0.67} & \underline{\textbf{0.68}} & \underline{\textbf{0.68}} & \underline{\textbf{0.36}} & \textbf{0.35} & \textbf{0.35} & \textbf{0.35} \\
Copernicus-FM & 0.62 & \underline{0.65} & \underline{0.65} & \underline{0.65} & 0.15 & 0.19 & \underline{0.20} & \underline{0.20} \\
Galileo & 0.63 & 0.63 & \underline{0.64} & \underline{0.64} & \underline{0.24} & 0.16 & 0.17 & 0.17 \\
ConvNeXtV2A-MM & \underline{0.65} & 0.63 & 0.64 & 0.64 & \underline{-0.12} & -0.18 & -0.23 & -0.24 \\
\bottomrule
\end{tabular}

}
\end{table}

\begin{table}
\centering
\caption{Soil pH test R$^2$ by architecture, adaptation mode, and split (Seed 43)}
\label{tab:soil_pH_R2_seed43}
\resizebox{\linewidth}{!}{%
\begin{tabular}{l|cccc|cccc}
\toprule
 & \multicolumn{4}{c|}{\textbf{Random}} & \multicolumn{4}{c}{\textbf{Geographic}} \\
\textbf{Model} & FT & JT & TTT-MMR & TTT-MMR-Geo & FT & JT & TTT-MMR & TTT-MMR-Geo \\
\midrule\midrule
ConvNeXtV2A & 0.37 & 0.40 & 0.40 & \underline{0.44} & -0.04 & 0.03 & \underline{0.10} & 0.08 \\
Scale-MAE & 0.39 & 0.43 & 0.45 & \underline{0.46} & -0.10 & -0.03 & \underline{0.00} & \underline{0.00} \\
DINOv3 Web & 0.55 & 0.58 & \underline{0.59} & \underline{0.59} & 0.18 & 0.24 & \underline{0.25} & \underline{0.25} \\
DINOv3 Sat & 0.54 & 0.56 & \underline{0.58} & \underline{0.58} & 0.16 & 0.21 & \underline{0.22} & \underline{0.22} \\
SatlasNet & 0.50 & 0.54 & 0.55 & \underline{0.56} & \underline{0.23} & 0.22 & \underline{0.23} & 0.20 \\
MPMAE & 0.58 & 0.59 & 0.60 & \underline{0.61} & 0.22 & 0.21 & 0.23 & \underline{0.24} \\
TerraMind & \textbf{0.67} & \textbf{0.67} & \underline{\textbf{0.68}} & \underline{\textbf{0.68}} & \underline{\textbf{0.34}} & \textbf{0.33} & \textbf{0.33} & \textbf{0.33} \\
Copernicus-FM & 0.62 & \underline{0.65} & \underline{0.65} & \underline{0.65} & 0.15 & \underline{0.20} & \underline{0.20} & \underline{0.20} \\
Galileo & 0.63 & 0.63 & \underline{0.64} & \underline{0.64} & 0.18 & 0.22 & \underline{0.23} & 0.22 \\
ConvNeXtV2A-MM & \underline{0.65} & 0.62 & 0.62 & 0.63 & -0.27 & \underline{-0.24} & -0.28 & -0.31 \\
\bottomrule
\end{tabular}

}
\end{table}

\begin{table}
\centering
\caption{Species test mAP by architecture, adaptation mode, and split (Seed 41)}
\label{tab:species_mAP_seed41}
\resizebox{\linewidth}{!}{%
\begin{tabular}{l|cccc|cccc}
\toprule
 & \multicolumn{4}{c|}{\textbf{Random}} & \multicolumn{4}{c}{\textbf{Geographic}} \\
\textbf{Model} & FT & JT & TTT-MMR & TTT-MMR-Geo & FT & JT & TTT-MMR & TTT-MMR-Geo \\
\midrule\midrule
ConvNeXtV2A & 0.47 & 0.53 & 0.55 & \underline{0.60} & 0.32 & 0.32 & 0.33 & \underline{0.34} \\
Scale-MAE & 0.67 & 0.72 & 0.76 & \underline{0.77} & 0.33 & 0.33 & \underline{0.34} & \underline{0.34} \\
DINOv3 Web & 0.82 & 0.84 & \underline{0.87} & \underline{0.87} & 0.36 & 0.36 & \underline{0.37} & \underline{0.37} \\
DINOv3 Sat & 0.82 & 0.84 & \underline{0.86} & \underline{0.86} & 0.36 & 0.36 & \underline{0.37} & \underline{0.37} \\
SatlasNet & 0.84 & 0.86 & 0.89 & \underline{0.90} & 0.36 & 0.38 & 0.38 & \underline{0.39} \\
MPMAE & 0.90 & 0.90 & \underline{0.92} & 0.91 & 0.39 & 0.40 & \underline{0.41} & \underline{0.41} \\
TerraMind & \underline{0.95} & \underline{0.95} & \underline{0.95} & \underline{0.95} & \underline{0.41} & 0.40 & \underline{0.41} & \underline{0.41} \\
Copernicus-FM & \underline{\textbf{0.99}} & \underline{\textbf{0.99}} & \underline{\textbf{0.99}} & \underline{\textbf{0.99}} & \textbf{0.45} & \textbf{0.47} & \textbf{0.47} & \underline{\textbf{0.48}} \\
Galileo & \underline{\textbf{0.99}} & \underline{\textbf{0.99}} & \underline{\textbf{0.99}} & \underline{\textbf{0.99}} & 0.43 & \underline{0.44} & \underline{0.44} & \underline{0.44} \\
ConvNeXtV2A-MM & \underline{\textbf{0.99}} & \underline{\textbf{0.99}} & \underline{\textbf{0.99}} & \underline{\textbf{0.99}} & 0.41 & \underline{0.45} & \underline{0.45} & \underline{0.45} \\
\bottomrule
\end{tabular}

}
\end{table}

\begin{table}
\centering
\caption{Species test mAP by architecture, adaptation mode, and split (Seed 42)}
\label{tab:species_mAP_seed42}
\resizebox{\linewidth}{!}{%
\begin{tabular}{l|cccc|cccc}
\toprule
 & \multicolumn{4}{c|}{\textbf{Random}} & \multicolumn{4}{c}{\textbf{Geographic}} \\
\textbf{Model} & FT & JT & TTT-MMR & TTT-MMR-Geo & FT & JT & TTT-MMR & TTT-MMR-Geo \\
\midrule\midrule
ConvNeXtV2A & 0.48 & 0.53 & 0.55 & \underline{0.59} & 0.32 & 0.33 & 0.34 & \underline{0.35} \\
Scale-MAE & 0.67 & 0.72 & 0.76 & \underline{0.77} & 0.33 & 0.34 & \underline{0.35} & \underline{0.35} \\
DINOv3 Web & 0.82 & 0.84 & \underline{0.87} & \underline{0.87} & 0.36 & 0.36 & \underline{0.37} & \underline{0.37} \\
DINOv3 Sat & 0.83 & 0.83 & \underline{0.86} & \underline{0.86} & \underline{0.37} & 0.36 & \underline{0.37} & \underline{0.37} \\
SatlasNet & 0.84 & 0.87 & 0.89 & \underline{0.90} & 0.36 & 0.37 & 0.38 & \underline{0.39} \\
MPMAE & 0.90 & 0.90 & \underline{0.91} & \underline{0.91} & 0.39 & 0.40 & 0.41 & \underline{0.42} \\
TerraMind & \underline{0.95} & \underline{0.95} & \underline{0.95} & \underline{0.95} & \underline{0.41} & \underline{0.41} & \underline{0.41} & \underline{0.41} \\
Copernicus-FM & \underline{\textbf{0.99}} & \underline{\textbf{0.99}} & \underline{\textbf{0.99}} & \underline{\textbf{0.99}} & 0.43 & \textbf{0.46} & \underline{\textbf{0.47}} & \underline{\textbf{0.47}} \\
Galileo & \underline{\textbf{0.99}} & \underline{\textbf{0.99}} & \underline{\textbf{0.99}} & \underline{\textbf{0.99}} & \underline{\textbf{0.44}} & \underline{0.44} & \underline{0.44} & \underline{0.44} \\
ConvNeXtV2A-MM & \underline{\textbf{0.99}} & \underline{\textbf{0.99}} & \underline{\textbf{0.99}} & \underline{\textbf{0.99}} & 0.42 & \underline{0.45} & \underline{0.45} & \underline{0.45} \\
\bottomrule
\end{tabular}

}
\end{table}

\begin{table}
\centering
\caption{Species test mAP by architecture, adaptation mode, and split (Seed 43)}
\label{tab:species_mAP_seed43}
\resizebox{\linewidth}{!}{%
\begin{tabular}{l|cccc|cccc}
\toprule
 & \multicolumn{4}{c|}{\textbf{Random}} & \multicolumn{4}{c}{\textbf{Geographic}} \\
\textbf{Model} & FT & JT & TTT-MMR & TTT-MMR-Geo & FT & JT & TTT-MMR & TTT-MMR-Geo \\
\midrule\midrule
ConvNeXtV2A & 0.48 & 0.53 & 0.56 & \underline{0.61} & 0.32 & 0.32 & 0.34 & \underline{0.35} \\
Scale-MAE & 0.67 & 0.72 & \underline{0.77} & \underline{0.77} & 0.34 & 0.34 & \underline{0.35} & \underline{0.35} \\
DINOv3 Web & 0.82 & 0.84 & \underline{0.87} & \underline{0.87} & 0.36 & 0.36 & \underline{0.37} & \underline{0.37} \\
DINOv3 Sat & 0.83 & 0.83 & \underline{0.86} & \underline{0.86} & 0.36 & \underline{0.37} & \underline{0.37} & \underline{0.37} \\
SatlasNet & 0.84 & 0.86 & 0.89 & \underline{0.90} & 0.36 & 0.37 & 0.38 & \underline{0.39} \\
MPMAE & 0.90 & 0.90 & 0.91 & \underline{0.92} & 0.38 & 0.40 & 0.41 & \underline{0.42} \\
TerraMind & \underline{0.95} & \underline{0.95} & \underline{0.95} & \underline{0.95} & 0.40 & \underline{0.41} & \underline{0.41} & \underline{0.41} \\
Copernicus-FM & \underline{\textbf{0.99}} & \underline{\textbf{0.99}} & \underline{\textbf{0.99}} & \underline{\textbf{0.99}} & \textbf{0.44} & \underline{\textbf{0.47}} & \underline{\textbf{0.47}} & \underline{\textbf{0.47}} \\
Galileo & \underline{\textbf{0.99}} & \underline{\textbf{0.99}} & \underline{\textbf{0.99}} & \underline{\textbf{0.99}} & 0.42 & 0.44 & 0.44 & \underline{0.45} \\
ConvNeXtV2A-MM & \underline{\textbf{0.99}} & \underline{\textbf{0.99}} & \underline{\textbf{0.99}} & \underline{\textbf{0.99}} & 0.42 & \underline{0.45} & \underline{0.45} & \underline{0.45} \\
\bottomrule
\end{tabular}

}
\end{table}

\FloatBarrier
\subsection{Linear probing}
The ``Random'' section of \crefrange{tab:LP-biomass}{tab:LP-species} shows the numbers used to create \cref{fig:rq1_LP_random}. The ``Geographic'' section of \crefrange{tab:LP-biomass}{tab:LP-species} shows the numbers used to create \cref{fig:rq1_LP_geographic}. The ``100\%'' columns in \crefrange{tab:LP-biomass}{tab:LP-species} were used to generate \cref{fig:rq2_LP}. We made \cref{fig:rq3_LP} based off the ``TerraMind S2'', ``TerraMind'', ``Copernicus-FM S2'', ``Copernicus-FM'', ``Galileo S2'', and ``Galileo'' rows in \crefrange{tab:LP-biomass}{tab:LP-species}.

\begin{table*}
\centering
\caption{Biomass linear probing test R$^2$}
\resizebox{\linewidth}{!}{%
\begin{tabular}{l|rrr|rrr|rrr|rrr|rrr|rrr}
\toprule
 & \multicolumn{9}{c|}{\textbf{Random}} & \multicolumn{9}{c}{\textbf{Geographic}} \\
 & \multicolumn{3}{c|}{\textbf{Seed 41}} & \multicolumn{3}{c|}{\textbf{Seed 42}} & \multicolumn{3}{c|}{\textbf{Seed 43}} & \multicolumn{3}{c|}{\textbf{Seed 41}} & \multicolumn{3}{c|}{\textbf{Seed 42}} & \multicolumn{3}{c}{\textbf{Seed 43}} \\
 & \textbf{5\%} & \textbf{50\%} & \textbf{100\%} & \textbf{5\%} & \textbf{50\%} & \textbf{100\%} & \textbf{5\%} & \textbf{50\%} & \textbf{100\%} & \textbf{5\%} & \textbf{50\%} & \textbf{100\%} & \textbf{5\%} & \textbf{50\%} & \textbf{100\%} & \textbf{5\%} & \textbf{50\%} & \textbf{100\%} \\
\midrule
\midrule
ConvNeXtV2A & -0.34 & -0.03 & 0.02 & -0.34 & -0.03 & 0.02 & -0.34 & -0.03 & 0.02 & -0.09 & -0.15 & -0.10 & -0.09 & -0.15 & -0.10 & -0.09 & -0.15 & -0.10 \\
Scale-MAE & -0.33 & 0.04 & 0.09 & -0.33 & 0.04 & 0.09 & -0.33 & 0.04 & 0.09 & -0.07 & -0.05 & -0.06 & -0.07 & -0.05 & -0.06 & -0.07 & -0.05 & -0.06 \\
DINOv3 Web & -0.35 & 0.01 & 0.11 & -0.35 & 0.01 & 0.11 & -0.35 & 0.01 & 0.11 & -0.10 & -0.02 & -0.06 & -0.10 & -0.02 & -0.06 & -0.10 & -0.02 & -0.06 \\
DINOv3 Sat & -0.40 & -0.19 & -0.03 & -0.40 & -0.19 & -0.03 & -0.40 & -0.19 & -0.03 & -0.13 & -0.00 & 0.04 & -0.13 & -0.01 & 0.04 & -0.13 & -0.00 & 0.04 \\
SatlasNet & -0.34 & 0.05 & 0.14 & -0.34 & 0.05 & 0.14 & -0.34 & 0.05 & 0.14 & -0.09 & 0.02 & 0.02 & -0.09 & 0.02 & 0.02 & -0.09 & 0.02 & 0.02 \\
MPMAE & -0.29 & 0.10 & 0.19 & -0.29 & 0.10 & 0.19 & -0.29 & 0.10 & 0.19 & -0.06 & \textbf{0.12} & \textbf{0.21} & -0.06 & \textbf{0.12} & \textbf{0.21} & -0.06 & \textbf{0.12} & \textbf{0.21} \\
TerraMind S2 & -0.38 & -0.06 & 0.13 & -0.38 & -0.06 & 0.13 & -0.38 & -0.06 & 0.13 & -0.12 & 0.07 & 0.13 & -0.12 & 0.07 & 0.13 & -0.12 & 0.07 & 0.13 \\
TerraMind & -0.36 & -0.00 & 0.13 & -0.36 & -0.00 & 0.13 & -0.36 & -0.00 & 0.13 & -0.10 & 0.08 & 0.05 & -0.10 & 0.08 & 0.05 & -0.10 & 0.08 & 0.05 \\
Copernicus-FM S2 & \textbf{-0.03} & 0.12 & 0.13 & \textbf{-0.03} & 0.12 & 0.13 & \textbf{-0.03} & 0.12 & 0.13 & -0.08 & 0.01 & -0.02 & -0.08 & 0.01 & -0.02 & -0.08 & 0.01 & -0.02 \\
Copernicus-FM & -0.09 & \textbf{0.16} & \textbf{0.24} & -0.08 & \textbf{0.16} & \textbf{0.24} & -0.08 & \textbf{0.16} & \textbf{0.24} & \textbf{0.02} & 0.08 & 0.19 & \textbf{0.02} & 0.08 & 0.19 & \textbf{0.02} & 0.08 & 0.19 \\
Galileo S2 & -0.34 & -0.01 & 0.06 & -0.34 & -0.01 & 0.06 & -0.34 & -0.01 & 0.06 & -0.09 & -0.00 & -0.04 & -0.09 & -0.00 & -0.04 & -0.09 & -0.00 & -0.04 \\
Galileo & -0.37 & -0.07 & 0.04 & -0.37 & -0.07 & 0.04 & -0.37 & -0.07 & 0.04 & -0.11 & 0.02 & -0.06 & -0.11 & 0.02 & -0.06 & -0.11 & 0.02 & -0.06 \\
ConvNeXtV2A-MM & -0.36 & 0.01 & 0.19 & -0.36 & 0.01 & 0.19 & -0.36 & 0.01 & 0.19 & -0.12 & -0.02 & 0.00 & -0.12 & -0.02 & 0.00 & -0.12 & -0.02 & 0.00 \\
\bottomrule
\end{tabular}

}
\label{tab:LP-biomass}
\end{table*}

\begin{table*}
\centering
\caption{Soil nitrogen linear probing test R$^2$}
\resizebox{\linewidth}{!}{%
\begin{tabular}{l|rrr|rrr|rrr|rrr|rrr|rrr}
\toprule
 & \multicolumn{9}{c|}{\textbf{Random}} & \multicolumn{9}{c}{\textbf{Geographic}} \\
 & \multicolumn{3}{c|}{\textbf{Seed 41}} & \multicolumn{3}{c|}{\textbf{Seed 42}} & \multicolumn{3}{c|}{\textbf{Seed 43}} & \multicolumn{3}{c|}{\textbf{Seed 41}} & \multicolumn{3}{c|}{\textbf{Seed 42}} & \multicolumn{3}{c}{\textbf{Seed 43}} \\
 & \textbf{5\%} & \textbf{50\%} & \textbf{100\%} & \textbf{5\%} & \textbf{50\%} & \textbf{100\%} & \textbf{5\%} & \textbf{50\%} & \textbf{100\%} & \textbf{5\%} & \textbf{50\%} & \textbf{100\%} & \textbf{5\%} & \textbf{50\%} & \textbf{100\%} & \textbf{5\%} & \textbf{50\%} & \textbf{100\%} \\
\midrule
\midrule
ConvNeXtV2A & -0.20 & 0.13 & 0.15 & -0.20 & 0.14 & 0.15 & -0.20 & 0.14 & 0.15 & -0.20 & -0.16 & -0.16 & -0.20 & -0.16 & -0.17 & -0.20 & -0.16 & -0.16 \\
Scale-MAE & -0.05 & 0.14 & 0.16 & -0.05 & 0.14 & 0.16 & -0.05 & 0.14 & 0.16 & \textbf{0.02} & -0.26 & -0.25 & \textbf{0.02} & -0.26 & -0.25 & \textbf{0.02} & -0.26 & -0.25 \\
DINOv3 Web & 0.05 & 0.24 & 0.27 & 0.05 & 0.24 & 0.27 & 0.05 & 0.24 & 0.27 & -0.00 & -0.33 & -0.33 & -0.00 & -0.33 & -0.33 & -0.00 & -0.33 & -0.32 \\
DINOv3 Sat & 0.05 & 0.30 & 0.34 & 0.05 & 0.30 & 0.34 & 0.05 & 0.30 & 0.34 & -0.01 & -0.10 & -0.07 & -0.01 & -0.10 & -0.07 & -0.01 & -0.10 & -0.07 \\
SatlasNet & \textbf{0.10} & 0.22 & 0.23 & \textbf{0.10} & 0.22 & 0.23 & \textbf{0.10} & 0.22 & 0.23 & -0.04 & -0.19 & -0.17 & -0.04 & -0.19 & -0.17 & -0.04 & -0.19 & -0.17 \\
MPMAE & -0.10 & 0.27 & 0.31 & -0.10 & 0.27 & 0.31 & -0.10 & 0.28 & 0.31 & -0.05 & -0.16 & -0.18 & -0.05 & -0.16 & -0.18 & -0.05 & -0.16 & -0.18 \\
TerraMind S2 & -0.24 & 0.33 & \textbf{0.41} & -0.24 & 0.34 & \textbf{0.41} & -0.24 & 0.34 & \textbf{0.41} & -0.12 & -0.01 & 0.03 & -0.12 & -0.01 & 0.03 & -0.12 & -0.01 & 0.03 \\
TerraMind & -0.16 & 0.28 & 0.36 & -0.16 & 0.28 & 0.36 & -0.16 & 0.28 & 0.36 & -0.06 & -0.13 & -0.04 & -0.06 & -0.13 & -0.04 & -0.06 & -0.13 & -0.04 \\
Copernicus-FM S2 & -0.05 & 0.19 & 0.23 & -0.05 & 0.19 & 0.23 & -0.05 & 0.19 & 0.23 & -0.00 & -0.22 & -0.26 & -0.00 & -0.22 & -0.26 & -0.00 & -0.22 & -0.26 \\
Copernicus-FM & -0.10 & 0.16 & 0.24 & -0.10 & 0.16 & 0.24 & -0.10 & 0.16 & 0.24 & -0.00 & -0.15 & -0.16 & -0.00 & -0.15 & -0.16 & -0.00 & -0.15 & -0.16 \\
Galileo S2 & 0.07 & 0.22 & 0.25 & 0.07 & 0.22 & 0.25 & 0.07 & 0.22 & 0.25 & -0.02 & -0.29 & -0.25 & -0.02 & -0.29 & -0.25 & -0.02 & -0.29 & -0.25 \\
Galileo & 0.05 & 0.31 & 0.36 & 0.05 & 0.31 & 0.36 & 0.05 & 0.31 & 0.36 & \textbf{0.02} & \textbf{0.07} & \textbf{0.08} & \textbf{0.02} & \textbf{0.07} & \textbf{0.09} & \textbf{0.02} & \textbf{0.07} & \textbf{0.09} \\
ConvNeXtV2A-MM & -0.21 & \textbf{0.35} & 0.39 & -0.21 & \textbf{0.35} & 0.39 & -0.21 & \textbf{0.35} & 0.39 & -0.45 & -0.00 & 0.06 & -0.45 & 0.00 & 0.06 & -0.45 & -0.00 & 0.06 \\
\bottomrule
\end{tabular}

}
\label{tab:LP-soil_nitrogen}
\end{table*}

\begin{table*}
\centering
\caption{Soil organic carbon linear probing test R$^2$}
\resizebox{\linewidth}{!}{%
\begin{tabular}{l|rrr|rrr|rrr|rrr|rrr|rrr}
\toprule
 & \multicolumn{9}{c|}{\textbf{Random}} & \multicolumn{9}{c}{\textbf{Geographic}} \\
 & \multicolumn{3}{c|}{\textbf{Seed 41}} & \multicolumn{3}{c|}{\textbf{Seed 42}} & \multicolumn{3}{c|}{\textbf{Seed 43}} & \multicolumn{3}{c|}{\textbf{Seed 41}} & \multicolumn{3}{c|}{\textbf{Seed 42}} & \multicolumn{3}{c}{\textbf{Seed 43}} \\
 & \textbf{5\%} & \textbf{50\%} & \textbf{100\%} & \textbf{5\%} & \textbf{50\%} & \textbf{100\%} & \textbf{5\%} & \textbf{50\%} & \textbf{100\%} & \textbf{5\%} & \textbf{50\%} & \textbf{100\%} & \textbf{5\%} & \textbf{50\%} & \textbf{100\%} & \textbf{5\%} & \textbf{50\%} & \textbf{100\%} \\
\midrule
\midrule
ConvNeXtV2A & -0.40 & -0.05 & 0.07 & -0.40 & -0.05 & 0.07 & -0.40 & -0.05 & 0.07 & -0.26 & -0.37 & -1.32 & -0.26 & -0.37 & -1.32 & -0.26 & -0.37 & -1.32 \\
Scale-MAE & -0.37 & 0.02 & 0.04 & -0.37 & 0.02 & 0.04 & -0.37 & 0.02 & 0.04 & -0.17 & -1.43 & -1.66 & -0.17 & -1.43 & -1.66 & -0.17 & -1.43 & -1.66 \\
DINOv3 Web & \textbf{-0.30} & 0.06 & 0.12 & \textbf{-0.30} & 0.06 & 0.12 & \textbf{-0.30} & 0.06 & 0.12 & \textbf{-0.07} & -1.51 & -1.83 & \textbf{-0.07} & -1.51 & -1.83 & \textbf{-0.07} & -1.51 & -1.83 \\
DINOv3 Sat & -0.33 & 0.08 & 0.15 & -0.33 & 0.08 & 0.15 & -0.33 & 0.08 & 0.15 & -0.14 & -1.11 & -1.23 & -0.14 & -1.11 & -1.23 & -0.14 & -1.11 & -1.23 \\
SatlasNet & \textbf{-0.30} & \textbf{0.11} & 0.18 & \textbf{-0.30} & \textbf{0.11} & 0.18 & \textbf{-0.30} & \textbf{0.11} & 0.18 & -0.09 & -1.01 & -0.93 & -0.09 & -1.01 & -0.93 & -0.09 & -1.01 & -0.93 \\
MPMAE & -0.40 & -0.05 & 0.09 & -0.40 & -0.05 & 0.09 & -0.40 & -0.05 & 0.09 & -0.26 & \textbf{-0.11} & -1.14 & -0.25 & \textbf{-0.12} & -1.14 & -0.25 & \textbf{-0.11} & -1.13 \\
TerraMind S2 & -0.41 & 0.03 & 0.11 & -0.41 & 0.03 & 0.11 & -0.41 & 0.03 & 0.11 & -0.28 & -0.51 & -1.29 & -0.28 & -0.51 & -1.29 & -0.28 & -0.51 & -1.29 \\
TerraMind & -0.40 & 0.05 & 0.09 & -0.40 & 0.05 & 0.09 & -0.40 & 0.05 & 0.09 & -0.24 & -0.87 & -1.56 & -0.24 & -0.87 & -1.56 & -0.24 & -0.87 & -1.56 \\
Copernicus-FM S2 & -0.37 & -0.01 & 0.03 & -0.37 & -0.01 & 0.03 & -0.37 & -0.01 & 0.03 & -0.17 & -1.28 & -1.48 & -0.17 & -1.28 & -1.47 & -0.17 & -1.28 & -1.48 \\
Copernicus-FM & -0.38 & -0.01 & 0.02 & -0.38 & -0.01 & 0.02 & -0.38 & -0.01 & 0.02 & -0.21 & -1.14 & -1.64 & -0.21 & -1.14 & -1.64 & -0.21 & -1.14 & -1.64 \\
Galileo S2 & -0.31 & 0.04 & 0.11 & -0.31 & 0.04 & 0.11 & -0.31 & 0.04 & 0.11 & -0.08 & -1.33 & -1.45 & -0.08 & -1.32 & -1.45 & -0.08 & -1.33 & -1.45 \\
Galileo & -0.33 & 0.05 & 0.13 & -0.33 & 0.05 & 0.13 & -0.33 & 0.05 & 0.13 & -0.13 & -1.21 & -1.17 & -0.13 & -1.21 & -1.17 & -0.13 & -1.21 & -1.17 \\
ConvNeXtV2A-MM & -0.41 & 0.00 & \textbf{0.29} & -0.41 & 0.00 & \textbf{0.29} & -0.41 & 0.00 & \textbf{0.29} & -0.37 & -0.47 & \textbf{-0.22} & -0.37 & -0.47 & \textbf{-0.21} & -0.37 & -0.47 & \textbf{-0.21} \\
\bottomrule
\end{tabular}

}
\label{tab:LP-soil_organic_carbon}
\end{table*}

\begin{table*}
\centering
\caption{Soil pH linear probing test R$^2$}
\resizebox{\linewidth}{!}{%
\begin{tabular}{l|rrr|rrr|rrr|rrr|rrr|rrr}
\toprule
 & \multicolumn{9}{c|}{\textbf{Random}} & \multicolumn{9}{c}{\textbf{Geographic}} \\
 & \multicolumn{3}{c|}{\textbf{Seed 41}} & \multicolumn{3}{c|}{\textbf{Seed 42}} & \multicolumn{3}{c|}{\textbf{Seed 43}} & \multicolumn{3}{c|}{\textbf{Seed 41}} & \multicolumn{3}{c|}{\textbf{Seed 42}} & \multicolumn{3}{c}{\textbf{Seed 43}} \\
 & \textbf{5\%} & \textbf{50\%} & \textbf{100\%} & \textbf{5\%} & \textbf{50\%} & \textbf{100\%} & \textbf{5\%} & \textbf{50\%} & \textbf{100\%} & \textbf{5\%} & \textbf{50\%} & \textbf{100\%} & \textbf{5\%} & \textbf{50\%} & \textbf{100\%} & \textbf{5\%} & \textbf{50\%} & \textbf{100\%} \\
\midrule
\midrule
ConvNeXtV2A & -12.37 & 0.02 & 0.23 & -12.44 & 0.02 & 0.23 & -12.41 & 0.02 & 0.23 & -19.61 & -0.64 & -0.32 & -19.67 & -0.64 & -0.33 & -19.63 & -0.64 & -0.32 \\
Scale-MAE & -1.34 & 0.26 & 0.33 & -1.33 & 0.26 & 0.33 & -1.34 & 0.26 & 0.33 & -2.49 & -0.16 & -0.11 & -2.48 & -0.15 & -0.11 & -2.49 & -0.15 & -0.11 \\
DINOv3 Web & -0.33 & 0.35 & 0.41 & -0.33 & 0.35 & 0.41 & -0.33 & 0.35 & 0.41 & -1.56 & -0.20 & -0.02 & -1.57 & -0.19 & -0.01 & -1.57 & -0.19 & -0.01 \\
DINOv3 Sat & -0.46 & 0.28 & 0.41 & -0.46 & 0.28 & 0.41 & -0.46 & 0.28 & 0.41 & -1.39 & -0.16 & 0.10 & -1.39 & -0.16 & 0.10 & -1.38 & -0.16 & 0.10 \\
SatlasNet & 0.09 & 0.34 & 0.39 & 0.09 & 0.34 & 0.39 & 0.09 & 0.34 & 0.39 & -0.20 & 0.03 & 0.13 & -0.20 & 0.03 & 0.13 & -0.20 & 0.03 & 0.13 \\
MPMAE & -4.64 & 0.20 & 0.37 & -4.64 & 0.20 & 0.37 & -4.64 & 0.20 & 0.37 & -7.51 & -0.45 & -0.14 & -7.52 & -0.44 & -0.14 & -7.52 & -0.44 & -0.14 \\
TerraMind S2 & -6.27 & \textbf{0.51} & \textbf{0.59} & -6.26 & \textbf{0.51} & \textbf{0.59} & -6.27 & \textbf{0.51} & \textbf{0.59} & -10.26 & -0.02 & 0.18 & -10.26 & -0.02 & 0.18 & -10.28 & -0.02 & 0.18 \\
TerraMind & -5.25 & 0.49 & 0.57 & -5.24 & 0.49 & 0.57 & -5.25 & 0.49 & 0.57 & -7.87 & 0.16 & \textbf{0.32} & -7.86 & 0.16 & \textbf{0.32} & -7.86 & 0.16 & \textbf{0.32} \\
Copernicus-FM S2 & -0.49 & 0.41 & 0.46 & -0.49 & 0.41 & 0.46 & -0.49 & 0.41 & 0.46 & -1.01 & 0.08 & 0.16 & -1.01 & 0.08 & 0.16 & -1.01 & 0.08 & 0.16 \\
Copernicus-FM & -1.61 & 0.41 & 0.49 & -1.61 & 0.41 & 0.49 & -1.61 & 0.41 & 0.49 & -2.62 & 0.16 & 0.15 & -2.61 & 0.16 & 0.15 & -2.62 & 0.16 & 0.15 \\
Galileo S2 & \textbf{0.27} & 0.42 & 0.46 & \textbf{0.26} & 0.42 & 0.46 & \textbf{0.27} & 0.42 & 0.46 & \textbf{0.07} & 0.09 & 0.18 & \textbf{0.07} & 0.09 & 0.18 & \textbf{0.07} & 0.09 & 0.18 \\
Galileo & 0.06 & 0.47 & 0.52 & 0.06 & 0.47 & 0.52 & 0.06 & 0.47 & 0.52 & -0.24 & \textbf{0.21} & 0.24 & -0.24 & \textbf{0.21} & 0.24 & -0.24 & \textbf{0.21} & 0.24 \\
ConvNeXtV2A-MM & -10.16 & 0.26 & 0.48 & -10.12 & 0.26 & 0.48 & -10.14 & 0.26 & 0.48 & -14.74 & -1.30 & -0.65 & -14.90 & -1.30 & -0.65 & -14.85 & -1.30 & -0.65 \\
\bottomrule
\end{tabular}

}
\label{tab:LP-soil_pH}
\end{table*}

\begin{table*}
\centering
\caption{Species linear probing test mAP}
\resizebox{\linewidth}{!}{%
\begin{tabular}{l|rrr|rrr|rrr|rrr|rrr|rrr}
\toprule
 & \multicolumn{9}{c|}{\textbf{Random}} & \multicolumn{9}{c}{\textbf{Geographic}} \\
 & \multicolumn{3}{c|}{\textbf{Seed 41}} & \multicolumn{3}{c|}{\textbf{Seed 42}} & \multicolumn{3}{c|}{\textbf{Seed 43}} & \multicolumn{3}{c|}{\textbf{Seed 41}} & \multicolumn{3}{c|}{\textbf{Seed 42}} & \multicolumn{3}{c}{\textbf{Seed 43}} \\
 & \textbf{5\%} & \textbf{50\%} & \textbf{100\%} & \textbf{5\%} & \textbf{50\%} & \textbf{100\%} & \textbf{5\%} & \textbf{50\%} & \textbf{100\%} & \textbf{5\%} & \textbf{50\%} & \textbf{100\%} & \textbf{5\%} & \textbf{50\%} & \textbf{100\%} & \textbf{5\%} & \textbf{50\%} & \textbf{100\%} \\
\midrule
\midrule
ConvNeXtV2A & 0.32 & 0.34 & 0.35 & 0.32 & 0.34 & 0.35 & 0.32 & 0.34 & 0.35 & 0.28 & 0.29 & 0.29 & 0.28 & 0.29 & 0.29 & 0.28 & 0.29 & 0.29 \\
Scale-MAE & 0.37 & 0.48 & 0.50 & 0.37 & 0.48 & 0.50 & 0.37 & 0.48 & 0.50 & 0.28 & 0.31 & 0.32 & 0.28 & 0.31 & 0.32 & 0.28 & 0.31 & 0.32 \\
DINOv3 Web & 0.46 & 0.59 & 0.62 & 0.46 & 0.59 & 0.62 & 0.46 & 0.59 & 0.62 & 0.33 & 0.36 & 0.36 & 0.33 & 0.36 & 0.36 & 0.33 & 0.36 & 0.36 \\
DINOv3 Sat & 0.52 & 0.68 & 0.71 & 0.52 & 0.68 & 0.71 & 0.52 & 0.68 & 0.71 & 0.33 & 0.37 & 0.37 & 0.33 & 0.37 & 0.37 & 0.33 & 0.37 & 0.37 \\
SatlasNet & 0.43 & 0.55 & 0.57 & 0.43 & 0.55 & 0.57 & 0.43 & 0.55 & 0.57 & 0.32 & 0.33 & 0.34 & 0.32 & 0.33 & 0.34 & 0.32 & 0.33 & 0.34 \\
MPMAE & 0.45 & 0.67 & 0.70 & 0.45 & 0.67 & 0.70 & 0.45 & 0.67 & 0.70 & 0.30 & 0.36 & 0.36 & 0.30 & 0.36 & 0.36 & 0.30 & 0.36 & 0.36 \\
TerraMind S2 & \textbf{0.66} & \textbf{0.85} & \textbf{0.87} & \textbf{0.65} & \textbf{0.85} & \textbf{0.87} & \textbf{0.66} & \textbf{0.85} & \textbf{0.87} & \textbf{0.39} & \textbf{0.43} & 0.43 & \textbf{0.39} & \textbf{0.43} & 0.43 & \textbf{0.39} & \textbf{0.43} & 0.43 \\
TerraMind & 0.58 & 0.82 & 0.86 & 0.58 & 0.82 & 0.86 & 0.58 & 0.82 & 0.86 & 0.35 & 0.42 & 0.42 & 0.35 & 0.42 & 0.42 & 0.35 & 0.42 & 0.42 \\
Copernicus-FM S2 & 0.38 & 0.53 & 0.57 & 0.38 & 0.53 & 0.57 & 0.38 & 0.53 & 0.57 & 0.29 & 0.31 & 0.32 & 0.29 & 0.31 & 0.32 & 0.29 & 0.31 & 0.32 \\
Copernicus-FM & 0.44 & 0.70 & 0.78 & 0.44 & 0.70 & 0.78 & 0.44 & 0.70 & 0.78 & 0.31 & 0.40 & 0.43 & 0.31 & 0.40 & 0.43 & 0.31 & 0.40 & 0.43 \\
Galileo S2 & 0.44 & 0.59 & 0.62 & 0.44 & 0.59 & 0.62 & 0.44 & 0.59 & 0.62 & 0.32 & 0.35 & 0.36 & 0.32 & 0.35 & 0.36 & 0.32 & 0.35 & 0.36 \\
Galileo & 0.55 & 0.79 & 0.84 & 0.55 & 0.79 & 0.84 & 0.55 & 0.79 & 0.84 & 0.37 & \textbf{0.43} & \textbf{0.45} & 0.37 & \textbf{0.43} & \textbf{0.45} & 0.37 & \textbf{0.43} & \textbf{0.45} \\
ConvNeXtV2A-MM & 0.52 & 0.79 & 0.81 & 0.52 & 0.79 & 0.81 & 0.52 & 0.79 & 0.81 & 0.37 & 0.42 & 0.42 & 0.37 & 0.42 & 0.42 & 0.38 & 0.42 & 0.42 \\
\bottomrule
\end{tabular}

}
\label{tab:LP-species}
\end{table*}

\end{document}